%% file: main.tex
\documentclass[10pt,twocolumn,letterpaper]{article}

\usepackage{iccv}
\usepackage{times}
\usepackage{epsfig}
\usepackage{graphicx}
\usepackage{amsmath}
\usepackage{amssymb}
\usepackage{gensymb}
\usepackage{subfig}
\usepackage{float}
\usepackage{mathtools}
\usepackage{dblfloatfix}

\usepackage{comment}
\usepackage{multirow}
\usepackage{array}
\usepackage{booktabs}
\usepackage[export]{adjustbox}
\usepackage{xfrac}
\usepackage{stmaryrd}

\usepackage{comment}
\usepackage{multirow}
\usepackage{array}
\usepackage{booktabs}

\usepackage[pagebackref=true,breaklinks=true,letterpaper=true,colorlinks,bookmarks=false]{hyperref}

 \iccvfinalcopy 

\def\iccvPaperID{2553} 

\ificcvfinal\fi

\begin{document}

\title{Full Surround Monodepth from Multiple Cameras} 

\author{
Vitor Guizilini\textsuperscript{1}\textsuperscript{*} \quad
Igor Vasiljevic\textsuperscript{2}\textsuperscript{*} \quad
Rareș Ambruș\textsuperscript{1}  \quad
Greg Shakhnarovich\textsuperscript{2} \quad
Adrien Gaidon\textsuperscript{1} \quad
\\~\\
\quad\quad\quad
\textsuperscript{1}Toyota Research Institute
\quad\quad\quad
\textsuperscript{2}Toyota Technological Institute at Chicago\\
{
\tt\small \{first.lastname\}@tri.global
\quad\quad\quad\quad\quad\quad\quad\quad
\{ivas,greg\}@ttic.edu
\quad\quad\quad
}
}

\maketitle
\thispagestyle{plain}
\pagestyle{plain}

\begin{abstract}
\vspace{-2mm}
\input{sections/abstract}
\end{abstract}

\vspace{-4mm}
\section{Introduction}

\input{sections/introduction}

\section{Related Work}
\input{sections/related}

\section{Methodology}
\input{sections/methodology}

\section{Experiments}
\input{sections/experiments}

\section{Conclusion}
\input{sections/conclusion}

\input{main_suppmat}


{\small
\bibliographystyle{ieee_fullname}
\bibliography{references}
}



\end{document}

%% file: sections/abstract.tex
Self-supervised monocular depth and ego-motion estimation is a promising approach to replace or supplement expensive depth sensors such as LiDAR for robotics applications like autonomous driving.
However, most research in this area focuses on a single monocular camera or stereo pairs that cover only a fraction of the scene around the vehicle.
In this work, we extend monocular self-supervised depth and ego-motion estimation to large-baseline multi-camera rigs.
Using generalized spatio-temporal contexts, pose consistency constraints, and carefully designed photometric loss masking, we learn a single network generating dense, consistent, and scale-aware point clouds that cover the same full surround $360\degree$ field of view as a typical LiDAR scanner.
We also propose a new scale-consistent evaluation metric more suitable to multi-camera settings. 
Experiments on two challenging benchmarks illustrate the benefits of our approach over strong baselines. 

%% file: sections/introduction.tex
\input{figures/teaser}

Self-supervised learning is a promising tool for 3D perception in robotics, forming an integral part of modern state-of-the-art depth estimation architectures~\cite{godard2019digging, gordon2019depth, packnet, zhou2017unsupervised}. With the potential to complement or even replace expensive LiDAR sensors, these 
methods typically take as input a monocular stream of images and produce dense depth and ego-motion predictions. Though recently released 
datasets contain multi-camera data that cover the same full $360\degree$ field of view as LiDAR~\cite{caesar2020nuscenes, packnet}, research has 
focused on forward-facing 
cameras or stereo pairs.
In this paper, we \emph{extend self-supervised depth and ego-motion learning to the general multi-camera setting}, where cameras can have different intrinsics and minimally overlapping regions, as required to minimize the number of cameras on the platform while providing full $360\degree$ coverage.
We describe why stereo-based learning techniques do not apply in this setting, and show that batching cameras independently does not effectively leverage all information available in a multi-camera dataset.

We propose instead to leverage \emph{cross-camera temporal contexts} via \emph{spatio-temporal photometric constraints} to increase the amount of overlap between cameras thanks to the system's ego-motion.
By exploiting known extrinsics between cameras, and enforcing \emph{pose consistency constraints} to ensure all cameras follow the same rigid body motion, we are able to learn \emph{scale-aware models} without any ground-truth depth or ego-motion labels.
Furthermore, our multi-camera constraints enable the prediction of consistent $360\degree$ point clouds, as reflected in our proposed \emph{shared median-scaling} evaluation protocol. 
Finally, we find that \emph{masking out non-overlapping and self-occluded areas} during photometric loss calculation has a drastic impact on performance.

In summary, our contributions are as follows:

\begin{itemize}
    \item We demonstrate, for the first time, self-supervised learning of scale-aware and consistent depth networks in a wide-baseline $360\degree$ multi-camera setting, which we refer to as \textbf{Full Surround Monodepth (FSM)}.    
    \item We introduce 
    key techniques to extend self-supervised depth and ego-motion learning to wide-baseline multi-camera systems: \textbf{multi-camera spatio-temporal contexts} and \textbf{pose consistency constraints}, as well as study the \textbf{impact of non-overlapping and self-occlusion photometric masking} in this novel setting.
    \item We ablate and show the benefits of our proposed approach on \textbf{two publicly available multi-camera datasets}: \emph{DDAD}~\cite{packnet} and \emph{nuScenes}~\cite{caesar2020nuscenes}.
\end{itemize}

%% file: figures/teaser.tex
\begin{figure}[t!]
\centering
\includegraphics[width=0.45\textwidth]{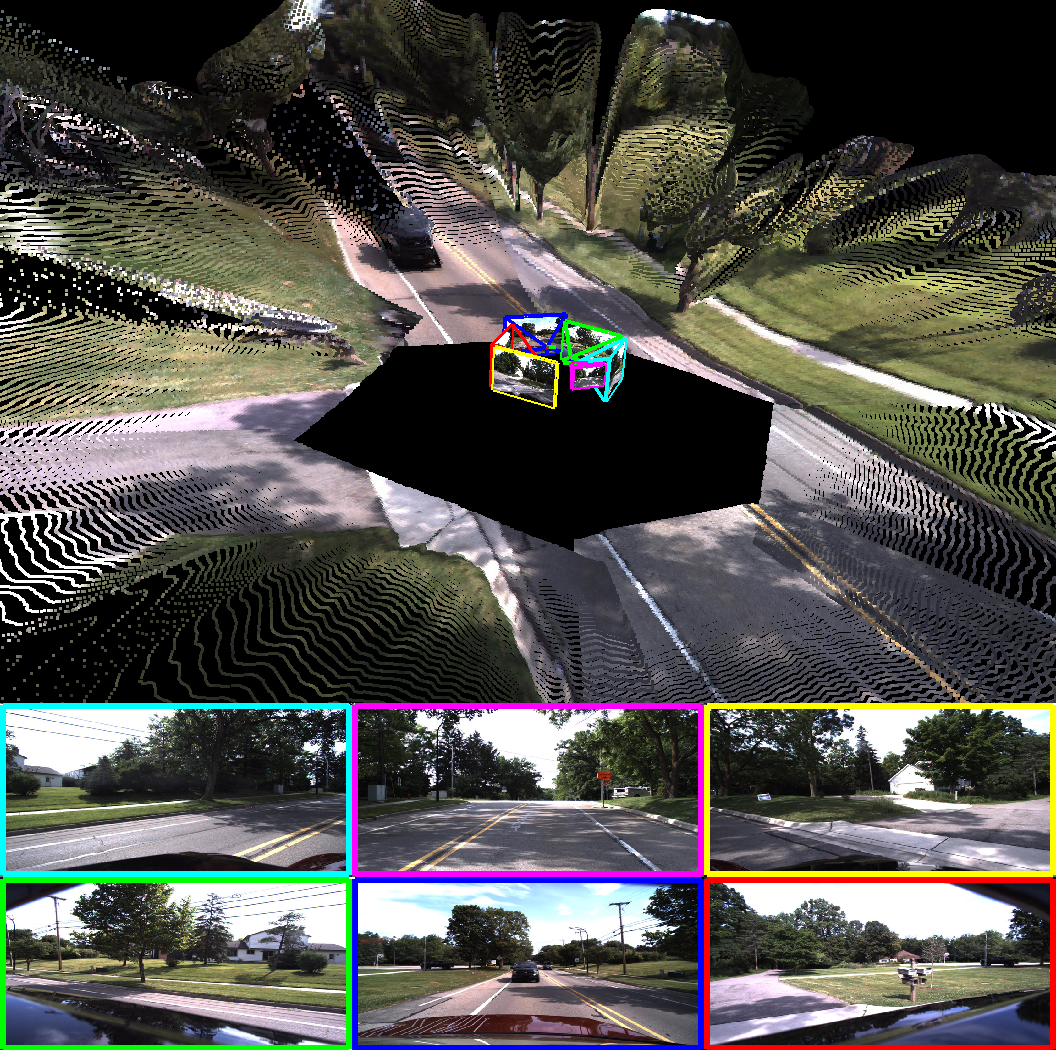}
\caption{\textbf{Consistent scale-aware Full Surround Monodepth (FSM) pointcloud} from multiple cameras.}
\label{fig:teaser}
\vspace{-5mm}
\end{figure}

%% file: sections/related.tex
\textbf{Learning with Stereo Supervision.}
Depth estimation from a rectified stereo pair is a classical task in computer vision~\cite{brown2003advances, saxena2007depth}. In this setting, the 2D matching problem is greatly simplified to a 1D disparity search.  In recent years, supervised stereo depth estimation methods~\cite{kendall2017end, zbontar2016stereo, zhan2018unsupervised}, as well as self-supervised techniques~\cite{garg2016unsupervised, godard2017unsupervised, kuznietsov2017semi}, have emerged as competitive learning-based approaches to this task.  Self-supervised methods take advantage of rectified stereo training data with large overlap to train a disparity estimation network.  Our proposed method is intended for multi-camera configurations with very large baselines (and thus minimal image overlap) where stereo rectification, and by extension disparity estimation, is not feasible.

\textbf{Monocular Depth Estimation.}
Early approaches to learning-based depth estimation were fully supervised~\cite{eigen2015predicting, eigen2014depth, saxena2008make3d}, using datasets collected using IR~\cite{Silberman:ECCV12} or laser scanners~\cite{geiger2012we}. Although achieving impressive results compared to non-learning baselines, these methods suffered from sparsity and high noise levels in the ``ground-truth" data, as well as the need for additional sensors during data collection. The pioneering work of Zhou \etal~\cite{zhou2017unsupervised} introduced the concept of self-supervised learning of depth and ego-motion by casting this problem as a task of view synthesis, using an image reconstruction objective. Further improvements in the view synthesis loss~\cite{godard2019digging} and network architectures~\cite{guizilini20203d}, have lead to accuracy that competes with supervised approaches~\cite{godard2019digging, gordon2019depth, packnet, klingner2020self}.  These learned depth estimators have found applications in several areas, including 3D object detection, where ``pseudo-LIDAR"~\cite{wang2019pseudo} point cloud estimates obtained from monocular depth maps are used to predict 3D bounding boxes.  However, these methods are designed for either monocular or rectified stereo images, and thus only capture a narrow slice of the LiDAR point cloud (typically less than $180\degree$). Consistent multi-camera depth estimation would allow these methods to operate on the full $360\degree$ point cloud annotations.

\textbf{Omnidirectional Depth Estimation.}
A popular approach to $360\degree$ depth estimation is through equirectangular or omnidirectional images~\cite{attal2020matryodshka, bertel2020omniphotos, chen2021distortion, karakottas2019360, sharma2019unsupervised, wang2020bifuse, wang2020360sd}. These methods operate on panoramic images to estimate depth, either monocular or through stereo~\cite{won2019omnimvs}.  For robotics tasks, these images suffer from major disadvantages as a $360\degree$ representation: (1) annotated datasets generally consist of perspective images, making transfer difficult; (2) specialized architectures are necessary; and (3) network training is limited by GPU memory, so resolution must be sacrificed to train using images with such a large field of view.

Catadioptric cameras are an example of an ``omnidirectional'' camera, and a self-supervised generalized camera model was proposed~\cite{vasiljevic2020neural} that produces $360\degree$ point clouds from single images. However, the resolution of catadioptric images drops dramatically at range, while our proposed approach generates much higher resolution pointclouds using perspective cameras.

\textbf{Deep Multi-view Stereo.}
Our multi-camera setting is related to the multi-view stereo (MVS) learning literature, which are generally supervised approaches where learned matching allows a network to predict matching cost-volumes~\cite{huang2018deepmvs, im2019dpsnet, kar2017learning, luo2020attention, xue2019mvscrf}.
Khot \etal~\cite{khot2019learning} relax the supervision requirements and propose a self-supervised MVS architecture, taking insights from self-supervised monocular depth estimation and using a photometric loss.  However, their proposed setting assumes a large collection of images surrounding a single object with known relative pose and large overlap for cost volume computation, and is thus very different from our setting -- our architecture is designed to work with image sequences from any location and with arbitrarily small overlapping between cameras.

%% file: sections/methodology.tex
We first describe the standard approach to single camera monocular self-supervised depth and ego-motion learning. Afterwards, we extend the description to our multi-camera setting and detail our three technical contributions.

\subsection{Single-Camera Monodepth}

Self-supervised depth and ego-motion architectures consist of a depth network that produces depth maps $\hat{D}_{t}$ for a target image $I_t$, as well as a pose network that predicts
the relative pose for pairs of target $t$ and context $c$ frames.  This pose prediction is a rigid transformation $\hat{\mathbf{X}}^{t \to c} = \begin{psmallmatrix}\mathbf{\hat{R}^{t\to c}} & \mathbf{\hat{t}^{t\to c}}\\ \mathbf{0} & \mathbf{1}\end{psmallmatrix} \in \text{SE(3)}$.
We train the networks jointly by minimizing the reprojection error between the actual target image $I_t$ and the synthesized image $\hat{I}_t$, obtaining the latter by projecting pixels from the context image $I_c$ (usually preceding or following $I_t$ in a sequence) onto the target image $I_t$~\cite{zhou2017unsupervised}.  The photometric reprojection loss~\cite{godard2017unsupervised, zhou2017unsupervised} consists of a structure similarity (SSIM) metric and an L1 loss term~\cite{wang2004image}:
\begin{equation}
\small
\mathcal{L}_{p}(I_t,\hat{I_t}) = \alpha~\frac{1 - \text{SSIM}(I_t,\hat{I_t})}{2} + (1-\alpha)~\| I_t - \hat{I_t} \|
\label{eq:photo_mono}
\end{equation}
To synthesize the target image, as in Zhou \etal~\cite{zhou2017unsupervised}, we use STN~\cite{jaderberg2015spatial} via grid sampling with bilinear interpolation. This view synthesis operation is thus fully differentiable, enabling gradient back-propagation for end-to-end training.  We define the pixel-warping operation as:
\begin{equation}
\hat{\mathbf{p}}^t =
\pi \big(\mathbf{\hat{R}}^{t \rightarrow c} \phi (\mathbf{p}^t, \hat{d}^t, \mathbf{K}) + \mathbf{\hat{t}}^{t \rightarrow c}, \mathbf{K}\big)
\label{eq:warp_mono}
\end{equation}
where $\phi(\mathbf{p}, \hat{d},\mathbf{K}) = \mathbf{P}$ is the unprojection of a pixel in homogeneous coordinates $\mathbf{p}$ to a 3D point $\mathbf{P}$ for a given estimated depth $\hat{d}$. Denote the projection of a 3D point back onto the image plane as $\pi(\mathbf{P},\mathbf{K}) = \mathbf{p}$. Both operations require the camera parameters, which for the standard pinhole model  \cite{hartley2003multiple} is defined by the $3 \times 3$ intrinsics matrix $\mathbf{K}$. 

\subsection{Multi-Camera Spatio-Temporal Contexts}
\label{sec:stc}

Multi-camera approaches to self-supervised depth and ego-motion are currently restricted to the stereo setting with rectified images that enable predicting disparities~\cite{godard2017unsupervised}, which are then converted to depth through a known baseline. Although methods have been proposed that combine stereo and monocular self-supervision \cite{godard2019digging, wang2019unos}, directly regressing depth also from stereo pairs, these still assume the availability of highly-overlapping images, from datasets such as KITTI \cite{geiger2012we}. Our proposed approach differs from the stereo setting in the sense that it \textit{does not require stereo-rectified or highly-overlapping images}, but rather is capable of exploiting small overlaps (as low as $10\%$) between cameras with arbitrary locations as a way to both \textit{improve individual camera performance} and generate \textit{scale-aware} models from known extrinsics. Multi-camera rigs with such low overlap are common, e.g., in autonomous driving as a cost-effective solution to $360\degree$ vision~\cite{caesar2020nuscenes, packnet}.

Let $C_i$ and $C_j$ be two cameras with extrinsics $\mathbf{X}_i$ and $\mathbf{X}_j$, and intrinsics $\mathbf{K}_i$ and $\mathbf{K}_j$. Denoting the relative extrinsics as $\mathbf{X}_{i \rightarrow j}$ and abbreviating $\phi_i(\mathbf{p}, \hat{d}) = \phi(\mathbf{p}, \hat{d}, \mathbf{K}_i)$ and $\pi_i(\mathbf{P}) = \pi(\mathbf{P}, \mathbf{K}_i)$,
we can use Equation~\ref{eq:warp_mono} to warp images from these two cameras:

\begin{equation}
\hat{\mathbf{p}}_i =
\pi_j \big(\mathbf{R}_{i \rightarrow j} \phi_i (\mathbf{p}_i, \hat{d}_i) + \mathbf{t}_{i \rightarrow j}\big)
\label{eq:warp_spatial}
\end{equation}
Note that the above equation is purely \textit{spatial}, since it warps images between different cameras taken at the same timestep. Conversely, Equation~\ref{eq:warp_mono} is purely \textit{temporal}, since it is only concerned with warping images from the same camera taken at different timesteps. 

Therefore, for any given camera $C_i$ at a timestep $t$, a context image can be either temporal (i.e., from adjacent frames $t-1$ and $t+1$) or spatial (i.e., from any camera $j$ that overlaps with $i$). This allows us to further generalize the concept of ``context image'' in self-supervised learning to also include temporal contexts from other overlapping cameras. This is done by warping images between different cameras taken at different timesteps using a composition of known extrinsics with predicted ego-motion:
\begin{equation}
\hat{\mathbf{p}}^t_i =
\pi_j \big(\mathbf{R}_{i \rightarrow j}\big(\hat{\mathbf{R}}^{t \rightarrow c}_j \phi_j (\mathbf{p}^t_j, \hat{d}^t_j) + \hat{\mathbf{t}}^{t \to c}_j\big) + \mathbf{t}_{i \rightarrow j}\big)
\label{eq:warp_multi}
\end{equation}

\input{figures/multicam_diagram}

\input{figures/ddad_warps}

A diagram depicting such transformations can be found in Figure~\ref{fig:multicam_diagram}, and Figure~\ref{fig:ddad_warps} shows examples of warped images and corresponding photometric losses using the \textit{DDAD} dataset \cite{packnet}. Particularly, the fifth and sixth rows show examples of multi-camera photometric losses using purely spatial contexts (Equation~\ref{eq:warp_spatial}) and our proposed spatio-temporal contexts (Equation~\ref{eq:warp_multi}). 
As we can see, \emph{spatio-temporal contexts (STC) promote a larger overlap between cameras and smaller residual photometric loss}, due to occlusions and changes in brightness and viewpoint. This improved photometric loss leads to \emph{better self-supervision} for depth and ego-motion learning in a multi-camera setting, as validated in experiments.

\subsection{Multi-Camera Pose Consistency Constraints}
\label{sec:pcc}

Beyond cross-camera constraints due to image overlap, there are also natural pose constraints due to the fact that all cameras are rigidly attached to a single vehicle (i.e., relative camera extrinsics are constant and fixed).  Specifically, the pose network is used to predict independent poses for each camera, even though they should correspond to the same transformation, just in a different coordinate frame.
For a given camera $i$, the pose network predicts its transformation $\hat{\mathbf{X}}^{t \to t+1}_{i}$ from the current frame $t$ to the subsequent frame $t+1$. In order to obtain predictions from different cameras that are in the same coordinate frame, we transform this prediction to the coordinate frame of a ``canonical'' camera $C_j$. We denote $\hat{\mathbf{X}}^{t \to t+1}_{i}$ in $C_j$ coordinates as $\tilde{\mathbf{X}}^{t \to t+1}_{i}$.

To convert a predicted transformation $\hat{\mathbf{X}}^{t \to t+1}_{i}$ from the coordinate frame of camera $C_i$ to camera $C_j$, we can use the extrinsics $\mathbf{X}_{i}$ and $\mathbf{X}_{j}$ to generate $\tilde{\mathbf{X}}^{t \to t+1}_{i}$ as follows:
\begin{equation}
    \tilde{\mathbf{X}}_{i}^{t \to t+1} = \mathbf{X}_j^{-1}\mathbf{X}_i\hat{\mathbf{X}}_{i}^{t \to t+1}\mathbf{X}_i^{-1}\mathbf{X}_j
\end{equation}
where $\tilde{\mathbf{X}}_{i}^{t \to t+1} = \begin{psmallmatrix}{\tilde{\mathbf{R}}}^{t \to t+1}_{i} & {\tilde{\mathbf{t}}}^{t \to t+1}_i\\ \mathbf{0} & \mathbf{1}\end{psmallmatrix}$.  As a convention, we convert all predicted transformations to the coordinate frame of the front camera $C_1$. Once all predictions are in the same coordinate frame, we constrain the translation vectors $\mathbf{t}$ and rotation matrices $\mathbf{R}$ to be similar across all cameras.

\noindent\textbf{Translation.} We constrain all predicted translation vectors to be similar to the prediction for the \emph{front camera}, which generally performs best across all experiments. Defining the predicted front camera translation vector as $\hat{\mathbf{t}}^{t+1}_{1}$, for $N$ cameras the translation consistency loss is given by:
\begin{equation}
    \mathbf{t}_{\mathit{loss}} = \sum_{j=2}^{N}\|\hat{\mathbf{t}}^{t+1}_{1} - \tilde{\mathbf{t}}^{t+1}_{j} \|^2 
\end{equation}
\textbf{Rotation.} Similarly, we want to constrain other cameras to predict a rotation matrix similar to the front camera. To accomplish that, once the predictions are in the same coordinate frame we convert them to Euler angles $(\phi_i, \theta_i, \psi_i)$ and calculate the rotation consistency loss such that: 
\begin{equation}
    \mathbf{R}_{\mathit{loss}} = \sum_{j=2}^{N}\|\hat{\phi}_1 - \tilde{\phi}_j \|^2 + \|\hat{\theta}_1 - \tilde{\theta}_j \|^2 + \|\hat{\psi}_1 - \tilde{\psi}_j \|^2 
\end{equation}
Similar to the trade-off between rotation losses and translation loss in the original PoseNet~\cite{kendall2015posenet, kendall2017geometric}, we trade off between the two constraints by defining $\mathcal{L}_{\mathit{pcc}} = \alpha_{t}\mathbf{t}_{\mathit{loss}} + \alpha_{r}\mathbf{R}_{\mathit{loss}}$, where $\alpha_{t}$ and $\alpha_r$ are weight coefficients.

\subsection{The Importance of Masks}
\label{sec:mask}

The photometric loss, as used for self-supervised monocular depth and ego-motion learning, has several assumptions that are not entirely justified in real-world scenarios. These include the static world assumption (violated by dynamic objects), brightness constancy (violated by luminosity changes), and dense overlap between frames (violated by large viewpoint changes). Although several works have been proposed to relax some of these assumptions~\cite{gordon2019depth}, more often than not methods are developed to mask out those regions \cite{godard2019digging}, to avoid spurious information from contaminating the final model. 

In a multi-camera setting there are two scenarios that are particularly challenging for self-supervised depth and ego-motion learning: \textit{non-overlapping areas}, due to large changes in viewpoint between cameras, and \textit{self-occlusions}, due to camera positioning that results in the platform (i.e., ego-vehicle) partially covering the image.  
Here we describe how our proposed approach addresses each of these scenarios.
The final masked photometric loss used during training (Equation~\ref{eq:photo_mono}) takes the form: 

\begin{equation}
    \mathcal{L}_{\mathit{mp}}(I_t,\hat{I_t}) = \mathcal{L}_{p}(I_t,\hat{I_t}) \odot \mathcal{M}_{\mathit{no}} \odot \mathcal{M}_{\mathit{so}}
    \label{eq:overall-loss}
\end{equation}
where $\odot$ denotes element-wise multiplication, and $\mathcal{M}_{\mathit{no}}$ and $\mathcal{M}_{\mathit{so}}$ are binary masks respectively for non-overlapping and self-occluded areas, as described below.

\input{figures/masking}

\subsubsection{Non-Overlapping Areas}

We generate non-overlapping area masks by jointly warping with each context image a unitary tensor of the same spatial dimensions, using nearest-neighbor interpolation. The warped tensor is used to mask the photometric loss, thus avoiding gradient backpropagation in areas where there is no overlap between frames. Note that this unitary warping also uses network predictions, and therefore is constantly updated at training time. This is similar to the motion masks described in~\cite{mahjourian2018unsupervised}, however here we extend this concept to a spatio-temporal multi-camera setting. Figure~\ref{fig:ddad_warps} shows examples of spatial and temporal non-overlapping masks on the \textit{DDAD} dataset, with a trained model. As expected, temporal contexts have a large amount of frame-to-frame overlap ($>90\%$), even considering side-pointing cameras. Spatial context overlaps, on the other hand, are much smaller ($10-20\%$), due to radically different camera orientations.  

\subsubsection{Self-Occlusions}
\label{sec:self-occ}
A common technique in self-supervised learning is the ``auto-mask'' procedure, which filters out pixels with a synthesized reprojection error higher than the original source image~\cite{godard2019digging}.  This mask particularly targets the ``infinite depth problem'', which occurs when scene elements move at the same speed as the vehicle, causing zero parallax and thus an infinitely-far away depth estimate for that region.  However, this technique assumes brightness constancy, and the self-occlusions created by the robot (or car) body are often highly specular (Figure~\ref{fig:mask_rgb}), especially in the case of modern passenger vehicles. 

Using a network with auto-masking enabled, specular self-occlusions create serious errors in the depth predictions, as shown in Figure~\ref{fig:mask_without}.  We propose a simpler approach, consisting of creating manual masks for each camera (this needs only to be done once, assuming that the extrinsics remain constant).  As shown in Figure~\ref{fig:mask_with}, and ablated in experiments, the introduction of these self-occlusion masks results in a substantial improvement in overall performance, to the point of enabling self-supervised depth and ego-motion learning under these conditions. 
Interestingly, self-occluded areas in the predicted depth maps are correctly ``inpainted" to include the hidden ground plane.  We posit that this is due to multi-camera training, in which a region unoccluded in one camera can be used to resolve self-occlusions in another camera. 

%% file: figures/multicam_diagram.tex
\begin{figure}[t!]
    \centering
    \subfloat{
    \includegraphics[width=0.9\linewidth, height=4.5cm, trim=0 10 0 0, clip]{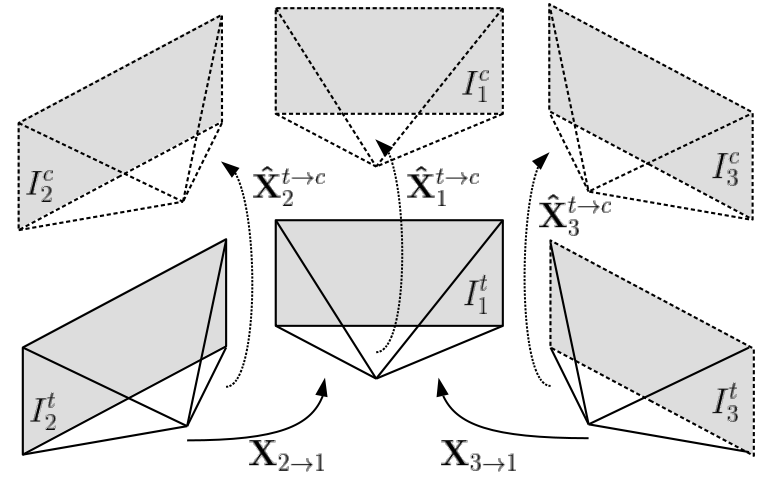}
    }
    \caption{\textbf{Multi-camera spatio-temporal transformation matrices}. Solid cameras are \textit{target} (current frames), and dotted cameras are \textit{context} (adjacent frames). Spatial transformations ($\mathbf{X}_{i \rightarrow 1}$) are obtained from known extrinsics, and temporal transformations ($\hat{\mathbf{X}}^{\mathit{t} \rightarrow \mathit{c}}_i$) from the pose network.}
    \label{fig:multicam_diagram}
\vspace{-3mm}
\end{figure}

%% file: figures/ddad_warps.tex
\begin{figure*}[t!]
\centering
\subfloat{
\textcolor{green}{\rule{2.62cm}{1.5mm}}
\textcolor{cyan}{\rule{2.62cm}{1.5mm}}
\textcolor{magenta}{\rule{2.62cm}{1.5mm}}
\textcolor{yellow}{\rule{2.62cm}{1.5mm}}
\textcolor{red}{\rule{2.62cm}{1.5mm}}
\textcolor{blue}{\rule{2.62cm}{1.5mm}}
}
\vspace{-4.0mm}
\\
\subfloat{
\includegraphics[width=0.15\linewidth,height=1.6cm]{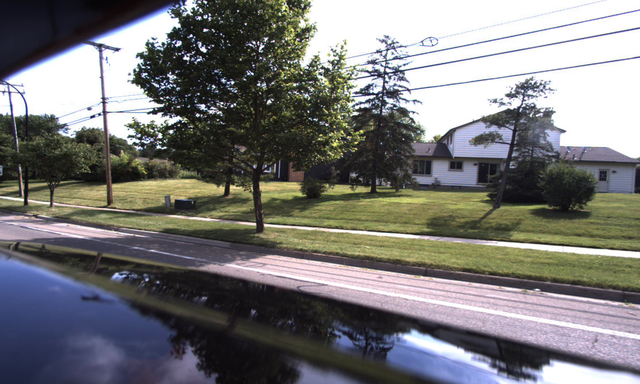}
\includegraphics[width=0.15\linewidth,height=1.6cm]{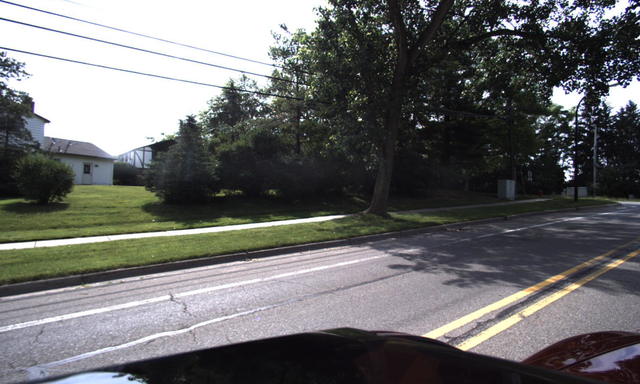}
\includegraphics[width=0.15\linewidth,height=1.6cm]{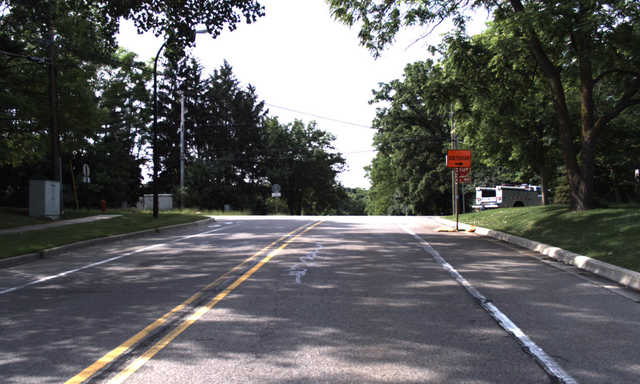}
\includegraphics[width=0.15\linewidth,height=1.6cm]{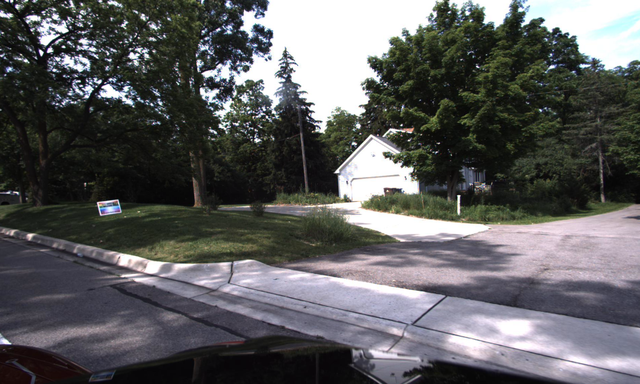}
\includegraphics[width=0.15\linewidth,height=1.6cm]{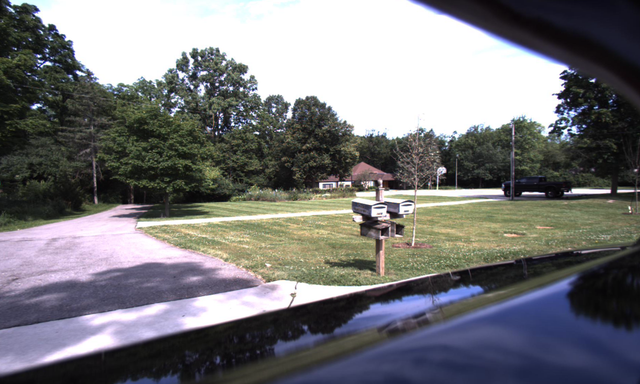}
\includegraphics[width=0.15\linewidth,height=1.6cm]{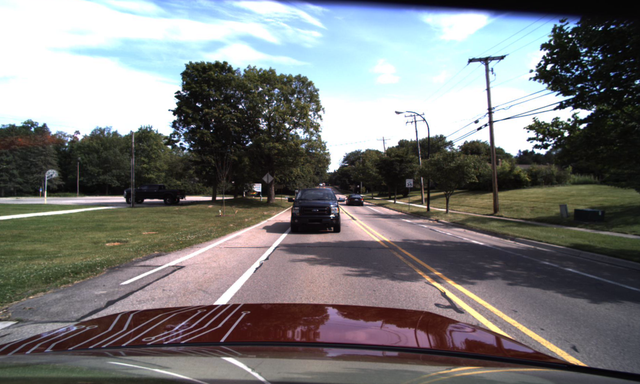}
}
\vspace{-4.5mm}
\\
\subfloat{
\includegraphics[width=0.15\linewidth,height=1.6cm]{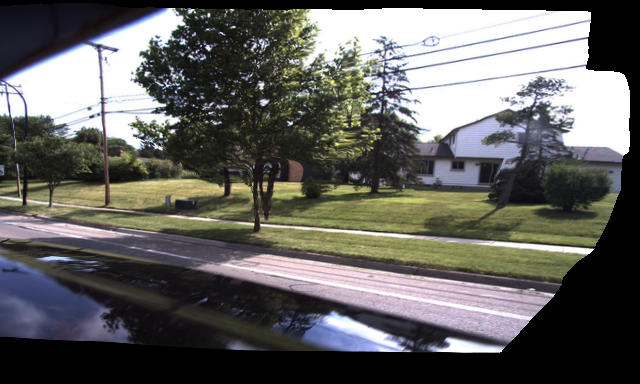}
\includegraphics[width=0.15\linewidth,height=1.6cm]{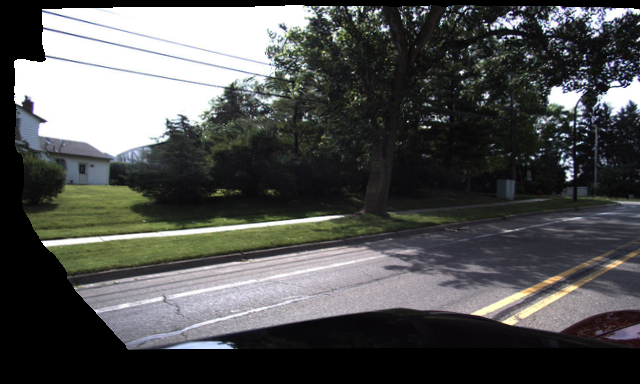}
\includegraphics[width=0.15\linewidth,height=1.6cm]{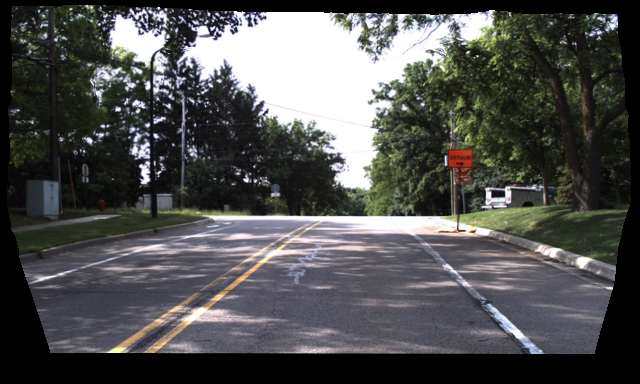}
\includegraphics[width=0.15\linewidth,height=1.6cm]{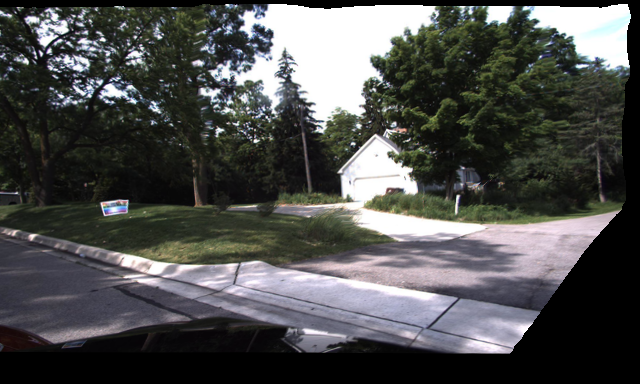}
\includegraphics[width=0.15\linewidth,height=1.6cm]{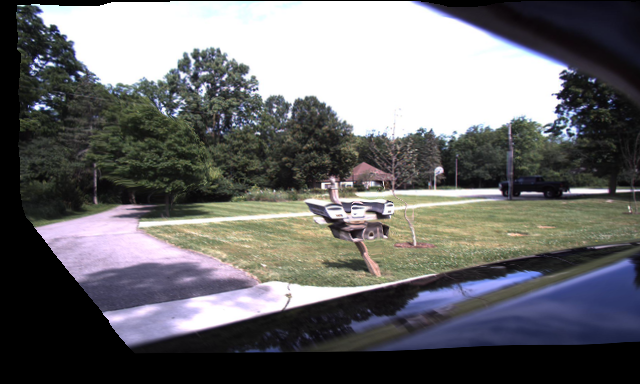}
\includegraphics[width=0.15\linewidth,height=1.6cm]{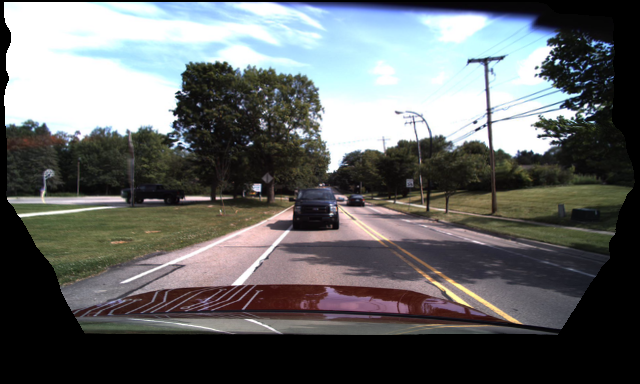}
}
\vspace{-4.5mm}
\\
\subfloat{
\includegraphics[width=0.15\linewidth,height=1.6cm]{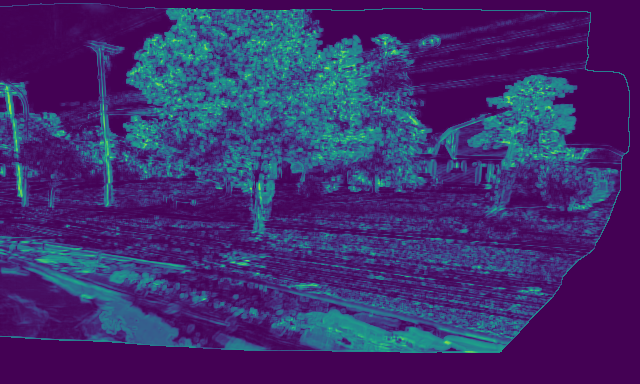}
\includegraphics[width=0.15\linewidth,height=1.6cm]{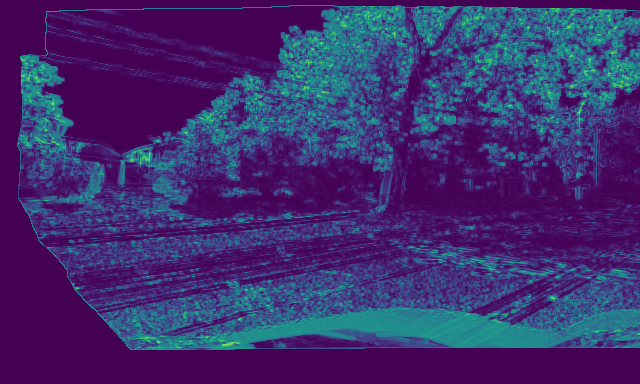}
\includegraphics[width=0.15\linewidth,height=1.6cm]{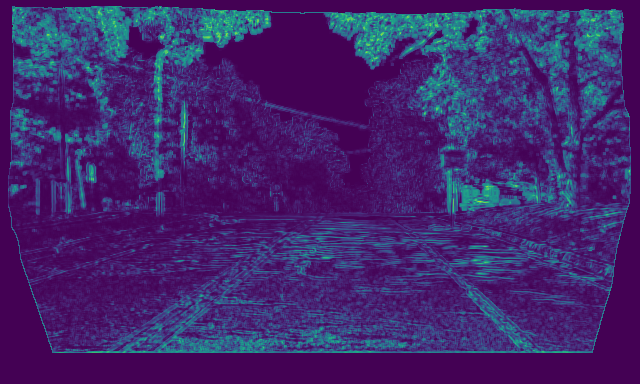}
\includegraphics[width=0.15\linewidth,height=1.6cm]{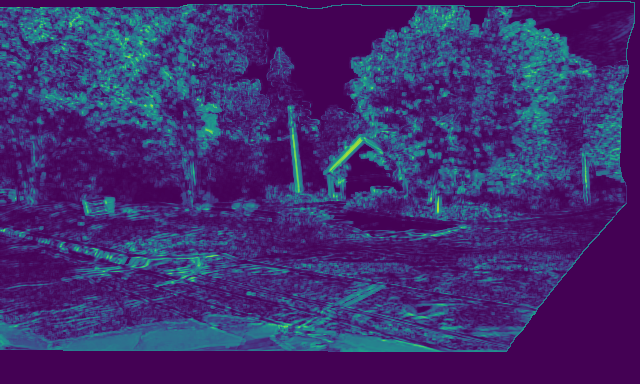}
\includegraphics[width=0.15\linewidth,height=1.6cm]{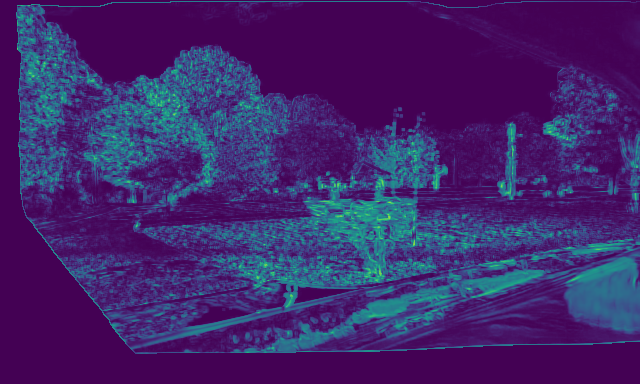}
\includegraphics[width=0.15\linewidth,height=1.6cm]{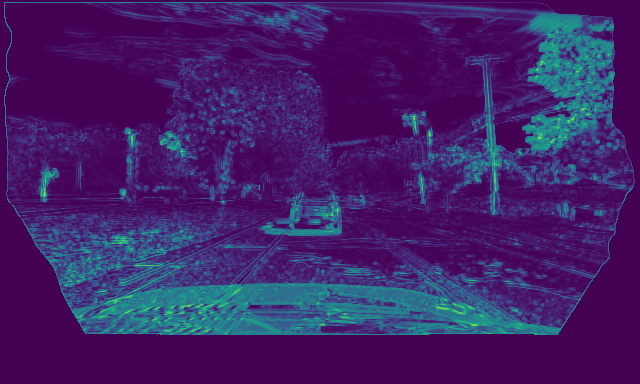}
}
\vspace{-4.5mm}
\\
\subfloat{
\includegraphics[width=0.15\linewidth,height=1.6cm]{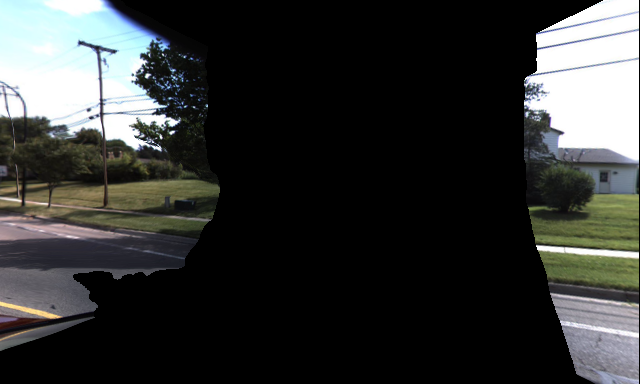}
\includegraphics[width=0.15\linewidth,height=1.6cm]{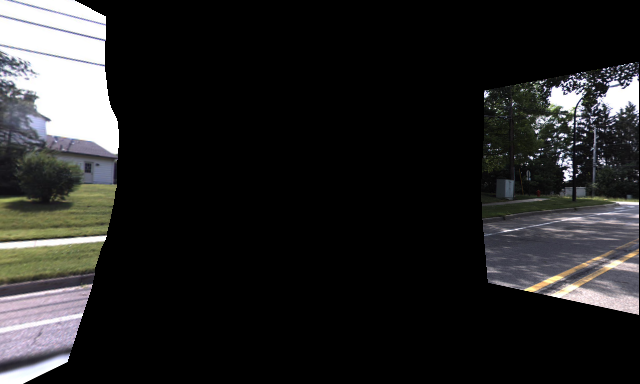}
\includegraphics[width=0.15\linewidth,height=1.6cm]{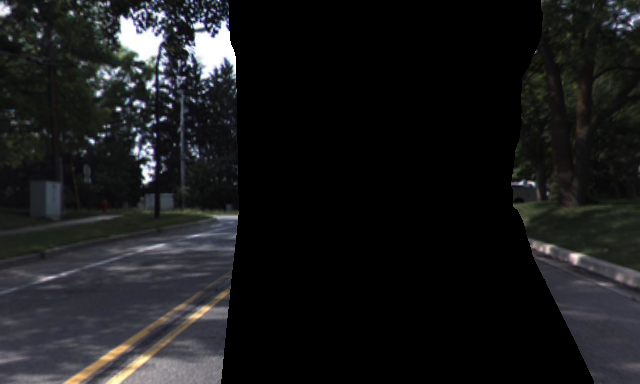}
\includegraphics[width=0.15\linewidth,height=1.6cm]{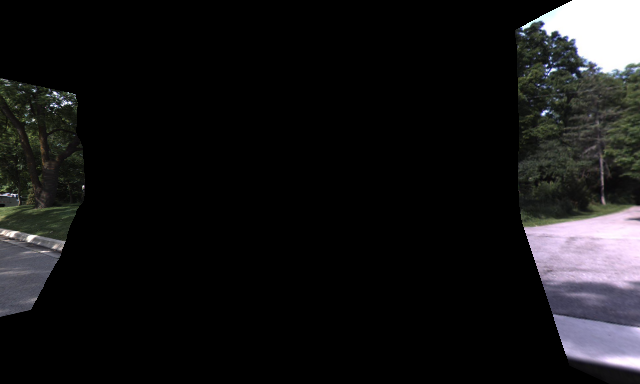}
\includegraphics[width=0.15\linewidth,height=1.6cm]{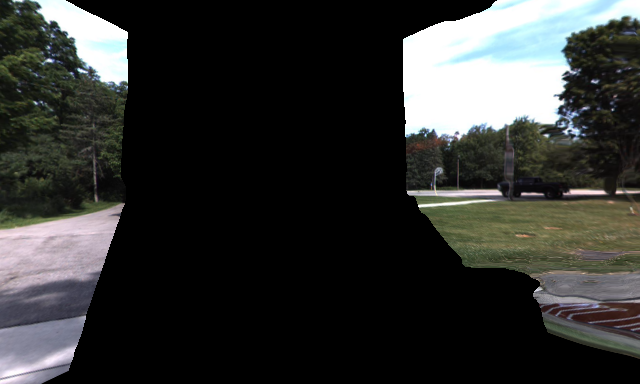}
\includegraphics[width=0.15\linewidth,height=1.6cm]{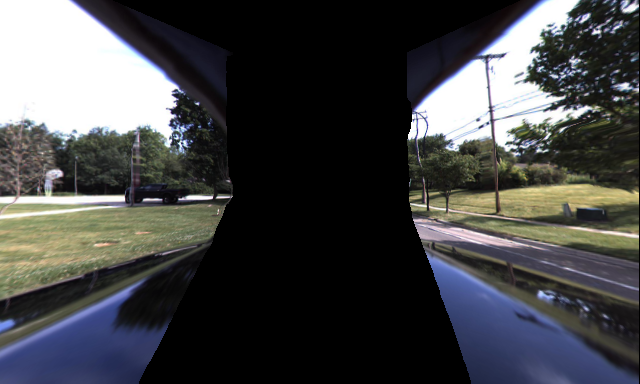}
}
\vspace{-4.5mm}
\\ 
\subfloat{
\includegraphics[width=0.15\linewidth,height=1.6cm]{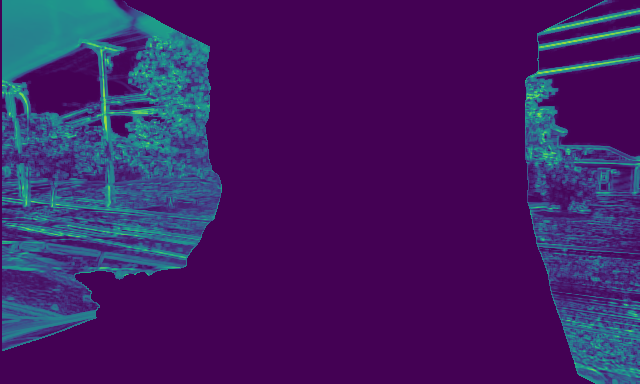}
\includegraphics[width=0.15\linewidth,height=1.6cm]{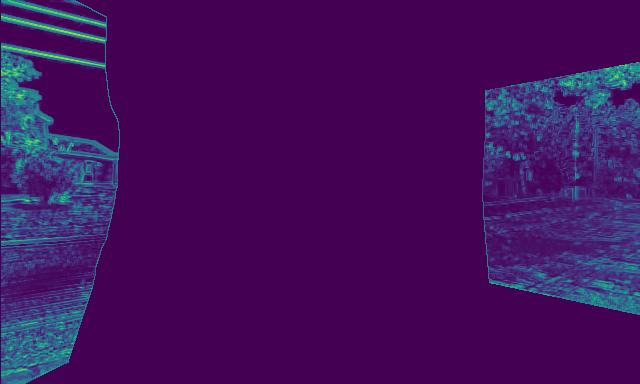}
\includegraphics[width=0.15\linewidth,height=1.6cm]{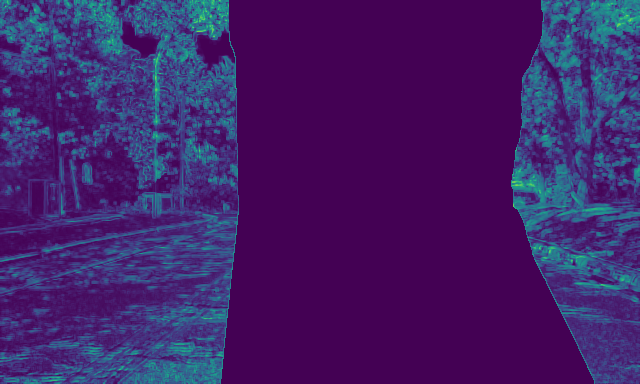}
\includegraphics[width=0.15\linewidth,height=1.6cm]{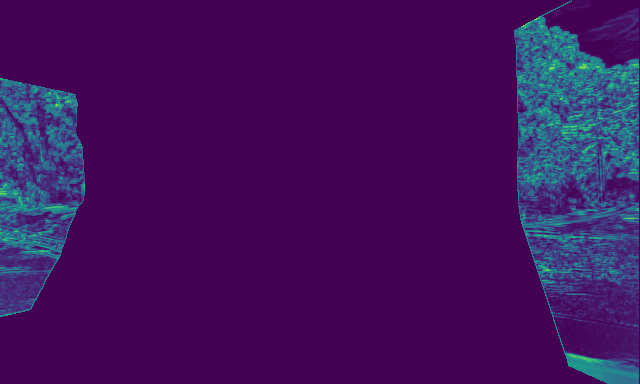}
\includegraphics[width=0.15\linewidth,height=1.6cm]{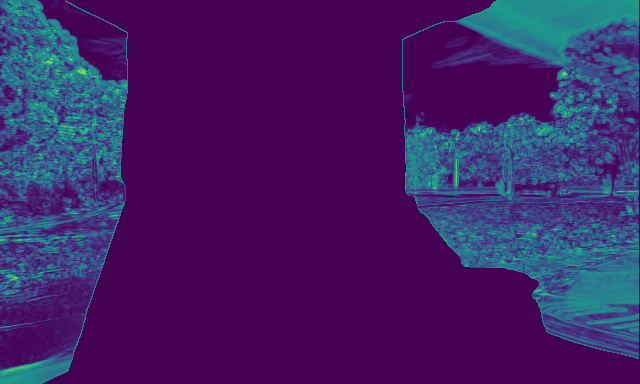}
\includegraphics[width=0.15\linewidth,height=1.6cm]{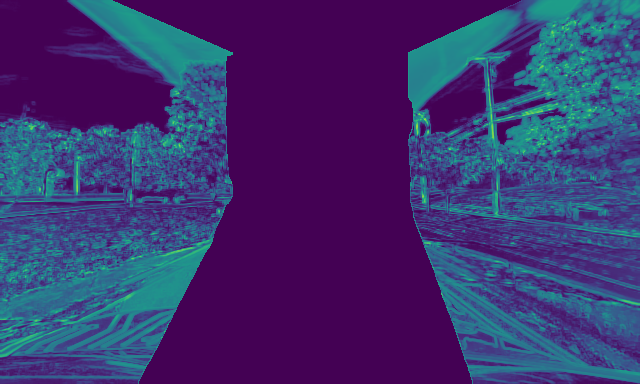}
}
\vspace{-4.5mm}
\\ 
\subfloat{
\includegraphics[width=0.15\linewidth,height=1.6cm]{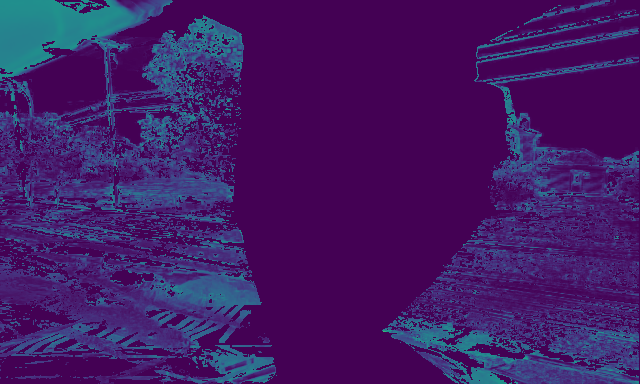}
\includegraphics[width=0.15\linewidth,height=1.6cm]{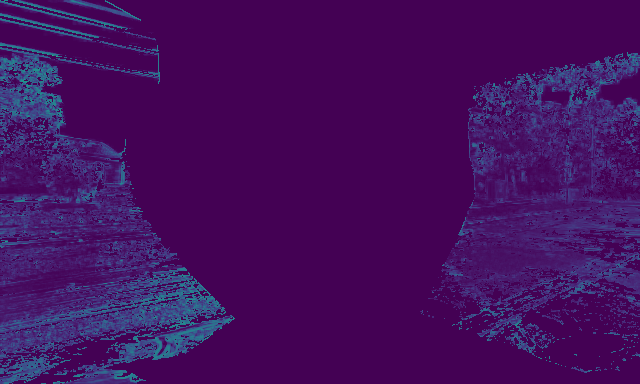}
\includegraphics[width=0.15\linewidth,height=1.6cm]{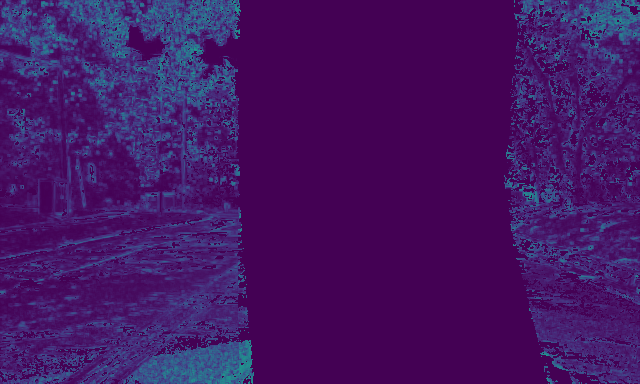}
\includegraphics[width=0.15\linewidth,height=1.6cm]{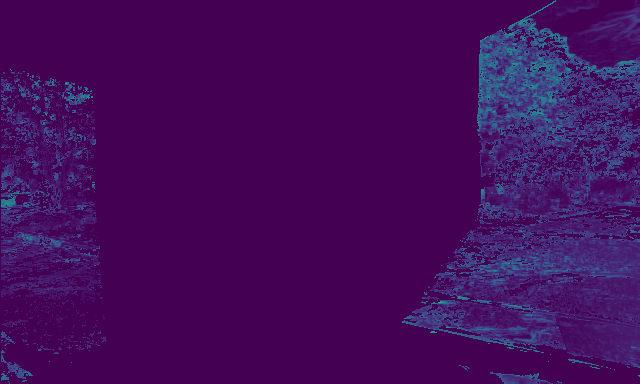}
\includegraphics[width=0.15\linewidth,height=1.6cm]{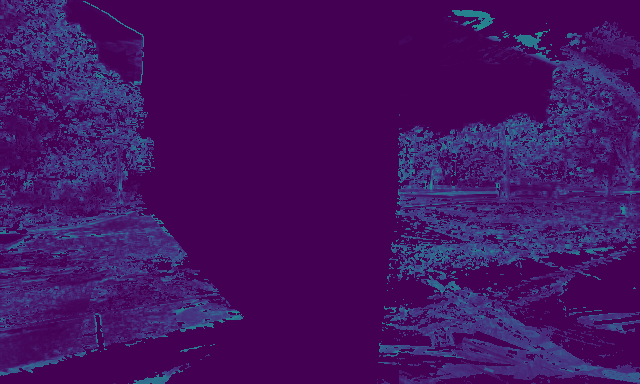}
\includegraphics[width=0.15\linewidth,height=1.6cm]{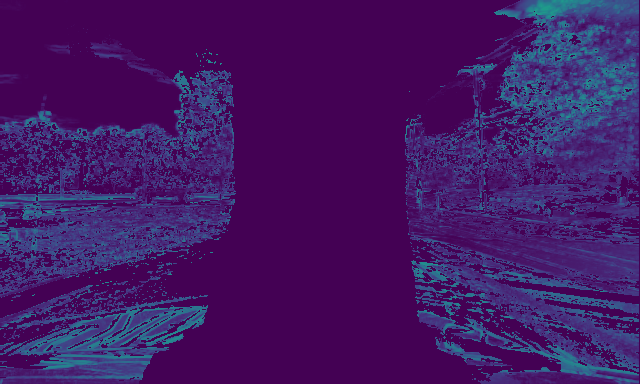}
}
\caption{\textbf{Examples of spatial and temporal image warping} on the \textit{DDAD} dataset (camera colors from Figure~\ref{fig:teaser}, clockwise). \textbf{First row:} Input RGB images. \textbf{Second and third rows:} Synthesized views from temporal contexts (Equation~\ref{eq:warp_mono}) and photometric losses. \textbf{Fourth and fifth rows:} Synthesized views from surrounding cameras (Equation~\ref{eq:warp_spatial}), and photometric losses using only spatial contexts. \textbf{Sixth row} Photometric losses using our proposed spatio-temporal contexts. By also leveraging temporal contexts during cross-camera photometric warping, we are able to generate larger overlapping areas between images, as well as a smaller residual photometric error (darker colors) for optimization.}
\label{fig:ddad_warps}
\vspace{-5mm}
\end{figure*}

%% file: figures/masking.tex
\begin{figure}[t!]
\centering
\subfloat[Input RGB image.]{
\includegraphics[width=0.48\linewidth]{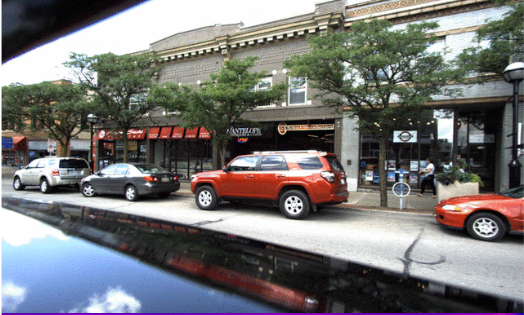}
\label{fig:mask_rgb}
}
\subfloat[Self-occlusion mask.]{
\includegraphics[width=0.48\linewidth]{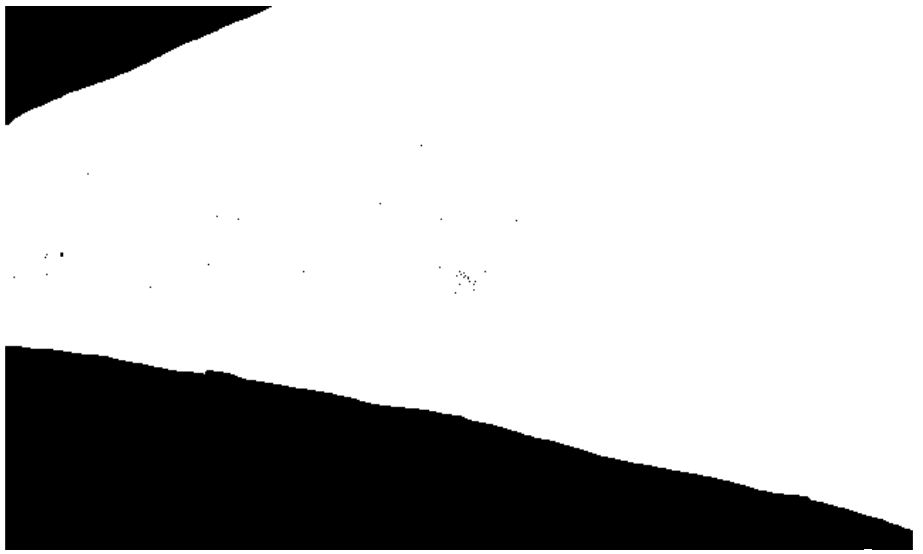}
\label{fig:mask_binary}
}
\vspace{-3mm}
\\
\subfloat[Without self-occlusion masking.]{
\includegraphics[width=0.48\linewidth]{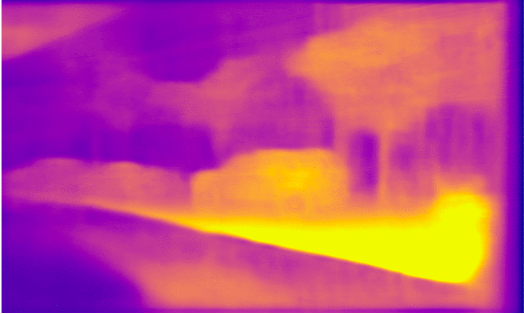}
\label{fig:mask_without}
}
\subfloat[With self-occlusion masking.]{
\includegraphics[width=0.48\linewidth]{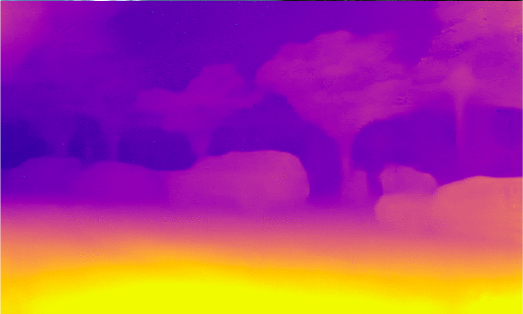}
\label{fig:mask_with}
}
\caption{\textbf{Impact of self-occlusion masks on depth estimation} on the \textit{DDAD} dataset. These masks remove self-occluded regions from the self-supervised photometric loss, enabling easier optimization (lower loss, cf. Figure~\ref{fig:ddad_warps}) and better generalization (e.g., on the ground plane).}
\label{fig:self_occ_masks}
\vspace{-3mm}
\end{figure}


%% file: sections/experiments.tex
\subsection{Datasets}

Traditionally, self-supervised depth and ego-motion learning uses monocular sequences~\cite{zhou2017unsupervised, godard2019digging, gordon2019depth, packnet} or rectified stereo pairs~\cite{godard2019digging, superdepth} from forward-facing cameras in the KITTI~\cite{geiger2012we} dataset. Recently, several datasets have been released with synchronized multi-camera sequences that cover the entire surrounding of the ego-vehicle~\cite{caesar2020nuscenes, packnet}. We focus on these datasets for our experiments, showing that our proposed approach, FSM, produces substantial improvements across all cameras.
For more information on these datasets, please refer to the supplementary material.

\textbf{KITTI \cite{geiger2012we}}.
The KITTI dataset is the standard benchmark for depth and ego-motion estimation. Although it only contains forward-facing stereo pairs, we show that FSM accommodates the special case of high-overlapping rectified images, achieving competitive results with stereo methods.

\textbf{DDAD \cite{packnet}}. 
The DDAD dataset is the main focus of our experiments, since it contains six cameras with relatively small overlap and highly accurate dense ground-truth depth maps for evaluation. We show that, by jointly training FSM on all cameras, we considerably improve results and establish a new state of the art on this dataset by a large margin. 

\textbf{NuScenes \cite{caesar2020nuscenes}}. 
The nuScenes dataset is a popular benchmark for 2D and 3D object detection, as well as semantic and instance segmentation.  However, it is a challenging dataset for self-supervised depth estimation because of the relatively low resolution of the images, very small overlap between the cameras, high diversity of weather conditions and time of day, and unstructured environments. 
We show that FSM is robust enough to overcome these challenges and substantially improve results relative to the baseline. 

\subsection{Multi-Camera Depth Evaluation Metrics}

Our approach is \emph{scale-aware}, due to its use of known extrinsics to generate metrically accurate predictions. In contrast, existing self-supervised monocular depth and ego-motion architectures learn up-to-scale models, resorting to \textit{median-scaling} at test time in order to compare depth predictions against the (scaled) ground truth.  Median scaling ``borrows'' the true scale of ground truth by multiplying each depth estimate by $\frac{\mathit{med}(D_{\mathit{gt}})}{\mathit{med}(D_{\mathit{pred}})}$, where $\mathit{med}$ is the median operation and $D_{\mathit{gt}}$ and $D_{\mathit{pred}}$ are ground-truth and predicted depth maps.  This scaling enables quantitative evaluation, but requires ground truth information at test time, limiting the utility of such methods in real-world applications.

Furthermore, median-scaling hides certain undesired behaviors, such as frame-to-frame variability \cite{bian2019depth}. This is exacerbated in the setting proposed in this paper, where multiple images are used to generate a single, consistent pointcloud. If median-scaling is applied on a per-frame basis, the resulting depth maps will hide any scale discrepancies between camera predictions, which will not be reflected in the quantitative evaluation (Figure \ref{fig:pcl_align}). Thus, instead of the standard median-scaling protocol, we propose to use a single scale factor $\gamma$ shared by all $N$ considered cameras defined as:
\begin{equation}
\small
    \gamma = \frac
    {\mathit{med}\big( \{ D_{\mathit{gt}}^1 , \cdots , D_{\mathit{gt}}^N \} \big) }
    {\mathit{med}\big( \{ D_{\mathit{pred}}^1 , \cdots , D_{\mathit{pred}}^N \} \big) }
\label{eq:sharedmedscaling}
\end{equation}

\input{figures/pcl_align}
This is similar to single median-scaling \cite{godard2019digging}, in which the same factor is used to scale predictions for the entire \emph{dataset}.  In our setting, because multiple images are considered jointly, we instead produce a shared factor to scale all predictions at that \emph{timestep}. This forces all predicted depth maps in the same timestep to have the same scale factor (and thus be relatively consistent), with any deviation reflected in the calculated metrics. 
In practice, as our method is scale-aware, we report metrics both with (for comparison with baselines) and without median-scaling.

\subsection{Networks} 
For all of our experiments, we used a ResNet18-based depth and pose networks, based on \emph{monodepth2}~\cite{godard2019digging}.  For more details regarding network architectures and training schedules, please refer to our supplementary material. 

\input{tables/table_stereo}

\subsection{Stereo Methods}
Though our proposed approach is intended for multi-camera rigs with small overlap, it can also be used without modification on stereo datasets, allowing us to learn depth directly without rectifying the images for disparity learning. We show in Table \ref{tab:stereo} that despite the greater simplicity of stereo rectification on the KITTI dataset, our approach remains competitive with explicitly stereo methods.

\subsection{Single-Camera Methods}

Given that the majority of self-supervised monocular depth estimation papers focuses on single-image sequences with forward-facing cameras, we consider a variety of alternative baselines for our quantitative evaluation. In particular, we pick two state-of-the-art published methods: \emph{monodepth2} \cite{godard2019digging}, that uses a simpler architecture with a series of modifications to the photometric loss calculation; and \emph{PackNet} \cite{packnet}, that proposes an architecture especially designed for self-supervised depth learning.  We use results of these methods for the front camera, as reported by Guizilini \etal~\cite{packnet}, as baselines for our method.

\input{tables/ddad_depth}

We also take inspiration from the ``Learning SfM from SfM'' work~\cite{klodt2018supervising, li2018megadepth} and employ COLMAP~\cite{schonberger2016structure}, a state-of-the-art structure-from-motion system, on the unlabeled DDAD~\cite{packnet} training split to generate predicted depth maps, that are then used as pseudo-labels for supervised learning.  Note that, while this approach is also self-supervised (i.e., there is no ground-truth depth), it requires substantially more computation, since it processes all images from any given sequence simultaneously with a series of bundle adjustment techniques to produce a single reconstruction of the entire scene. For more details regarding this baseline, please refer to the supplementary material. 

The results of these experiments are summarized in Table \ref{table:ddad_depth}. Compared to other self-supervised methods on monocular sequences, our masking procedures for multi-camera training already significantly improve results from the previous state of the art~\cite{packnet}: from $0.162$ to $0.139$ absolute relative error (Abs Rel) on the front camera. By introducing spatial contexts (Equation \ref{eq:warp_spatial}), we not only further improve performance, but also learn \emph{scale-aware} models by leveraging the camera extrinsics.  Note that there is no limitation on the extrinsics transformation, only the assumption of some overlap between camera pairs (Figure \ref{fig:ddad_warps}).  Finally, by introducing our proposed spatio-temporal contexts (Equation \ref{eq:warp_multi}) and pose consistency constraints (Section \ref{sec:pcc}) we further boost performance to $0.130$, thus \textbf{achieving a new state of the art by a large margin}.  
\input{tables/both_scaling_small2}

\input{figures/ddad_qualitative}

\input{figures/nuscenes_qualitative}

\input{figures/camviz}

\subsection{Multi-Camera Depth Estimation}

We now evaluate FSM depth performance on all cameras of the \emph{DDAD} and \emph{nuScenes} datasets, ablating the impact of our contributions in the multi-camera setting (see Table ~\ref{tab:both_scaling_small}). 

\subsubsection{Photometric Masking}

As a baseline, we combine images from all cameras into a single dataset, without considering masking or cross-camera constraints.  This is similar to Gordon \etal~\cite{gordon2019depth}, where multiple datasets from different cameras are pooled together to train a single model.  As discussed in Section \ref{sec:self-occ}, the presence of self-occlusions on the DDAD dataset severely degrades depth performance when masking is not considered, reaching $0.380$ Abs Rel (average of all cameras) versus $0.211$ when self-occlusion masks are introduced (see Figure \ref{fig:self_occ_masks} for a qualitative comparison).  Note that these results are still unscaled, and therefore median-scaling is required at test time for a proper quantitative evaluation.
\subsubsection{Spatio-Temporal Contexts}

The introduction of our proposed spatio-temporal contexts (STC), as described in Section \ref{sec:stc}, boosts performance on all cameras, from $0.211$ to $0.202$ ($4.5\%$) on \emph{DDAD} and $0.327$ to $0.299$ ($9.1\%$) on \emph{nuScenes}, by leveraging different levels of overlapping between views. This improvement becomes more apparent when considering our \emph{shared median-scaling} evaluation protocol: $0.241$ to $0.208$ ($16.1\%$) on \emph{DDAD} and $0.428$ to $0.355$ ($20.6\%$) on \emph{nuScenes}. This is evidence that STC produces more consistent pointclouds across multiple cameras, as evidenced in Figure \ref{fig:pcl_align} and revealed by our proposed metric.
Furthermore, the known extrinsics between cameras enables the generation of \emph{scale-aware} models, with minimal degradation from their median-scaled counterparts: $0.207$ versus $0.202$ ($2.2\%$) on \emph{DDAD} and $0.311$ vs $0.299$ ($4.0\%$ on \emph{nuScenes}). In this setting, we also evaluated the impact of STC relative to using only spatial and temporal contexts independently. As expected, there is a noticeable degradation in performance when spatio-temporal contexts are not considered: $0.206$ to $0.218$ ($5.4\%$) on DDAD and $0.311$ to $0.391$ ($25.8\%$) on nuScenes. We hypothesize that degradation on \emph{nuScenes} is substantially higher due to an overall smaller overlapping between cameras (Figure \ref{fig:camviz}), which will benefit more from STC as a way to improve cross-camera photometric losses for self-supervised training (Figure \ref{fig:ddad_warps}).

\subsubsection{Pose Consistency Constraints}

Finally, we include the pose consistency constraints (PCC) described in Section \ref{sec:pcc}, as a way to enforce the learning of a similar rigid motion for all cameras. These additional constraints further boost performance, from $0.207$ to $0.201$ ($2.9\%$) on \emph{DDAD} and $0.311$ to $0.297$ ($4.5\%$) on \emph{nuScenes}. These improvements are more prominent on the side cameras, since per-frame pose estimation is harder in these cases due to larger relative motion and smaller overlap. Interestingly, the combination of all our contributions (masking, spatio-temporal contexts and pose consistency constraints) lead to \textbf{scale-aware results that surpass median-scaled results}. This is evidence that FSM is capable of generating state-of-the-art metrically accurate models that are useful for downstream tasks in real applications.

%% file: figures/pcl_align.tex
\begin{figure}[t!]
    \centering
    \subfloat[Monocular photometric loss (Abs Rel 0.211/0.241)]{
    \includegraphics[width=0.93\linewidth,height=2.5cm, trim=0 50 0 50, clip]{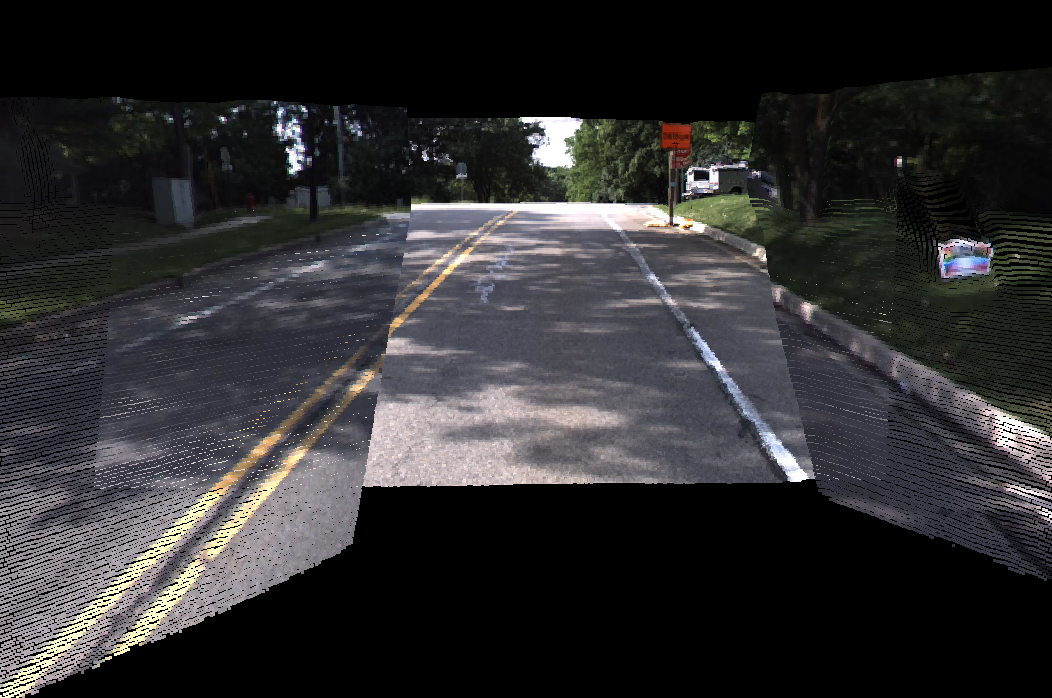}
    }
\\
    \vspace{-3mm}
    \subfloat[Spatio-temporal photometric loss (Abs Rel 0.201/0.207)]{
    \includegraphics[width=0.93\linewidth,height=2.5cm, trim=0 50 0 50, clip]{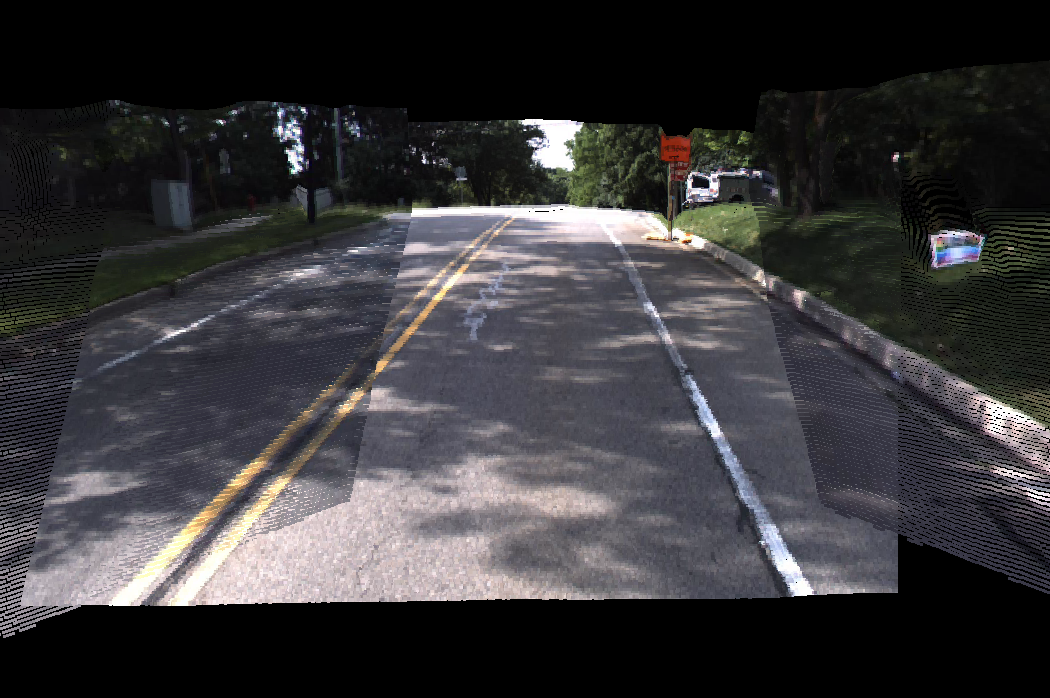}
    }
    \vspace{-1mm}
    \caption{\textbf{Multi-camera pointcloud alignment} on DDAD using (a) the standard monocular photometric loss, and (b) our proposed spatio-temporal photometric constraints. We also report per-frame and shared median-scaling Abs Rel results (average of all cameras, see Table~\ref{tab:both_scaling_small} for more details).}
    \vspace{-2mm}
    \label{fig:pcl_align}
\vspace{-3mm}
\end{figure}

%% file: tables/table_stereo.tex
\captionsetup[table]{skip=6pt}

\begin{table}[b!]
\vspace{-4mm}
\renewcommand{\arraystretch}{0.9}
\centering
{
\small
\setlength{\tabcolsep}{0.3em}
\begin{tabular}{l|c|cccc}
\toprule
\textbf{Method} & \textbf{Sup.} & 
Abs Rel$\downarrow$ &
Sq Rel$\downarrow$ &
RMSE$\downarrow$ &
$\delta_{1.25}$ $\uparrow$
\\
\toprule
UnDeepVO~\cite{li2018undeepvo}
& $S$ & 0.183 & 1.730 & 6.570 & - \\
Godard et al.~\cite{godard2017unsupervised}
& $S$ & 0.148 & 1.344 & 5.927 & 0.803 \\
SuperDepth~\cite{superdepth} 
& $S$ & 0.112 & 0.875 & 4.958 & 0.852 \\
Monodepth2~\cite{godard2019digging} 
& $M$ & 0.115 & 0.903 & 4.863 & 0.877 \\
Monodepth2~\cite{godard2019digging} 
& $S$ & 0.109 & 0.873 & 4.960 & 0.864 \\
\midrule
Monodepth2~\cite{godard2019digging} 
& $M+S$ & \textbf{0.106} & \underline{0.818} & \underline{4.750} & \textbf{0.874} \\
\textbf{FSM} & $M+S$ & \underline{0.108} & \textbf{0.737} & \textbf{4.615} & \underline{0.872} \\
\bottomrule
\end{tabular}
}
\caption{
\textbf{Depth estimation results on the KITTI dataset,} relative to stereo methods. Even though our approach relaxes several stereo assumptions, it remains competitive with published methods. 
}

\label{tab:stereo}
\vspace{-1mm}
\end{table}

%% file: tables/ddad_depth.tex
\captionsetup[table]{skip=6pt}

\begin{table}[t!]
\renewcommand{\arraystretch}{0.9}
\centering
{
\small
\setlength{\tabcolsep}{0.3em}
\begin{tabular}{l|cccc}
\toprule
\textbf{Method}  & 
Abs Rel$\downarrow$ &
Sq Rel$\downarrow$ &
RMSE$\downarrow$ &
$\delta_{1.25}$ $\uparrow$
\\
\toprule
COLMAP (pseudo-depth) & 
0.243 & 4.438 & 17.239 & 0.601 \\
\midrule
Monodepth2 (R18)~\cite{godard2019digging}  &
0.213 & 4.975 & 18.051 & 0.761 \\
Monodepth2 (R50)~\cite{godard2019digging} &
0.198 & 4.504 & 16.641 & 0.781 \\
PackNet~\cite{packnet}  &
0.162 & 3.917 & \textbf{13.452} & 0.823 \\
\midrule
\textbf{FSM} (w/o mask \& spatial) & 0.184 & 4.049 & 17.109 & 0.735 \\
\textbf{FSM} (w/o spatial) & 0.139 & 3.023 & 14.106 & 0.827 \\
\textbf{FSM}$^\star$ (w/ spatial) & 0.135 & 2.841 & 13.850 & 0.832 \\
\textbf{FSM}$^\star$ (w/ spatio-temporal) & \textbf{0.130} & \textbf{2.731} & 13.654 & \textbf{0.837} \\

\bottomrule
\end{tabular}
}
\caption{
\textbf{Quantitative depth evaluation of different methods on the DDAD \cite{packnet} dataset}, for distances up to 200m on the forward-facing camera. The symbol $^\star$ indicates a \emph{scale-aware} model, evaluated without median-scaling at test time.
}
\vspace{-3mm}
\label{table:ddad_depth}
\end{table}

%% file: tables/both_scaling_small2.tex
\begin{table}%
\centering
\small
\renewcommand{\arraystretch}{1.05}
\setlength{\tabcolsep}{0.22em}
\subfloat[DDAD]{
\begin{tabular}{l|cccccc|c}
    \toprule
    \multirow{2}{*}{\textbf{Method}} &
    \multicolumn{7}{c}{Abs.Rel.$\downarrow$} \\
    \cmidrule{2-8}
    & \textit{Front}
    & \textit{F.Left}
    & \textit{F.Right}
    & \textit{B.Left}
    & \textit{B.Right}
    & \textit{Back}
    & \textit{Avg.}
    \\
    \midrule
    Mono$^\dagger$ - M
    & 0.184
    & 0.366
    & 0.448
    & 0.417
    & 0.426
    & 0.438
    & 0.380
    \\
    Mono$^\dagger$
    & 0.139
    & 0.209
    & 0.236
    & 0.231
    & 0.247
    & 0.204
    & 0.211
    \\
    \textbf{FSM$^\dagger$}
    & \underline{0.131}
    & \underline{0.203}
    & \underline{0.226}
    & \textbf{0.223}
    & \textbf{0.240}
    & \underline{0.188}
    & \underline{0.202}
    \\
    \midrule
    Mono$^\ddagger$
    & 0.143
    & 0.238
    & 0.265
    & 0.277
    & 0.276
    & 0.247
    & 0.241
    \\
    \textbf{FSM$^\ddagger$}
    & 0.133
    & 0.212
    & 0.229
    & 0.231
    & \underline{0.246}
    & 0.194
    & 0.208
    \\ 
    \midrule
    \textbf{FSM - STC}
    & 0.133
    & 0.219
    & 0.246
    & 0.252
    & 0.259
    & 0.197
    & 0.218
    \\
    \textbf{FSM - PCC}
    & \underline{0.131}
    & 0.206
    & 0.228
    & 0.238 
    & 0.248
    & \underline{0.188}
    & 0.207
    \\
    \textbf{FSM}
    & \textbf{0.130}
    & \textbf{0.201}
    & \textbf{0.224}
    & \underline{0.229}
    & \textbf{0.240}
    & \textbf{0.186}
    & \textbf{0.201}
    \\
    \bottomrule
\end{tabular}
}
\\
\vspace{-2mm}
\subfloat[nuScenes]{
\begin{tabular}{l|cccccc|c}
    \toprule
    \multirow{2}{*}{\textbf{Method}} &
    \multicolumn{7}{c}{Abs.Rel.$\downarrow$} \\
    \cmidrule{2-8}
    & \textit{Front}
    & \textit{F.Left}
    & \textit{F.Right}
    & \textit{B.Left}
    & \textit{B.Right}
    & \textit{Back}
    & \textit{Avg.}
    \\
    \midrule
    Mono$^\dagger$
    & 0.214
    & 0.304
    & 0.388
    & 0.314 
    & 0.438 
    & 0.304 
    & 0.327 
    \\
    \textbf{FSM$^\dagger$}
    & 0.198
    & 0.297
    & \textbf{0.364}
    & \underline{0.301} 
    & \textbf{0.392}
    & \underline{0.240}
    & \underline{0.299}
    \\
    \midrule
    Mono$^\ddagger$
    & 0.251
    & 0.403 
    & 0.546
    & 0.429 
    & 0.616 
    & 0.321 
    & 0.428
    \\
    \textbf{FSM$^\ddagger$}
    & 0.200 
    & 0.337 
    & 0.448
    & 0.354 
    & 0.521 
    & 0.267 
    & 0.355 
    \\ 
    \midrule
    \textbf{FSM - STC}
    & 0.208
    & 0.382
    & 0.510
    & 0.393
    & 0.595
    & 0.258
    & 0.391
    \\ 
    \textbf{FSM - PCC}
    & \underline{0.187} 
    & \underline{0.291}
    & 0.392
    & 0.311
    & 0.448
    & 0.235
    & 0.311
    \\ 
    \textbf{FSM}
    & \textbf{0.186}
    & \textbf{0.287}
    & \underline{0.375}
    & \textbf{0.296}
    & \underline{0.418}
    & \textbf{0.221}
    & \textbf{0.297}
    \\
    \bottomrule
\end{tabular}
}
\vspace{-2mm}
\caption{\textbf{Depth estimation results on multi-camera datasets}, using \emph{FSM} relative to the single-camera photometric loss (\emph{Mono}). The symbol $^\dagger$ denotes per-frame median-scaling, and $^\ddagger$ shared median-scaling (Eq.~\ref{eq:sharedmedscaling}). \emph{M} denotes the removal of masking (Sec.~\ref{sec:mask}), \emph{STC} the removal of spatio-temporal contexts (Sec.~\ref{sec:stc}), and \emph{PCC} the removal of pose consistency constraints (Sec.~\ref{sec:pcc}).}
\label{tab:both_scaling_small}
\vspace{-3mm}
\end{table}

%% file: figures/ddad_qualitative.tex
\begin{figure*}[t!]
\vspace{-2mm}
\centering
\subfloat{
\includegraphics[width=0.16\linewidth,height=1.5cm]{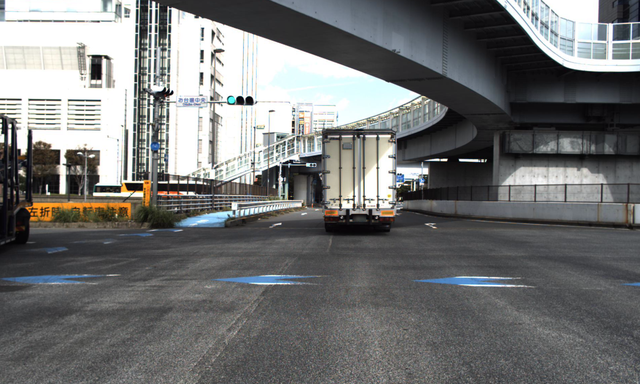}
\includegraphics[width=0.16\linewidth,height=1.5cm]{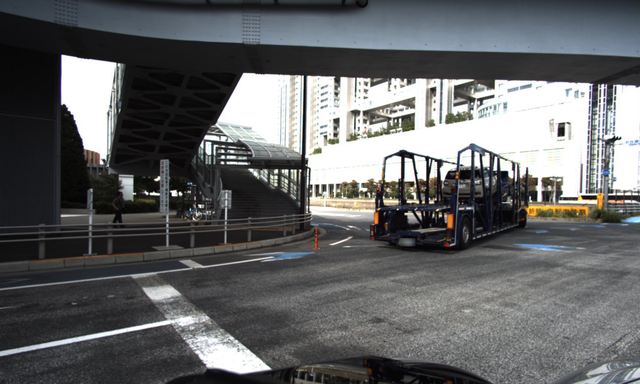}
\includegraphics[width=0.16\linewidth,height=1.5cm]{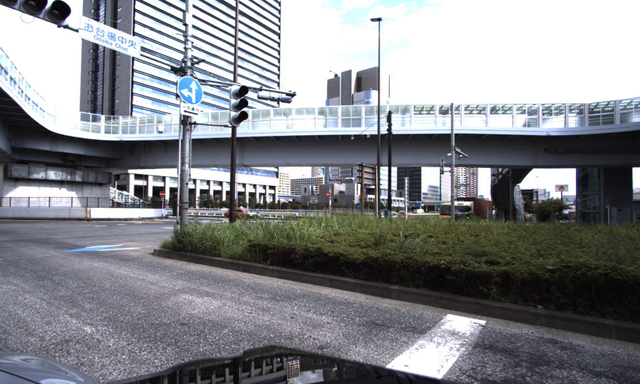}
\includegraphics[width=0.16\linewidth,height=1.5cm]{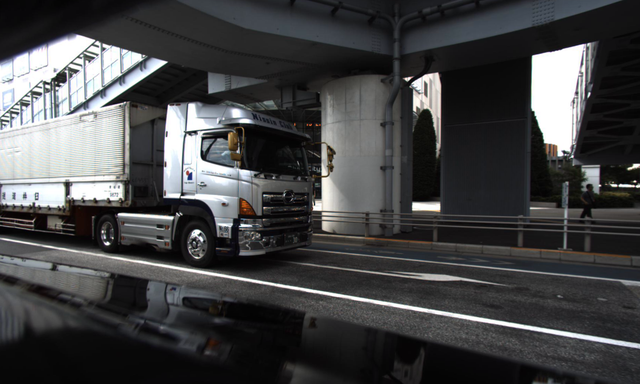}
\includegraphics[width=0.16\linewidth,height=1.5cm]{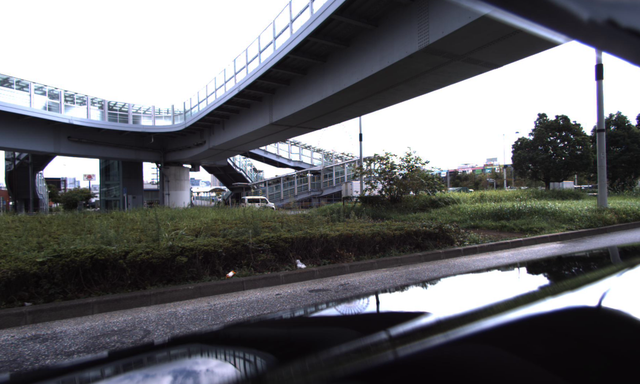}
\includegraphics[width=0.16\linewidth,height=1.5cm]{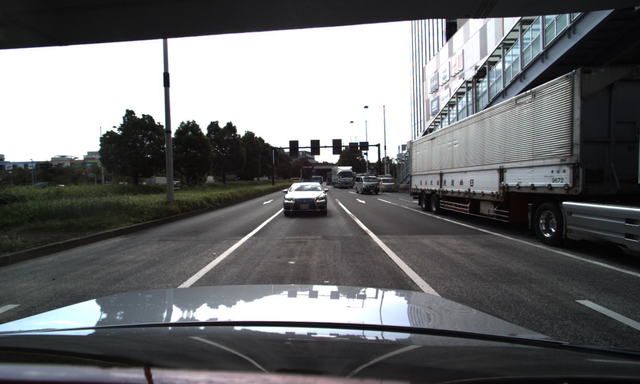}
}
\vspace{-4mm}
\\
\subfloat{
\includegraphics[width=0.16\linewidth,height=1.5cm,height=1.5cm]{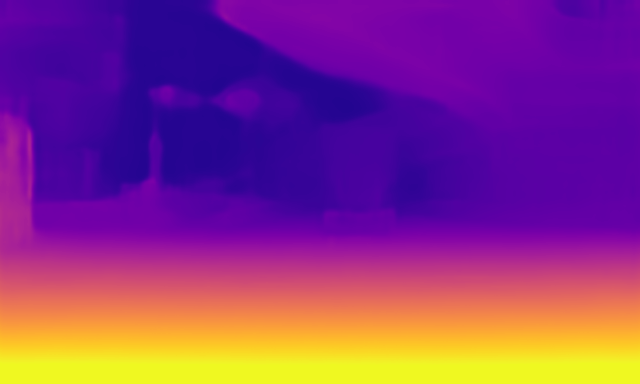}
\includegraphics[width=0.16\linewidth,height=1.5cm,height=1.5cm]{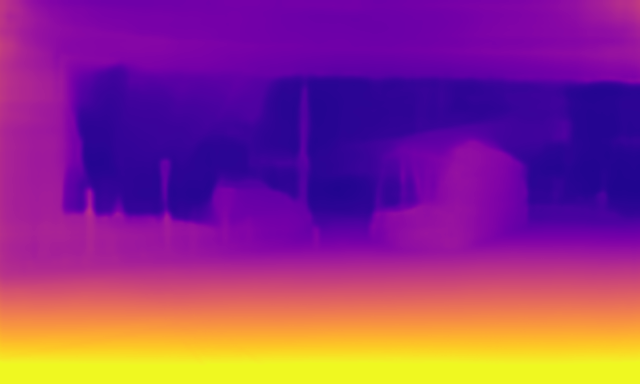}
\includegraphics[width=0.16\linewidth,height=1.5cm,height=1.5cm]{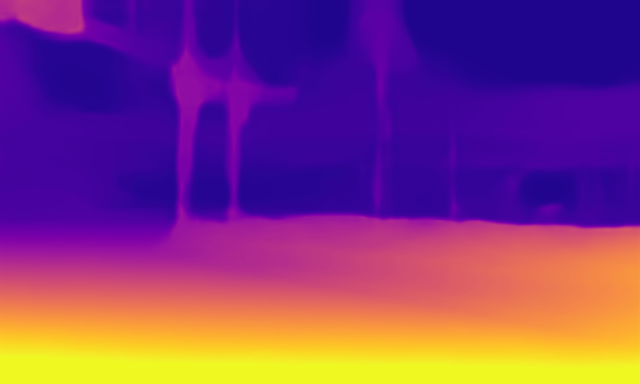}
\includegraphics[width=0.16\linewidth,height=1.5cm,height=1.5cm]{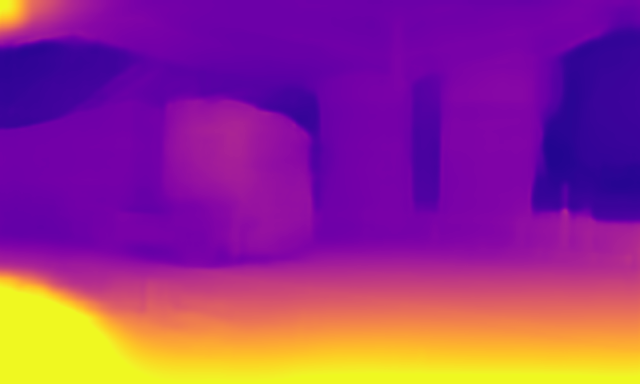}
\includegraphics[width=0.16\linewidth,height=1.5cm,height=1.5cm]{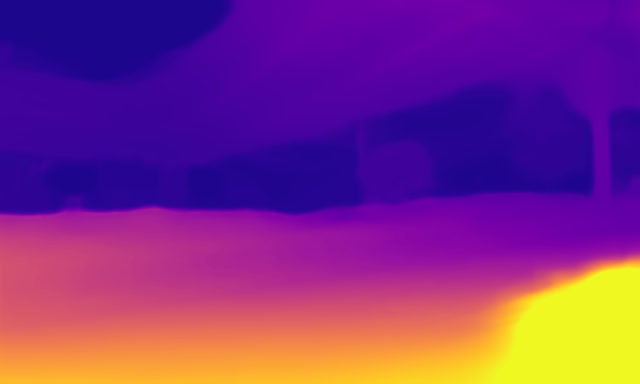}
\includegraphics[width=0.16\linewidth,height=1.5cm,height=1.5cm]{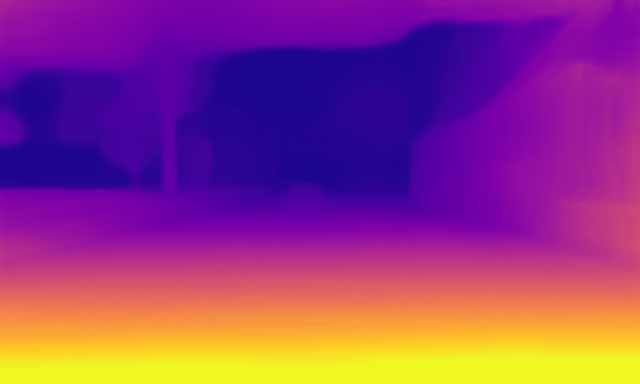}
}
\\
\caption{
\textbf{Self-Supervised depth estimation results} using FSM on the \textit{DDAD} dataset.
}
\label{fig:ddad_qualitative}
\end{figure*}

%% file: figures/nuscenes_qualitative.tex
\begin{figure*}[t!]
\vspace{-2mm}
\centering
\subfloat{
\includegraphics[width=0.16\linewidth, height=1.5cm]{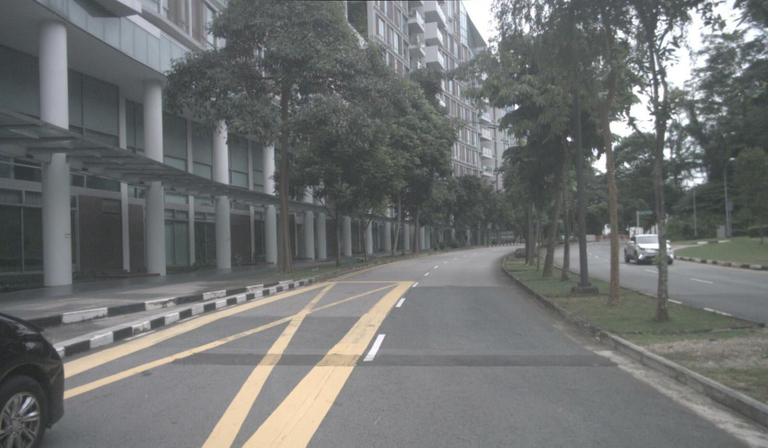}
\includegraphics[width=0.16\linewidth, height=1.5cm]{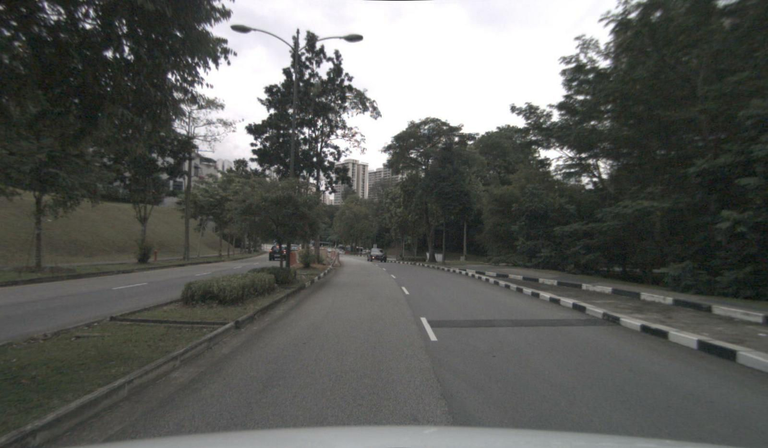}
\includegraphics[width=0.16\linewidth, height=1.5cm]{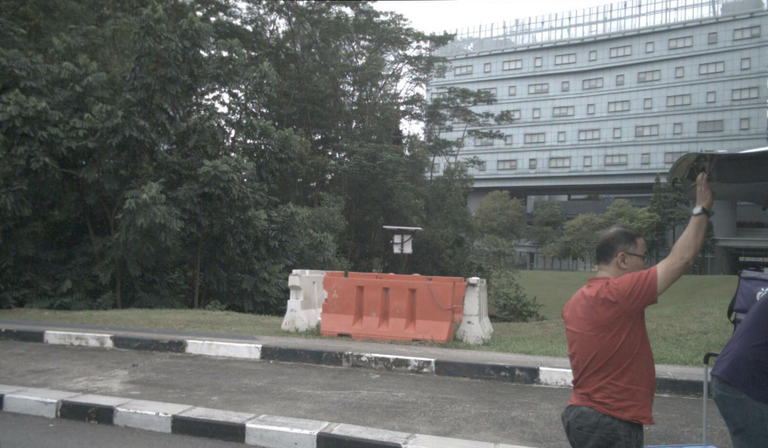}
\includegraphics[width=0.16\linewidth, height=1.5cm]{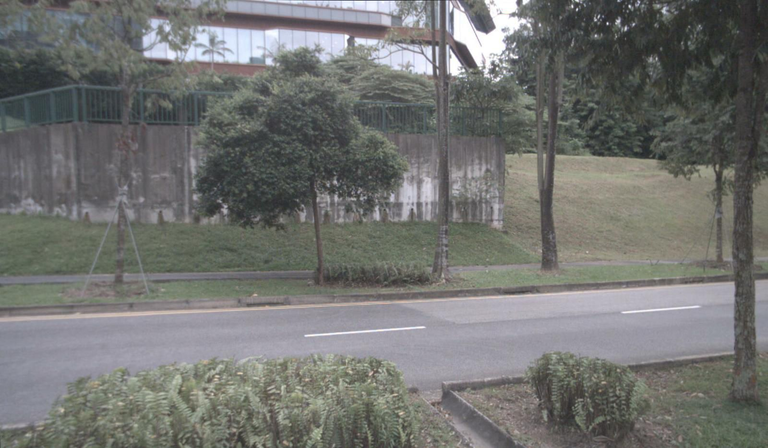}
\includegraphics[width=0.16\linewidth, height=1.5cm]{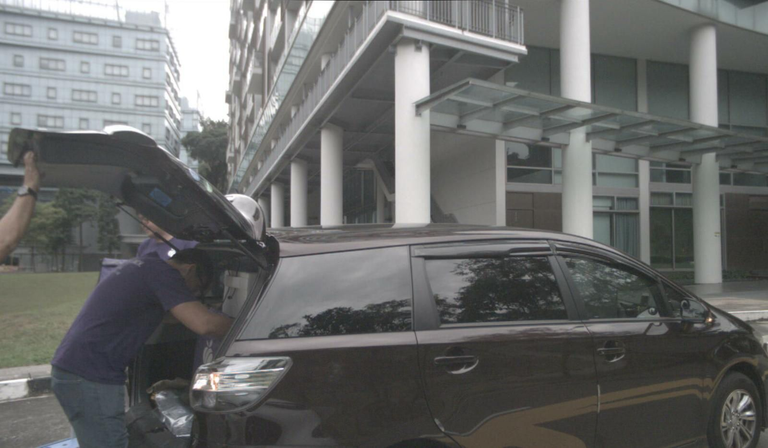}
\includegraphics[width=0.16\linewidth, height=1.5cm]{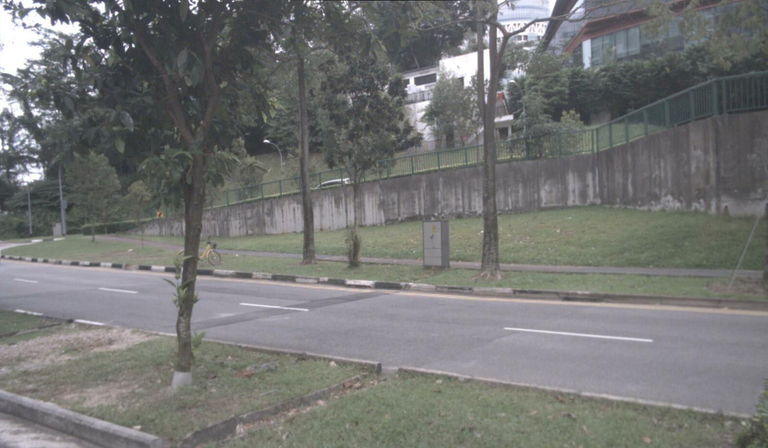}
}
\vspace{-4mm}
\\
\subfloat{
\includegraphics[width=0.16\linewidth, height=1.5cm]{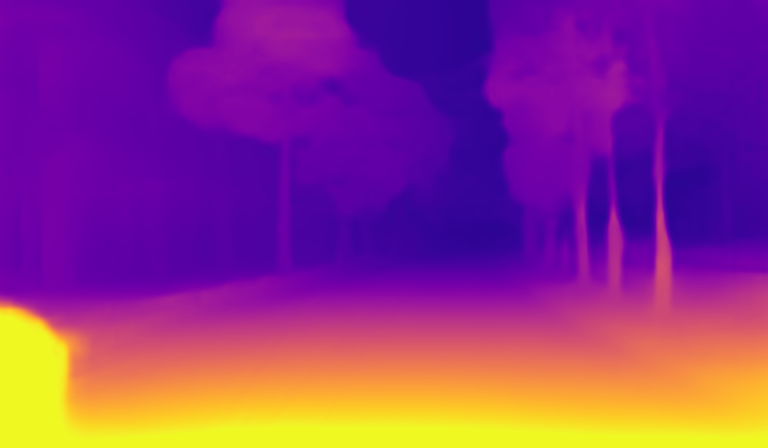}
\includegraphics[width=0.16\linewidth, height=1.5cm]{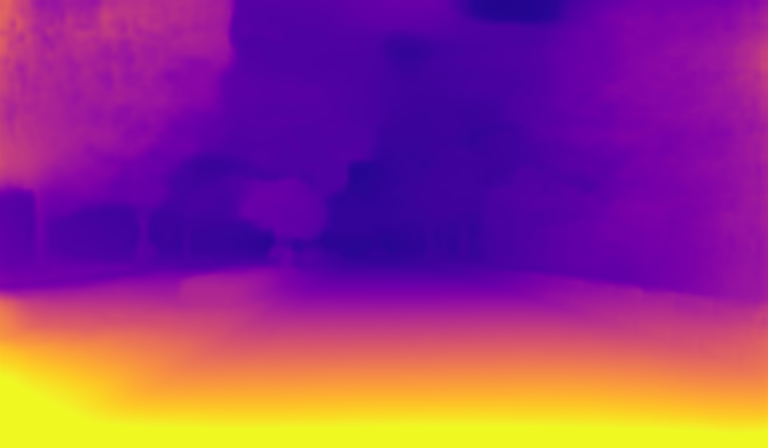}
\includegraphics[width=0.16\linewidth, height=1.5cm]{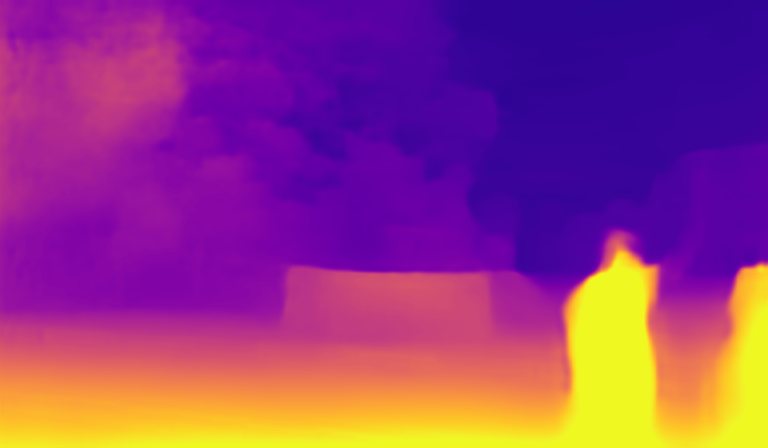}
\includegraphics[width=0.16\linewidth, height=1.5cm]{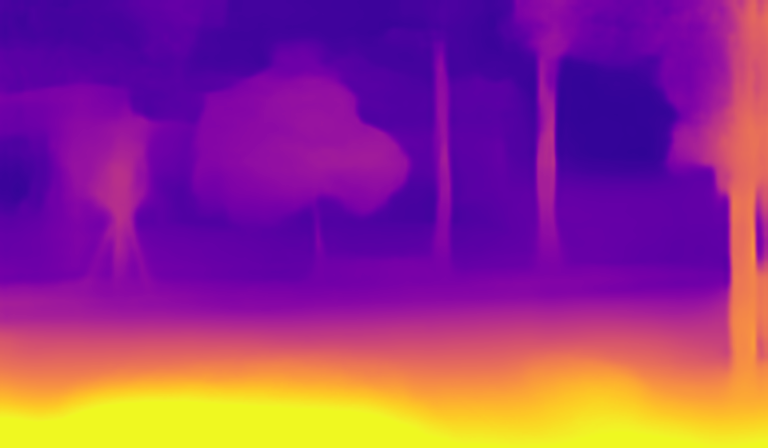}
\includegraphics[width=0.16\linewidth, height=1.5cm]{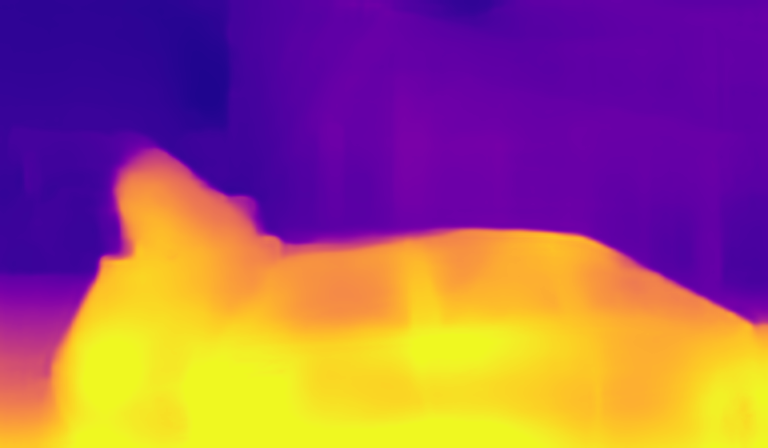}
\includegraphics[width=0.16\linewidth, height=1.5cm]{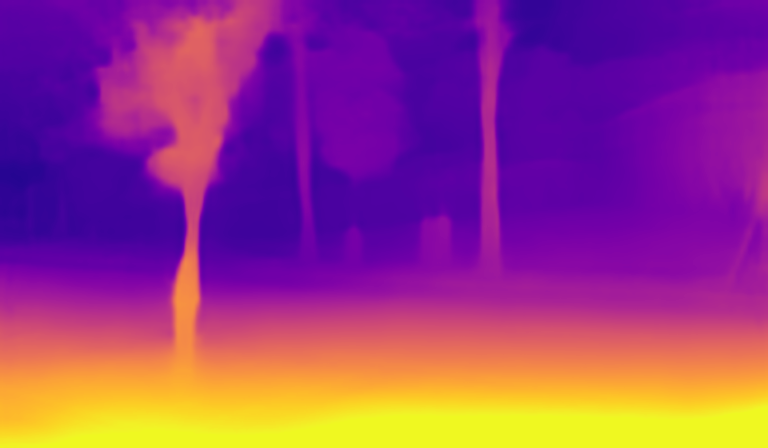}
}
\\
\caption{
\textbf{Self-Supervised depth estimation results} using FSM on the \textit{nuScenes} dataset.
}
\label{fig:nuscenes_qualitative}
\end{figure*}

%% file: figures/camviz.tex
\begin{figure*}[t!]
    \vspace{-4mm}
    \centering
    \subfloat[DDAD]{
    \includegraphics[width=0.49\linewidth,height=3cm]{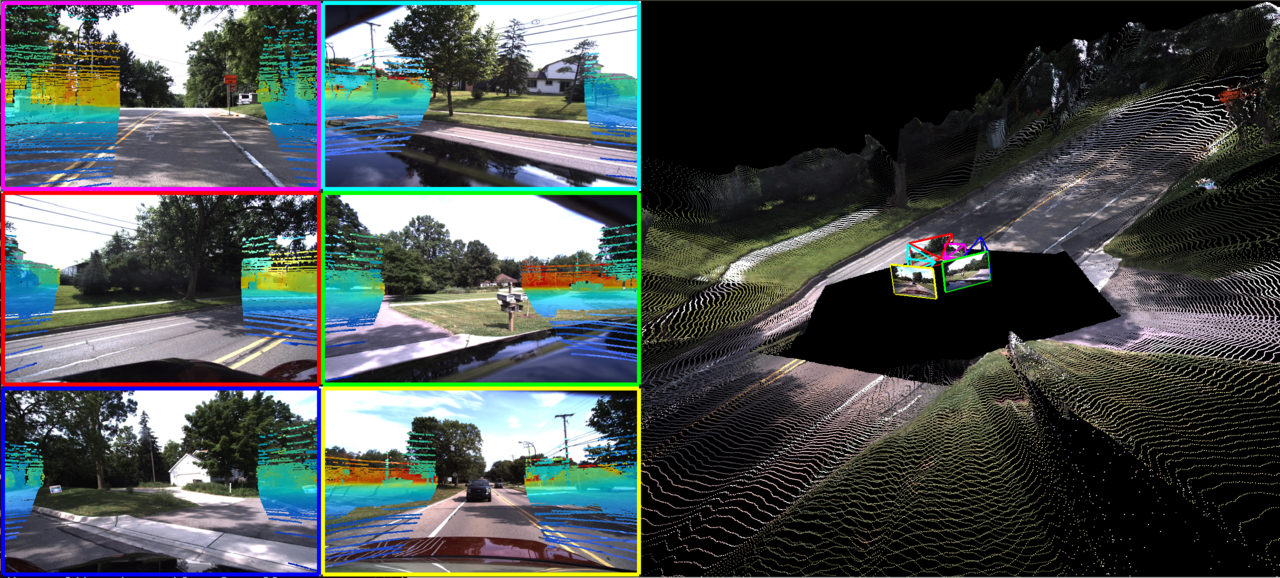}
    }
    \subfloat[NuScenes]{
    \includegraphics[width=0.49\linewidth,height=3cm]{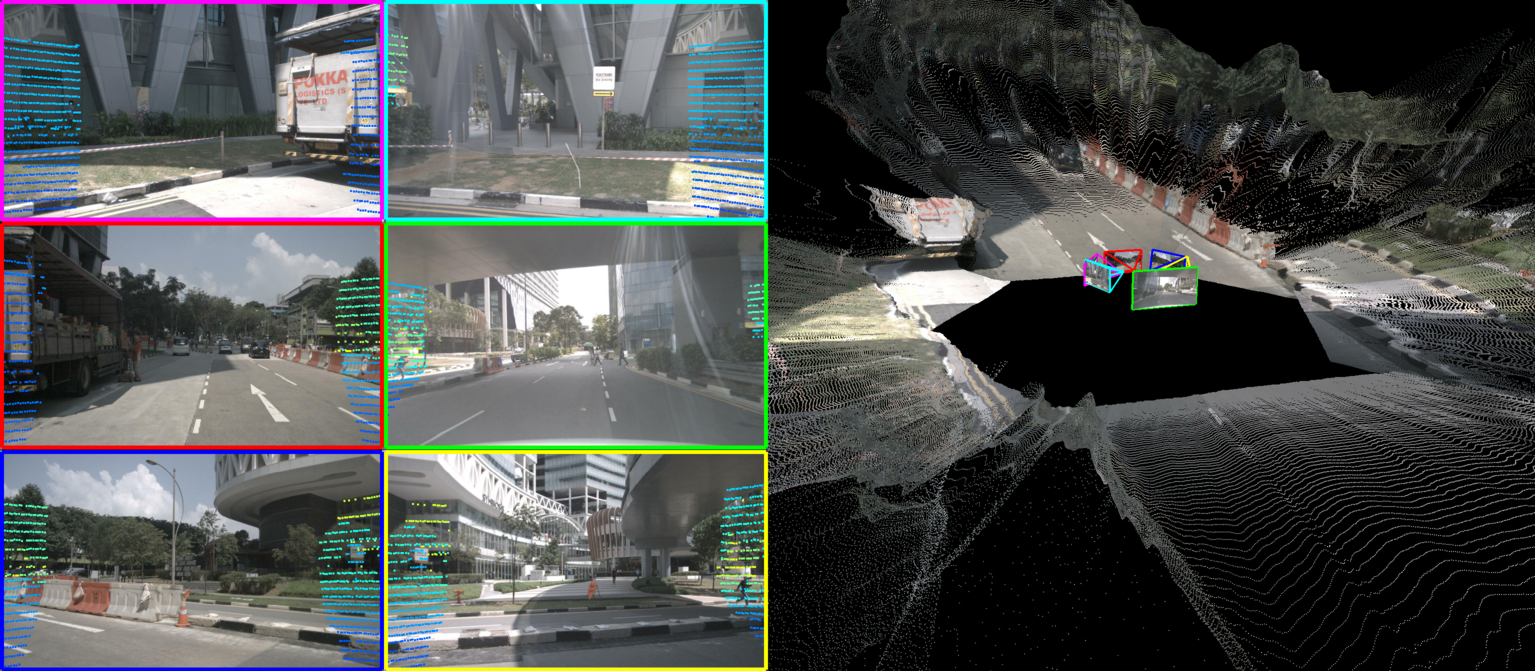}
    }
    \vspace{-2mm}
    \caption{\textbf{Predicted pointclouds using FSM on the \textit{DDAD} and \textit{nuScenes} datasets}. For each dataset, the same network is used in all six images, predicted depth maps are lifted to 3D using camera intrinsics and extrinsics, and then combined \emph{without any post-processing}. As a way to visualize camera overlapping, we also show projected LiDAR points from adjacent views overlaid on each RGB image (this information is not used at training or test time).}
    \label{fig:camviz}
    \vspace{-4mm}
\end{figure*}

%% file: sections/conclusion.tex
Using cameras for 3D perception to complement or replace LiDAR scanners is an exciting new frontier for robotics.  
We have extended self-supervised learning of depth and ego-motion from monocular and stereo settings to the general multi-camera setting, predicting \emph{scale-aware} and dense point clouds around the ego-vehicle. 
We also introduced a series of key techniques that boost performance in this new setting by leveraging known extrinsics between cameras: \emph{spatial-temporal contexts}, \emph{pose consistency constraints}, and studied the effects of \emph{non-overlapping} and \emph{self-occlusion} photometric masking. In extensive experiments we demonstrated the capabilities of our methods and how they advance the state of the art. 
As future work, we plan to relax the assumption of known intrinsics and extrinsics, and estimate these parameters jointly with depth and ego-motion to enable vehicle self-calibration.

%% file: main_suppmat.tex
\appendix

\section{Network Architectures}

For our experiments we use a ResNet18-based depth and pose architecture similar to~\cite{godard2019digging}. For more details please refer to Table~\ref{tab:networks}. Note that our FSM constraints do not require any particular depth and pose network architectures.

\section{Datasets}

\subsection{DDAD}

The Dense Depth for Automated Driving (DDAD)~\cite{packnet} is an urban driving dataset captured with six synchronized cameras and depth ranges of up to 250 meters. It has a total of 12,650 training samples, from which we consider all six cameras for a total of 63,250 images and ground-truth depth maps (not used in this work). The validation set contains 3,950 samples (15,800 images) and ground-truth depth maps, used only for evaluation. Following the procedure outlined in~\cite{packnet}, input images were downsampled to a $640 \times 384$ resolution, and for evaluation we considered distances up to 200m without any cropping.

\input{tables/suppmat_networks}

\input{supp_mat/ddad_qualitative}

\input{supp_mat/nuscenes_qualitative}

\subsection{nuScenes}

The nuScenes~\cite{caesar2020nuscenes} dataset is an urban driving dataset that contains images from a synchronized six-camera array, comprised of $1000$ scenes with a total of $1.4$M images.  The dataset contains 2D and 3D annotations, and is primarily used as a detection dataset.  It is annotated at a 2Hz framerate, however images are captured at 30Hz, and we use the larger dataset for training.  The raw images are $1600 \times 900$, which are downsampled to $768 \times 448$, and evaluated at distances up to 80m without any cropping. Though the data is diverse, it is also very challenging for self-supervised monocular depth estimation, containing many scenes in very low illumination, with rain droplets occluding the lens, and where the road is not visible from the side cameras (see Figure~\ref{fig:nuscenes_qualitative_supp}).

\input{supp_mat/overlapping}

\subsection{KITTI}
For our stereo KITTI experiments we train and evaluate on the standard \textit{Eigen} split \cite{zhou2017unsupervised}, which is comprised of $23,488$ training, $888$ validation and $697$ testing images. Corresponding projected depth maps for ground truth evaluation are obtained from raw LiDAR scans. Images were downsampled to $640 \times 192$, and evaluated at distances up to 80m with the \emph{garg} crop \cite{zhou2017unsupervised}.

\section{COLMAP}
COLMAP~\cite{schonberger2016structure} is a leading SfM technique used for large-scale 3D reconstruction. It takes a collection of images and produces a 3D pointcloud, from which dense depth maps can be obtained by projecting this reconstructed pointcloud onto each image plane. 
We use COLMAP as a ``self-supervised'' baseline to our method, generating pseudo-ground truth projected depth maps to train a supervised depth estimator. 
We used the following procedure to generate pseudo-ground truth COLMAP depth maps on the \emph{DDAD} dataset: first, original $1936\times 1216$ images are downsampled by $50\%$ to $968 \times 608$ to more closely match the resolution of self-supervised learning, and to make time-intensive dataset generation more manageable. Then, for each scene we consider the set of images from all six cameras and all timesteps and generate a COLMAP pointcloud (using the default parameters, which we have found to produce best overall results on average between diverse scene types). Finally, this 3D structure is reprojected as depth maps onto each image.  For some scenes, COLMAP failed to reconstruct a consistent pointcloud, leading to depth map generation only for a subset of images.
Note that this procedure is very time-consuming -- each \emph{DDAD} scene can take $3-5$ GPU-hours for 3D reconstruction (thus $600-1000$ GPU-hours for the entire dataset). In comparison, our approach to self-supervision trains the network to convergence over the entire dataset in a few hours. 
Using this procedure, we were able to generate pseudo-ground truth depth for approximately $80\%$ of the \emph{DDAD} dataset.  For the COLMAP-supervised training, we used the same depth network architecture shown in Table~\ref{tab:netdepth} and no pose network, since there are no temporal contexts.


\section{Training Details}

Our models were implemented using PyTorch~\cite{paszke2017automatic} and trained across eight V100 GPUs\footnote{Training and inference code, as well as pre-trained models, will be made available upon publication.}. To highlight the flexibility of our proposed framework, all experiments used the same training hyper-parameters: Adam optimizer~\cite{kingma2014adam}, with $\beta_1=0.9$ and $\beta_2=0.999$; batch size of $4$ per GPU for single camera and $6$ per GPU for multi-camera (all images from each sample in the same batch); learning rate of $2 \cdot 10^{-4}$ for $20$ epochs; the previous $t-1$ and subsequent $t+1$ images are used as temporal context; color jittering and horizontal flipping as data augmentation; SSIM weight of $\alpha=0.85$; and depth smoothness weight of $\lambda_d=0.001$. We also used coefficients $\lambda_s=0.1$ and $\lambda_t=1.0$ to weight spatial and temporal losses respectively.
    

\section{Multi-Camera Qualitative Results}
We show qualitative results for randomly-selected images from the \emph{DDAD} dataset in Figure~\ref{fig:ddad_qualitative_supp}, and \emph{nuScenes} results in Figure~\ref{fig:nuscenes_qualitative_supp}. The difference in data quality between these two datasets (image resolution, camera field-of-view overlap) is reflected in the quantitative results reported in the main paper. Even so, our proposed FSM constraints significantly boost performance in these two challenging settings to achieve a new state of the art in self-supervised monocular depth estimation, without any changes to the underlying framework. 


Similarly, in Figure \ref{fig:pcl360_overlapping} we show examples of reconstructed $360\degree$ pointclouds for each dataset, obtained by lifting 2D color information to 3D using predicted depth and and camera calibration (intrinsics and extrinsics). We also show how much overlap there is between cameras by overlaying projected LiDAR points from adjacent views, where we can see that \emph{DDAD} has significantly more overlap that \emph{nuScenes}. Regardless, FSM is capable of leveraging different levels of overlap between cameras to generate consistent, scale aware pointclouds without any post-processing. A video showing examples of reconstructed sequences using FSM is also provided as supplementary material. 

\paragraph{Acknowledgments} IV and GS were supported in part by AFOSR Center of Excellence Award FA9550-18-1-0166.

%% file: tables/suppmat_networks.tex
\begin{table}[t!]%
\small
  \centering
\resizebox{0.84\linewidth}{!}{
\subfloat[][Depth Network \cite{monodepth2}.]{
\begin{tabular}[b]{l|l|c|c|c}
\toprule
& \textbf{Layer Description} & \textbf{K} & \textbf{S} & \textbf{Out. Dim.} \\ 
\toprule
\multicolumn{5}{c}{\textbf{ResidualBlock (K, S)}} \\ 
\midrule
\#A & Conv2d $\shortrightarrow$ BN $\shortrightarrow$ ReLU & K & 1 &  \\
\#B & Conv2d $\shortrightarrow$ BN $\shortrightarrow$ ReLU & K & S &  \\
\toprule
\multicolumn{5}{c}{\textbf{UpsampleBlock (\#skip)}} \\ 
\midrule
\#C & Conv2d $\shortrightarrow$ BN $\shortrightarrow$ ReLU $\shortrightarrow$ Upsample         & 3 & 1 & \\
\#D & Conv2d ($\#C \oplus \#skip$) $\shortrightarrow$ BN $\shortrightarrow$ ReLU  & 3 & 1 & \\
\toprule
\toprule
\#0 & Input RGB image & - & - & 3$\times$H$\times$W \\ 
\midrule
\multicolumn{5}{c}{\textbf{Encoder}} \\ \hline
\#1  & Conv2d $\shortrightarrow$ BN $\shortrightarrow$ ReLU   & 7 & 1 &  64$\times$H$\times$W \\
\#2  & Max. Pooling                 & 3 & 2 &  64$\times$H/2$\times$W/2 \\
\#3  & ResidualBlock (x2)           & 3 & 2 &  64$\times$H/4$\times$W/4 \\
\#4  & ResidualBlock (x2)           & 3 & 2 & 128$\times$H/8$\times$W/8 \\
\#5  & ResidualBlock (x2)           & 3 & 2 & 256$\times$H/16$\times$W/16 \\
\#6  & ResidualBlock (x2)           & 3 & 2 & 512$\times$H/32$\times$W/32 \\
\midrule
\multicolumn{5}{c}{\textbf{Depth Decoder}} \\ 
\midrule
\#7 & UpsampleBlock (\#5)    & 3 & 1 & 256$\times$H/16$\times$W/16 \\
\#8 & UpsampleBlock (\#4)    & 3 & 1 & 128$\times$H/8$\times$W/8 \\
\#9 & UpsampleBlock (\#3)    & 3 & 1 & 64$\times$H/4$\times$W/4 \\
\#10 & UpsampleBlock (\#2)   & 3 & 1 & 32$\times$H/2$\times$W/2 \\
\#11 & UpsampleBlock (\#1)   & 3 & 1 & 32$\times$H$\times$W \\
\#12 & Conv2d $\shortrightarrow$ Sigmoid  & 3 & 1 & 1$\times$H$\times$W \\
\bottomrule
\end{tabular}
\label{tab:netdepth}
}}
\\
\resizebox{0.84\linewidth}{!}{
\subfloat[][Pose Network \cite{zhou2018unsupervised}.]{
\begin{tabular}[b]{l|c|c|c|c}
\toprule
& \textbf{Layer Description} & \textbf{K} & \textbf{S} & \textbf{Out. Dim.} \\ 
\toprule
\#0 & Input 2 RGB images & - & - & 6$\times$H$\times$W \\ 
\midrule
\#1  & \hspace{2mm} Conv2d $\shortrightarrow$ GN $\shortrightarrow$ ReLU \hspace{2mm} & 3 & 2 & 16$\times$H/2$\times$W/2 \\
\#2  & \hspace{2mm} Conv2d $\shortrightarrow$ GN $\shortrightarrow$ ReLU \hspace{2mm} & 3 & 2 & 32$\times$H/4$\times$W/4 \\
\#3  & \hspace{2mm} Conv2d $\shortrightarrow$ GN $\shortrightarrow$ ReLU \hspace{2mm} & 3 & 2 & 64$\times$H/8$\times$W/8 \\
\#4  & \hspace{2mm} Conv2d $\shortrightarrow$ GN $\shortrightarrow$ ReLU \hspace{2mm} & 3 & 2 & 128$\times$H/16$\times$W/16 \\
\#5  & \hspace{2mm} Conv2d $\shortrightarrow$ GN $\shortrightarrow$ ReLU \hspace{2mm} & 3 & 2 & 256$\times$H/32$\times$W/32 \\
\#6  & \hspace{2mm} Conv2d $\shortrightarrow$ GN $\shortrightarrow$ ReLU \hspace{2mm} & 3 & 2 & 256$\times$H/64$\times$W/64 \\
\#7  & \hspace{2mm} Conv2d $\shortrightarrow$ GN $\shortrightarrow$ ReLU \hspace{2mm} & 3 & 2 & 256$\times$H/128$\times$W/128 \\
\#8  & Conv2d & 1 & 1 & 6$\times$H/128$\times$W/128 \\
\midrule
\#9  & Global Pooling & - & - & 6 \\
\bottomrule
\end{tabular}
\label{tab:netpose}
}}
\\
\caption{
\textbf{Neural network architectures used in our proposed FSM framework}. The predicted depth maps are $1 \times H \times W$ tensors, and the predicted poses are $6$-dimensional vectors representing translation ($x,y,z$) and Euler rotation angles (pitch, yaw, roll).  \emph{BN} stands for Batch Normalization \cite{ioffe2015batch}, \emph{GN} for Group Normalization \cite{WuH18}, \emph{Upsample} doubles spatial dimensions using bilinear interpolation, and \emph{ReLU} denote Rectified Linear Units. The symbol $\oplus$ indicates feature concatenation. 
}
\label{tab:networks}
\vspace{-5mm}
\end{table}

%% file: supp_mat/ddad_qualitative.tex
\begin{figure*}[t!]
\centering
\subfloat{
\includegraphics[width=0.16\linewidth,height=1.8cm]{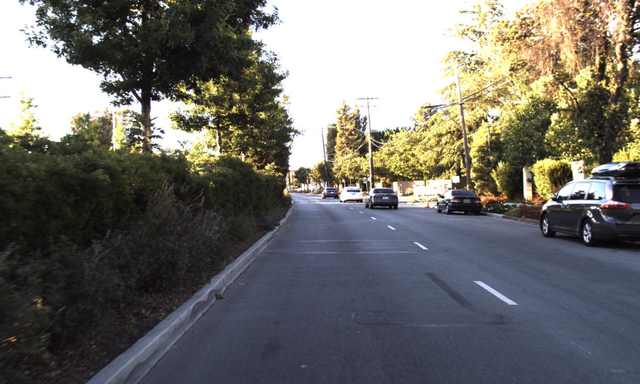}
\includegraphics[width=0.16\linewidth,height=1.8cm]{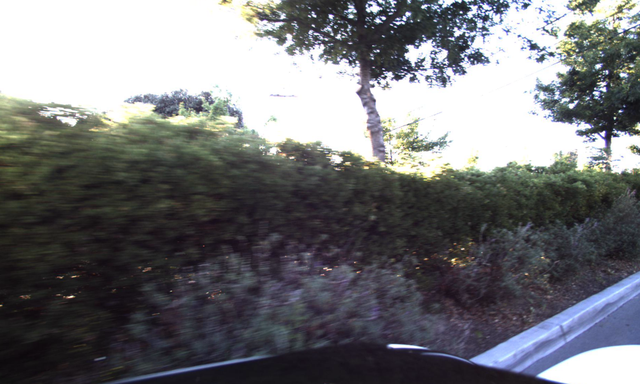}
\includegraphics[width=0.16\linewidth,height=1.8cm]{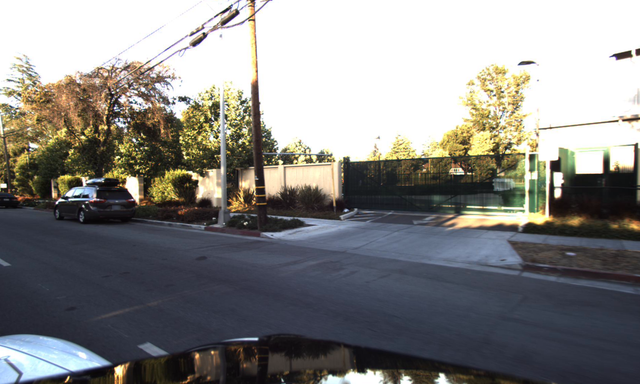}
\includegraphics[width=0.16\linewidth,height=1.8cm]{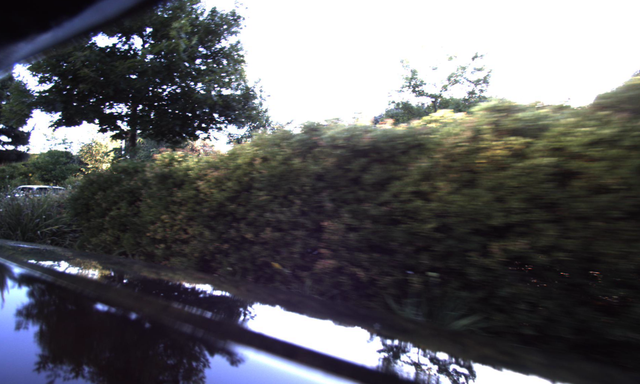}
\includegraphics[width=0.16\linewidth,height=1.8cm]{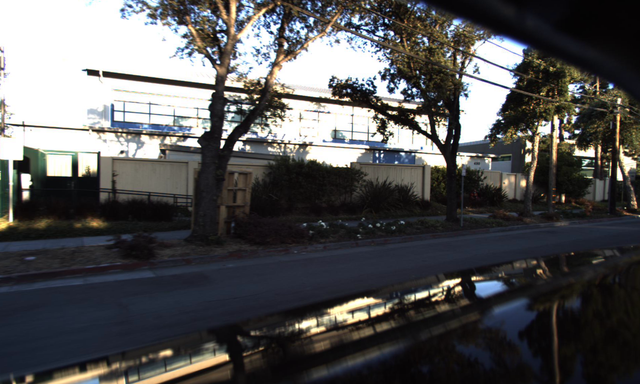}
\includegraphics[width=0.16\linewidth,height=1.8cm]{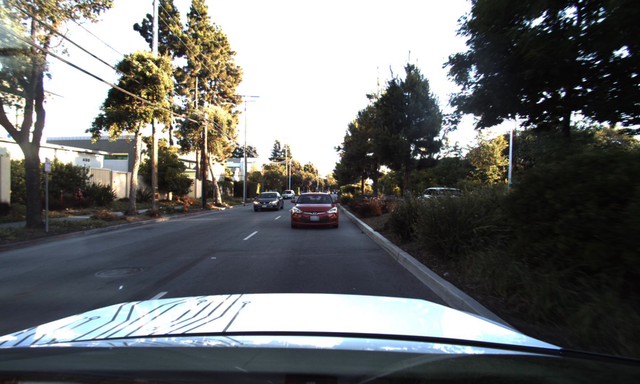}
}
\vspace{-4mm}
\\
\subfloat{
\includegraphics[width=0.16\linewidth,height=1.8cm]{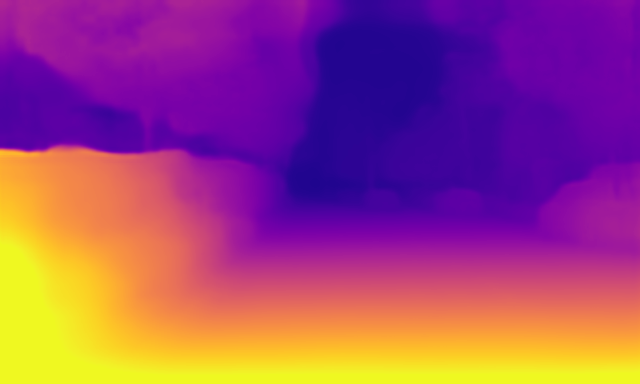}
\includegraphics[width=0.16\linewidth,height=1.8cm]{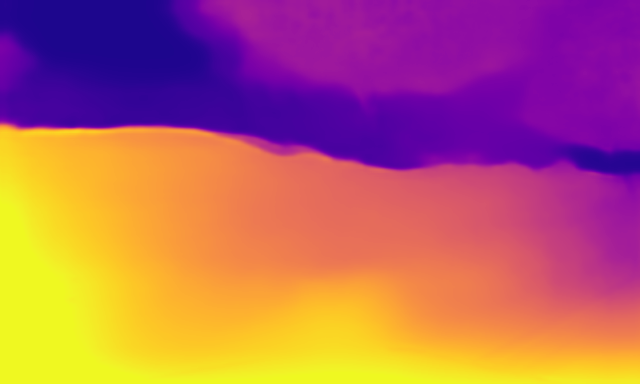}
\includegraphics[width=0.16\linewidth,height=1.8cm]{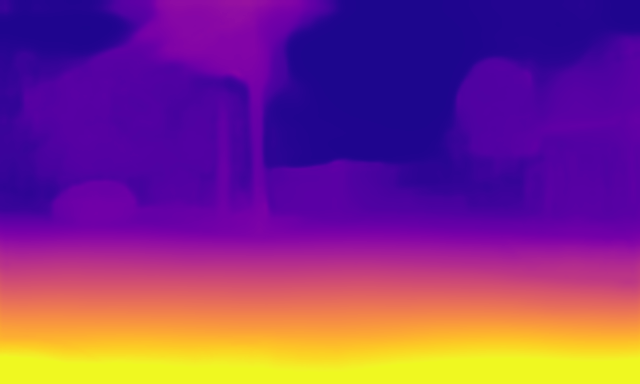}
\includegraphics[width=0.16\linewidth,height=1.8cm]{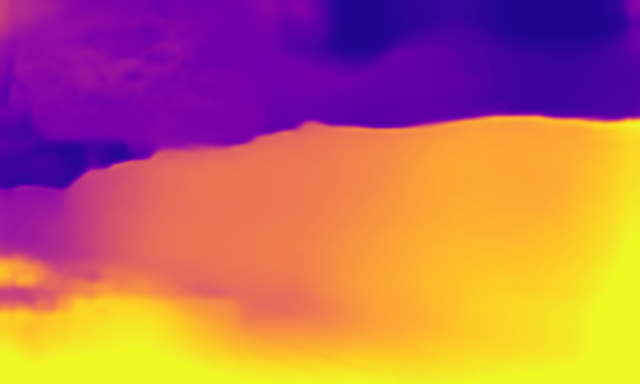}
\includegraphics[width=0.16\linewidth,height=1.8cm]{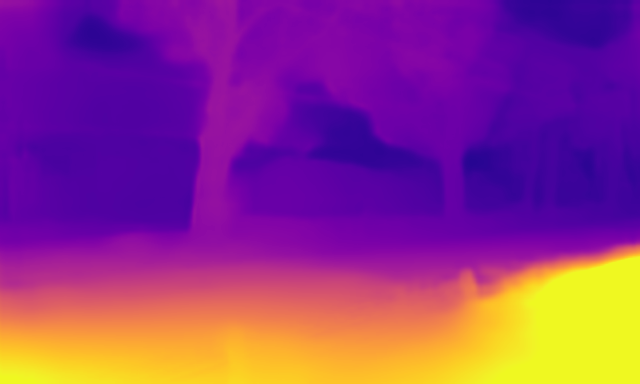}
\includegraphics[width=0.16\linewidth,height=1.8cm]{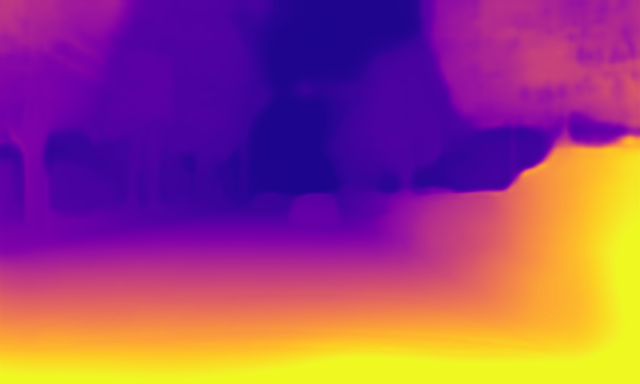}
}
\vspace{-2mm}
\\
\subfloat{
\includegraphics[width=0.16\linewidth,height=1.8cm]{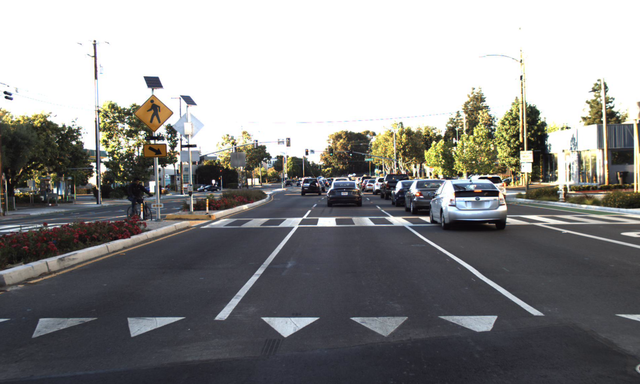}
\includegraphics[width=0.16\linewidth,height=1.8cm]{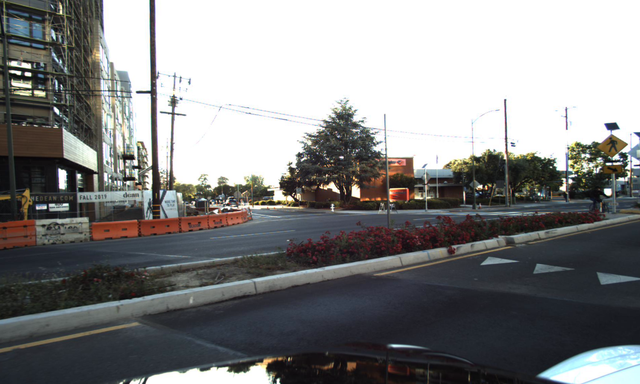}
\includegraphics[width=0.16\linewidth,height=1.8cm]{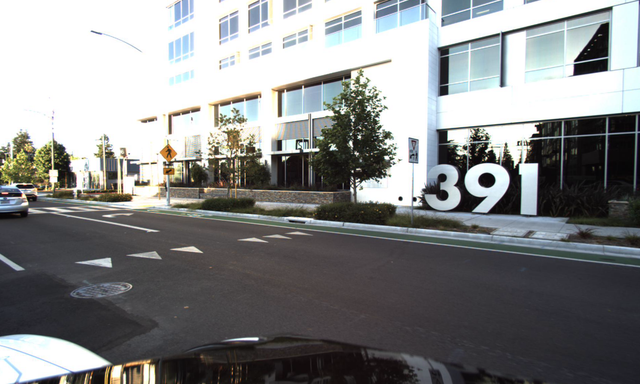}
\includegraphics[width=0.16\linewidth,height=1.8cm]{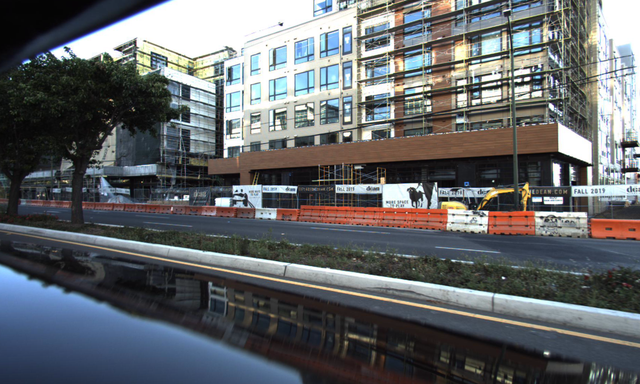}
\includegraphics[width=0.16\linewidth,height=1.8cm]{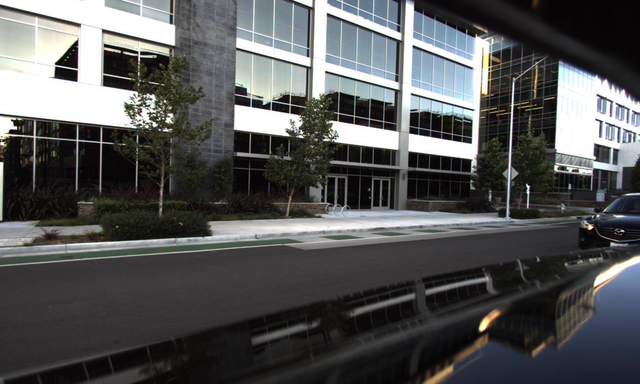}
\includegraphics[width=0.16\linewidth,height=1.8cm]{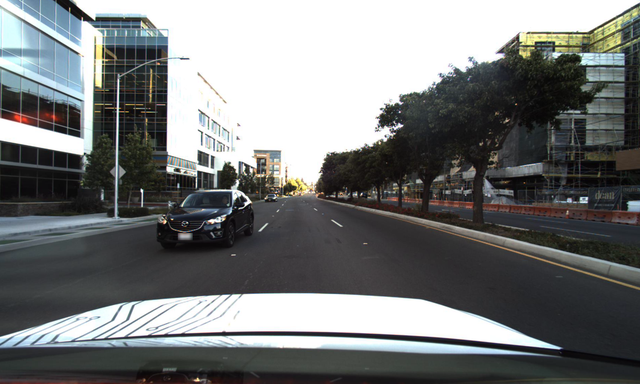}
}
\vspace{-4mm}
\\
\subfloat{
\includegraphics[width=0.16\linewidth,height=1.8cm]{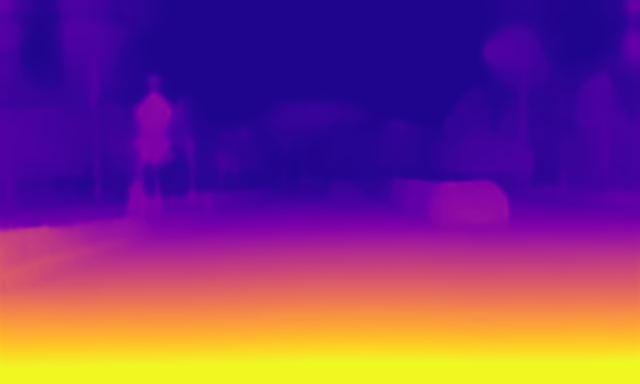}
\includegraphics[width=0.16\linewidth,height=1.8cm]{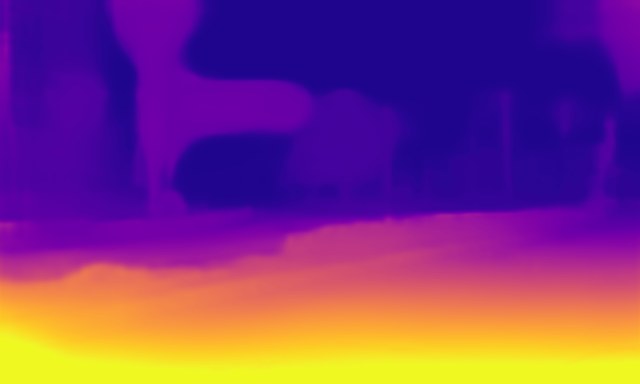}
\includegraphics[width=0.16\linewidth,height=1.8cm]{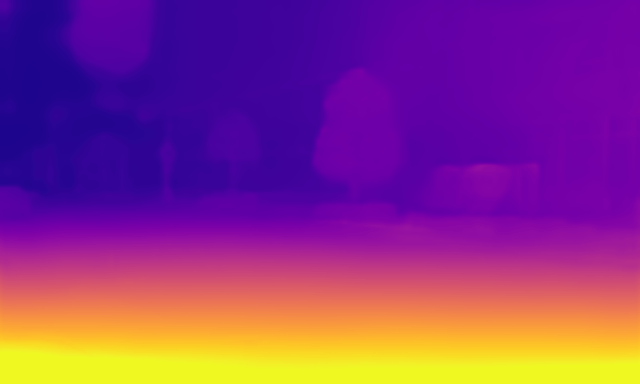}
\includegraphics[width=0.16\linewidth,height=1.8cm]{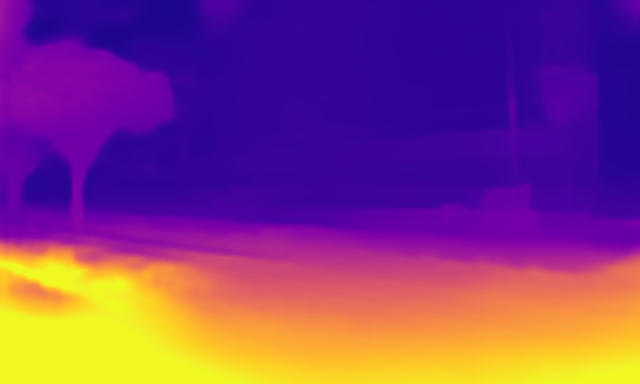}
\includegraphics[width=0.16\linewidth,height=1.8cm]{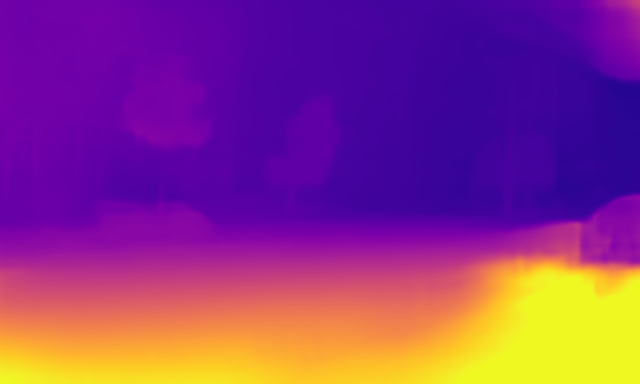}
\includegraphics[width=0.16\linewidth,height=1.8cm]{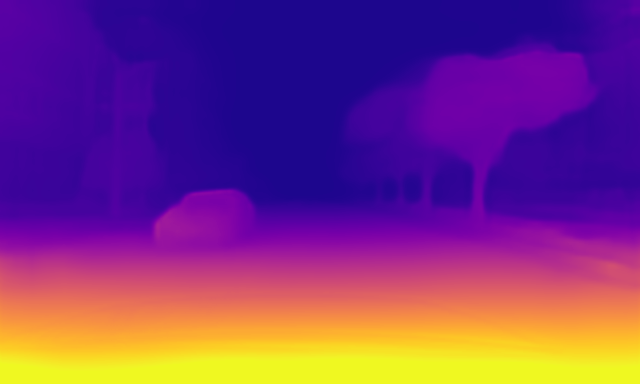}
}
\vspace{-2mm}
\\
\subfloat{
\includegraphics[width=0.16\linewidth,height=1.8cm]{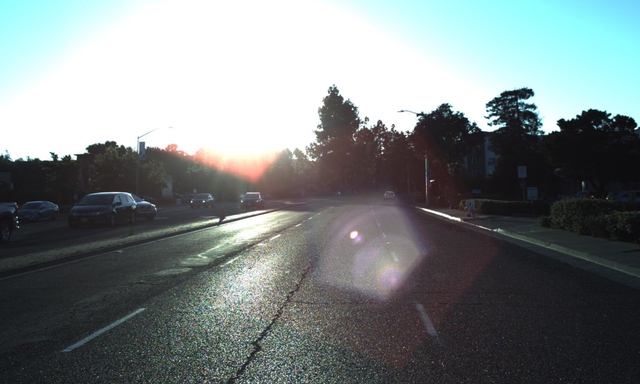}
\includegraphics[width=0.16\linewidth,height=1.8cm]{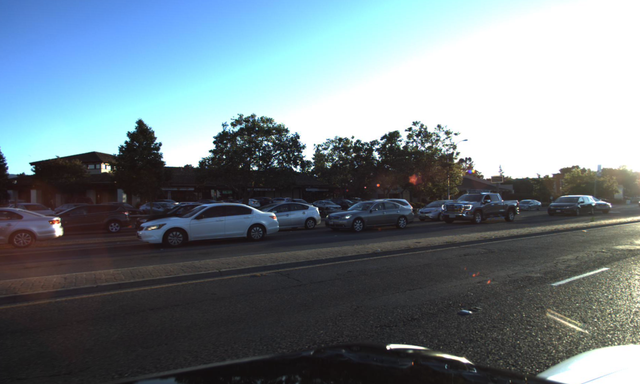}
\includegraphics[width=0.16\linewidth,height=1.8cm]{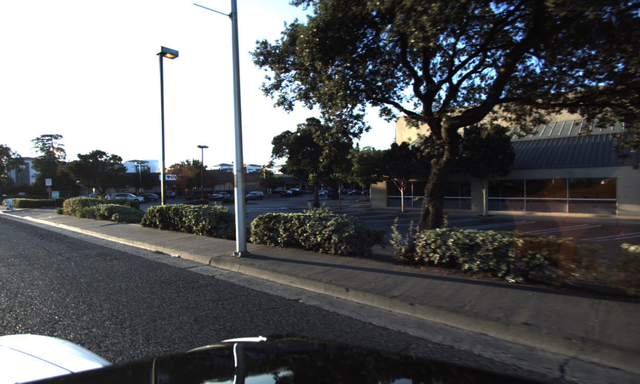}
\includegraphics[width=0.16\linewidth,height=1.8cm]{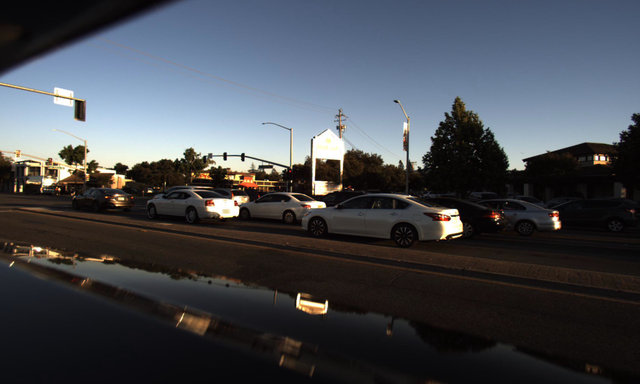}
\includegraphics[width=0.16\linewidth,height=1.8cm]{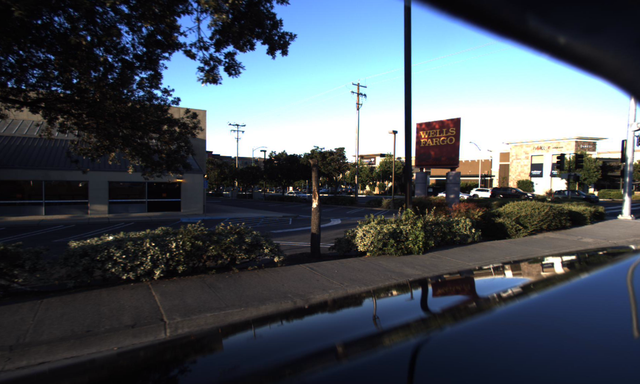}
\includegraphics[width=0.16\linewidth,height=1.8cm]{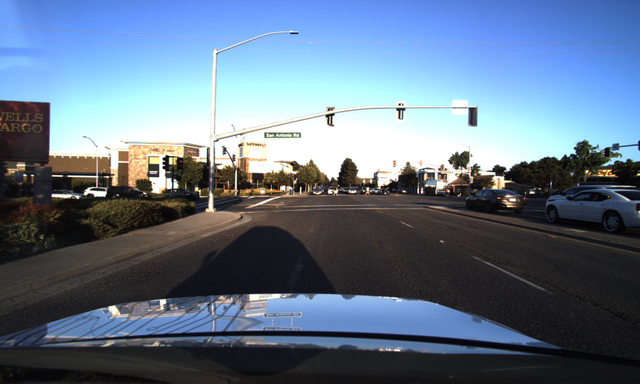}
}
\vspace{-4mm}
\\
\subfloat{
\includegraphics[width=0.16\linewidth,height=1.8cm]{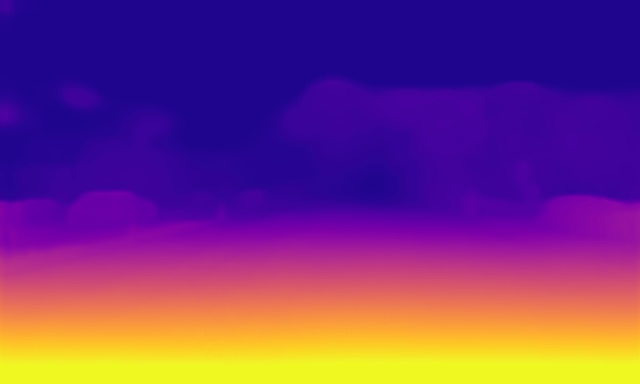}
\includegraphics[width=0.16\linewidth,height=1.8cm]{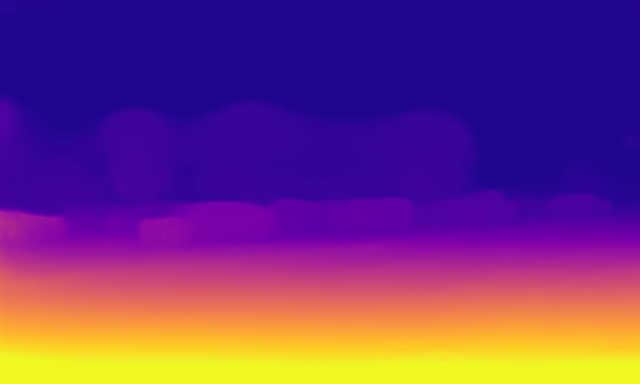}
\includegraphics[width=0.16\linewidth,height=1.8cm]{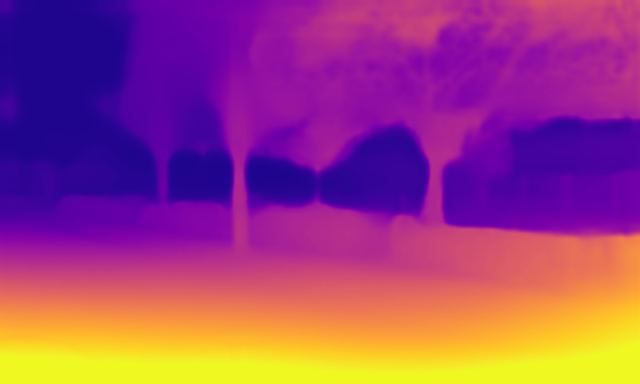}
\includegraphics[width=0.16\linewidth,height=1.8cm]{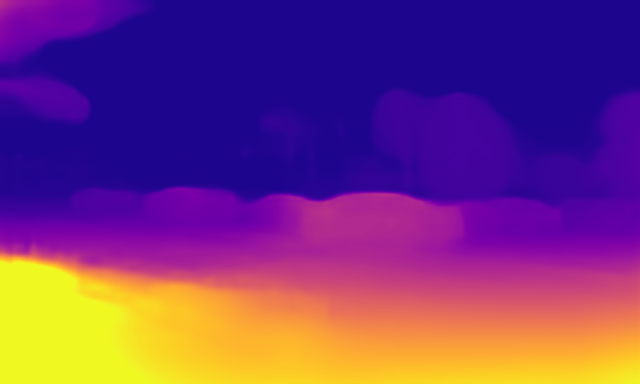}
\includegraphics[width=0.16\linewidth,height=1.8cm]{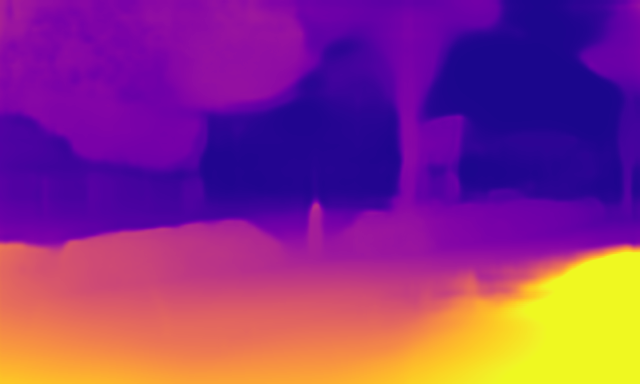}
\includegraphics[width=0.16\linewidth,height=1.8cm]{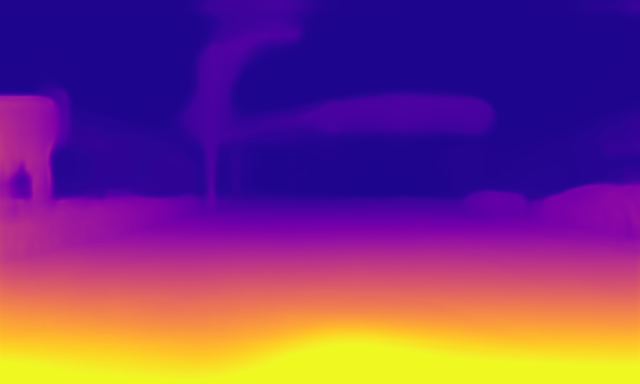}
}
\vspace{-2mm}
\\
\subfloat{
\includegraphics[width=0.16\linewidth,height=1.8cm]{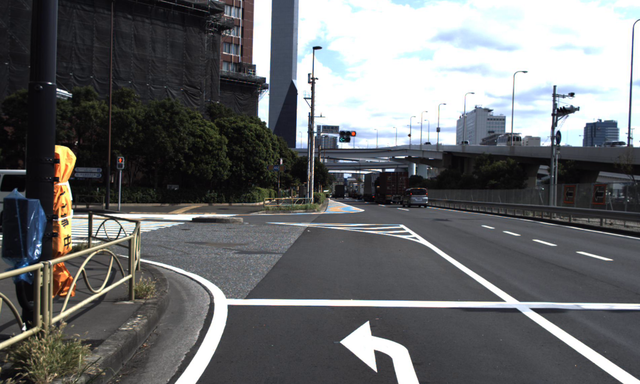}
\includegraphics[width=0.16\linewidth,height=1.8cm]{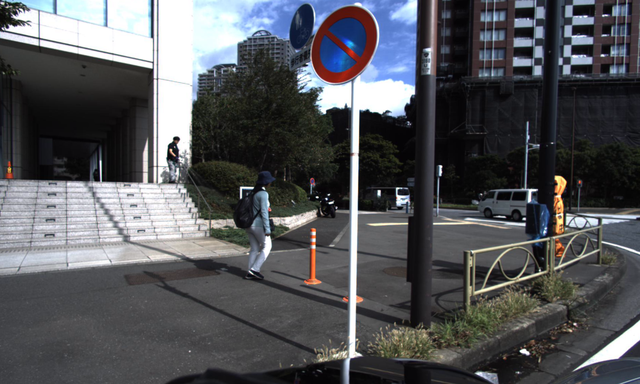}
\includegraphics[width=0.16\linewidth,height=1.8cm]{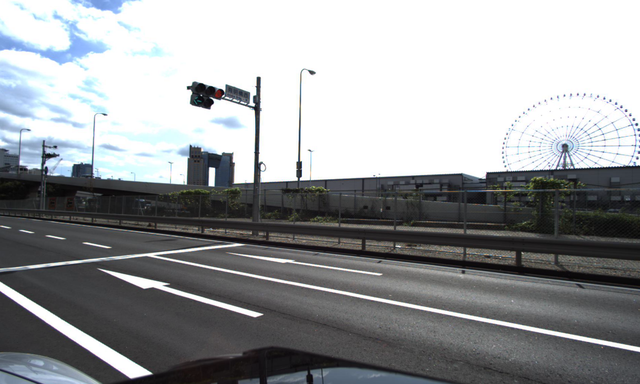}
\includegraphics[width=0.16\linewidth,height=1.8cm]{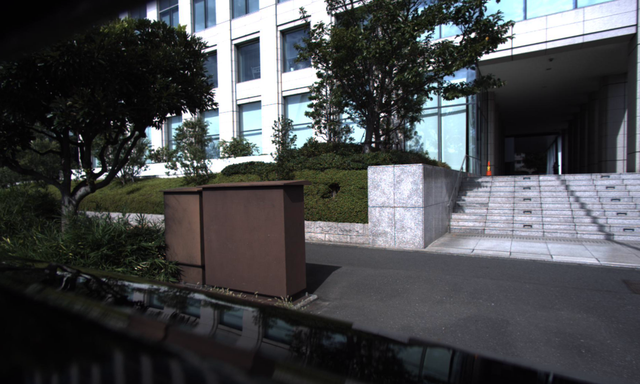}
\includegraphics[width=0.16\linewidth,height=1.8cm]{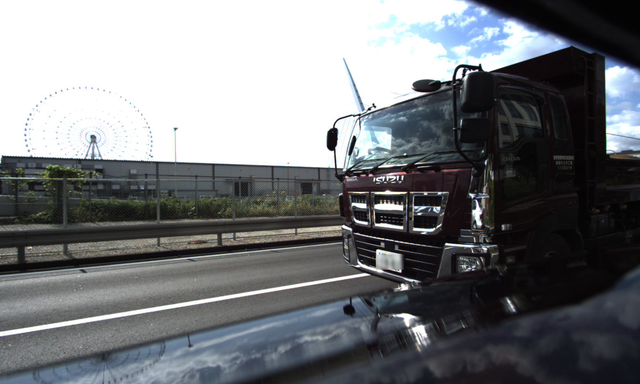}
\includegraphics[width=0.16\linewidth,height=1.8cm]{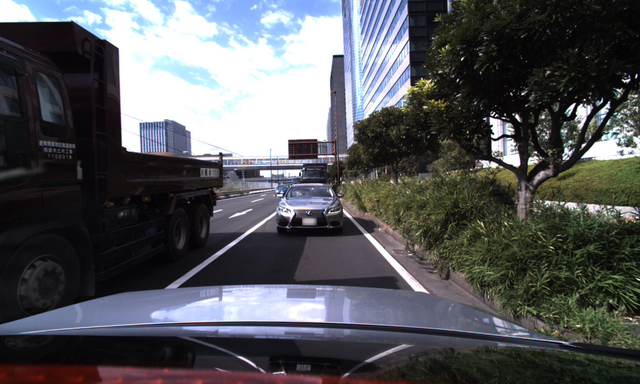}
}
\vspace{-4mm}
\\
\subfloat{
\includegraphics[width=0.16\linewidth,height=1.8cm]{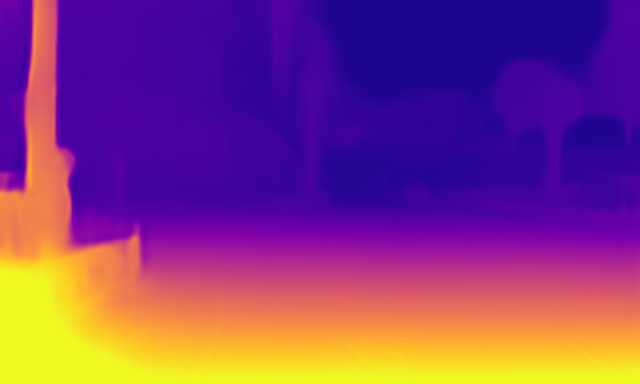}
\includegraphics[width=0.16\linewidth,height=1.8cm]{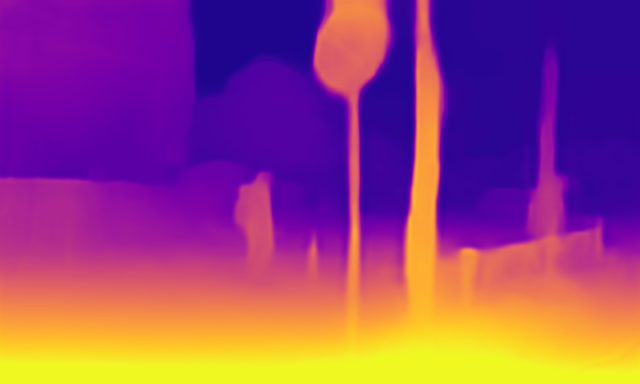}
\includegraphics[width=0.16\linewidth,height=1.8cm]{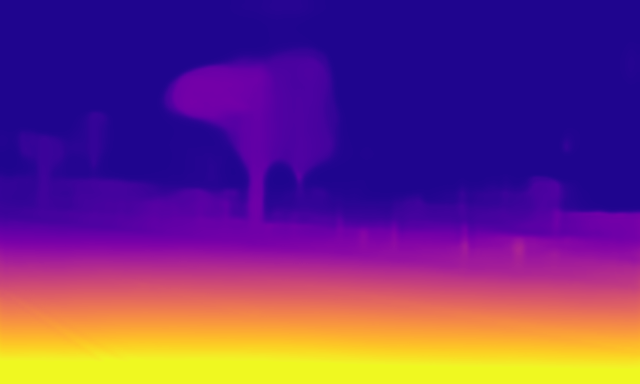}
\includegraphics[width=0.16\linewidth,height=1.8cm]{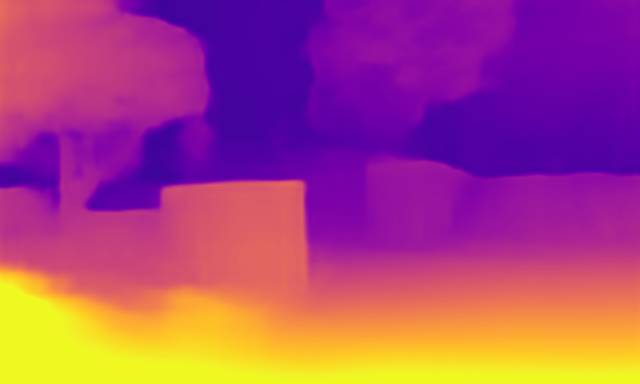}
\includegraphics[width=0.16\linewidth,height=1.8cm]{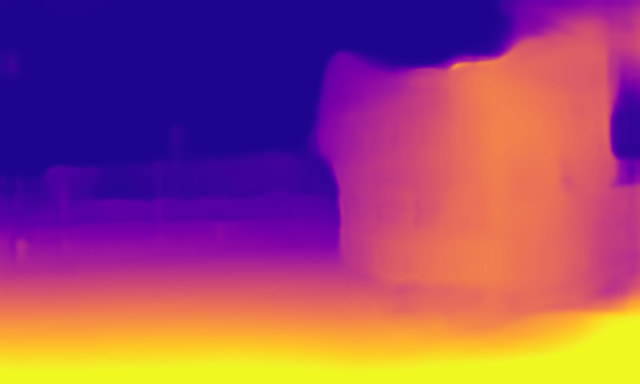}
\includegraphics[width=0.16\linewidth,height=1.8cm]{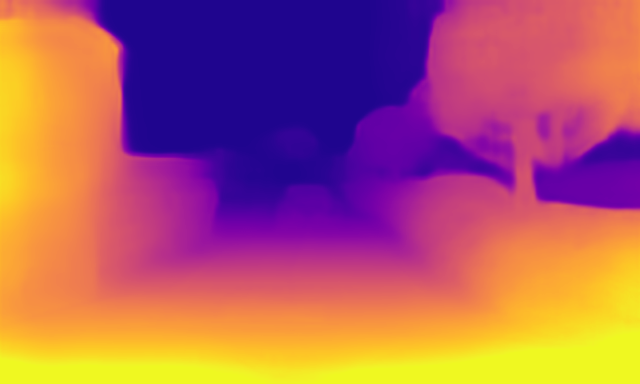}
}
\caption{\textbf{Self-Supervised depth estimation FSM results} on the \textit{DDAD} dataset.  Unlike \textit{nuScenes}, \emph{DDAD} scenes were curated with the intent of depth estimation training, containing higher quality images with larger overlap between cameras (compare with nuScenes images in Figure~\ref{fig:nuscenes_qualitative_supp}). This higher data quality leads to the difference in metrics reported in the paper on these two datasets.}
\label{fig:ddad_qualitative_supp}
\vspace{-2mm}
\end{figure*}


%% file: supp_mat/nuscenes_qualitative.tex
\begin{figure*}[t!]
\centering
\subfloat{
\includegraphics[width=0.16\linewidth,height=1.7cm]{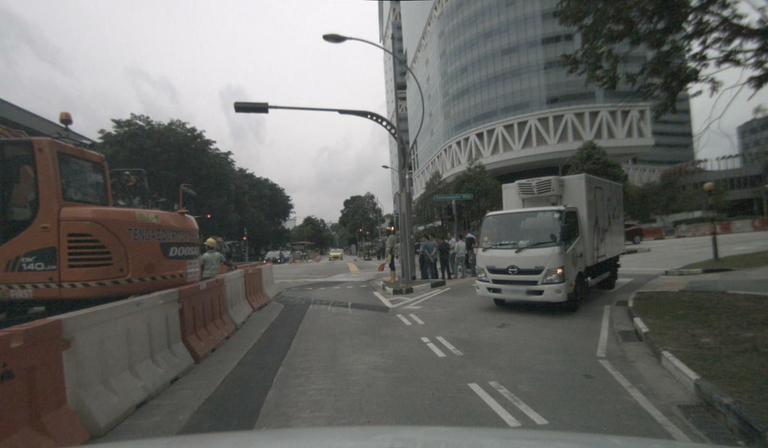}
\includegraphics[width=0.16\linewidth,height=1.7cm]{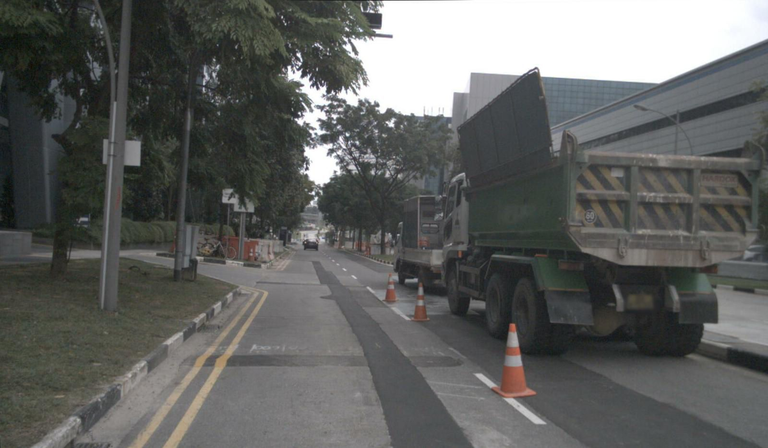}
\includegraphics[width=0.16\linewidth,height=1.7cm]{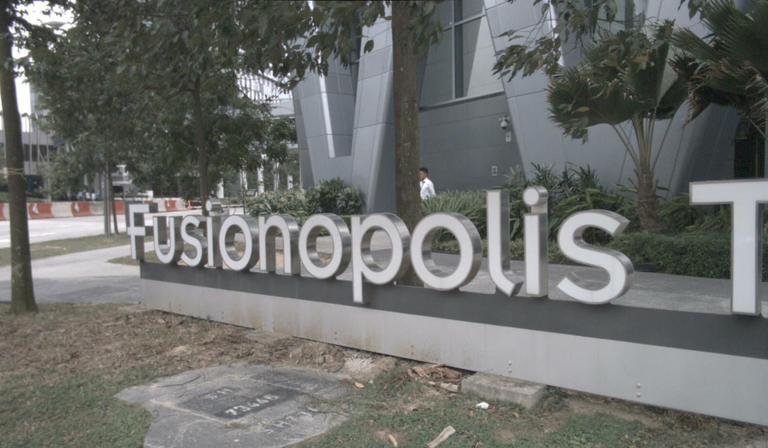}
\includegraphics[width=0.16\linewidth,height=1.7cm]{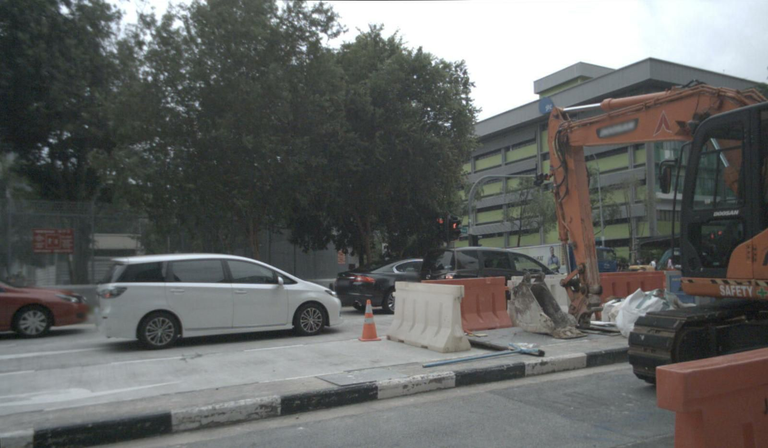}
\includegraphics[width=0.16\linewidth,height=1.7cm]{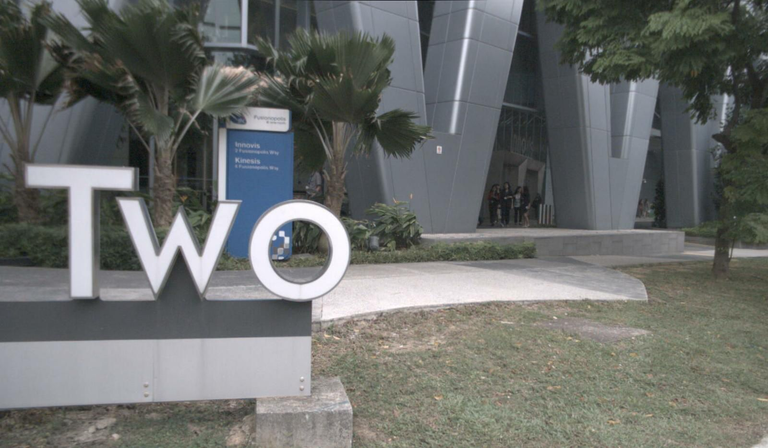}
\includegraphics[width=0.16\linewidth,height=1.7cm]{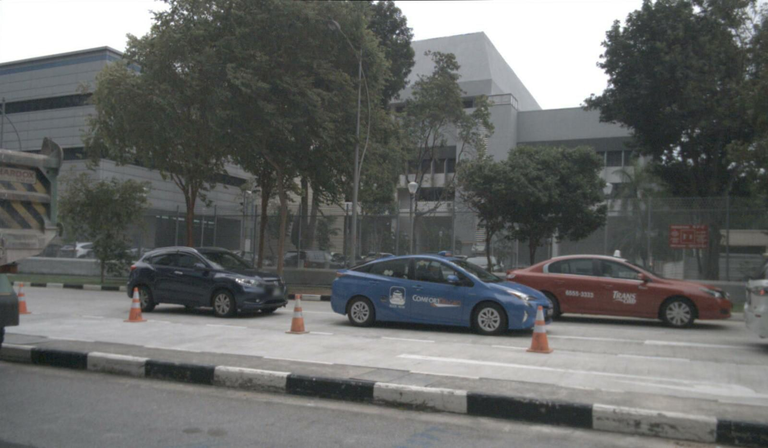}
}
\vspace{-4mm}
\\
\subfloat{
\includegraphics[width=0.16\linewidth,height=1.7cm]{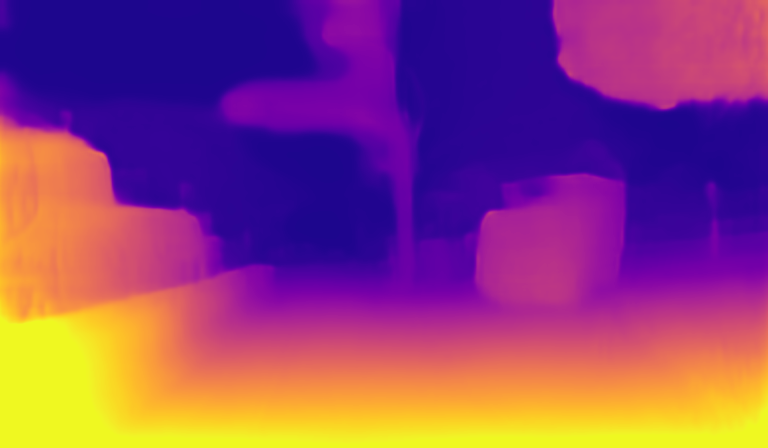}
\includegraphics[width=0.16\linewidth,height=1.7cm]{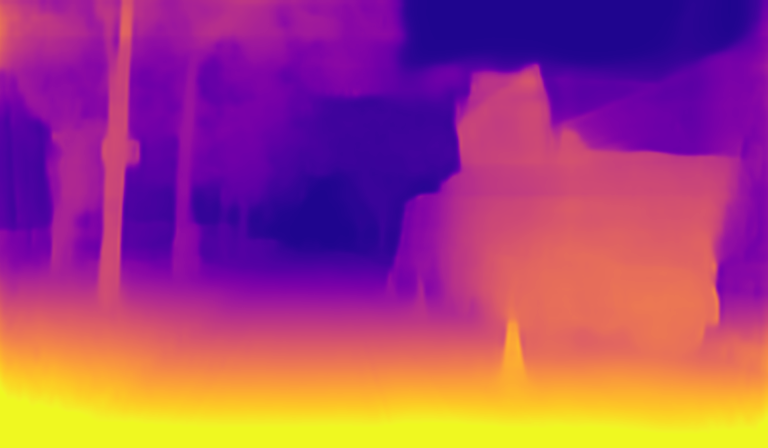}
\includegraphics[width=0.16\linewidth,height=1.7cm]{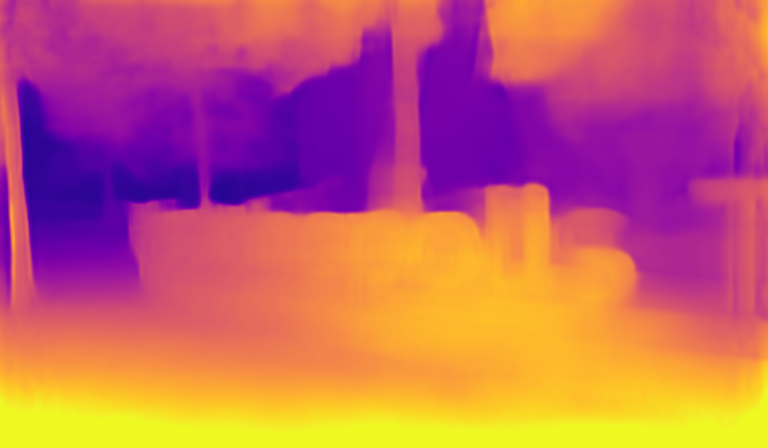}
\includegraphics[width=0.16\linewidth,height=1.7cm]{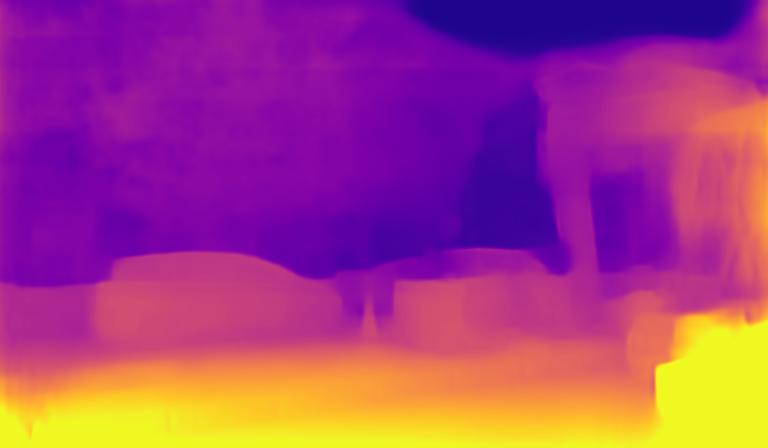}
\includegraphics[width=0.16\linewidth,height=1.7cm]{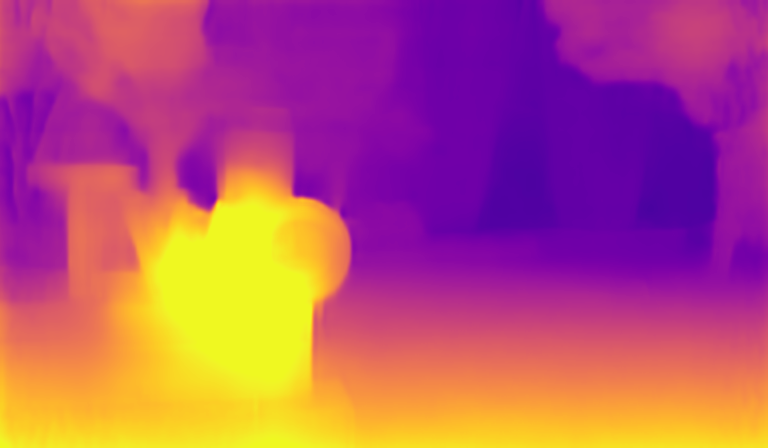}
\includegraphics[width=0.16\linewidth,height=1.7cm]{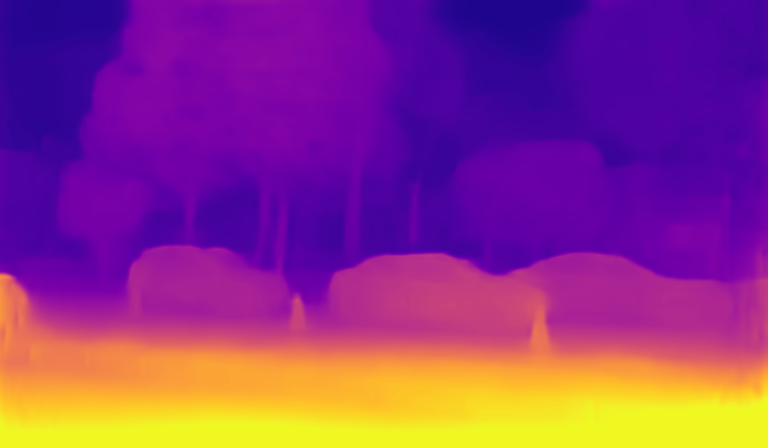}
}
\vspace{-2mm}
\\
\subfloat{
\includegraphics[width=0.16\linewidth,height=1.7cm]{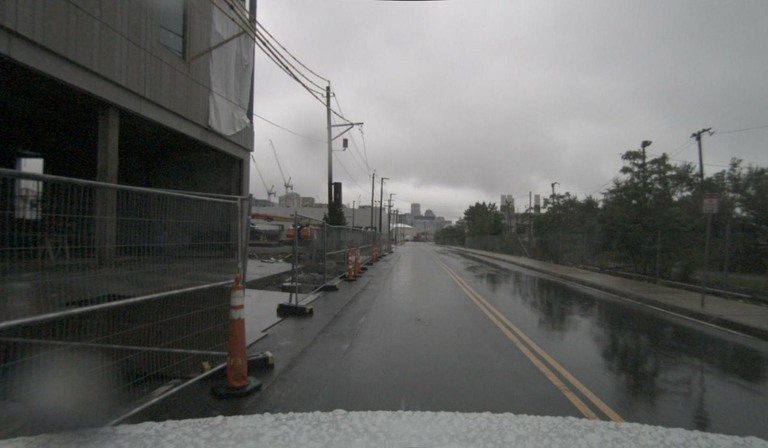}
\includegraphics[width=0.16\linewidth,height=1.7cm]{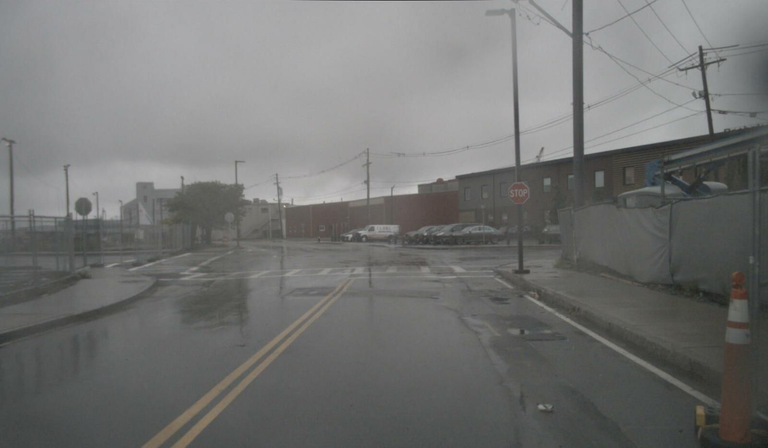}
\includegraphics[width=0.16\linewidth,height=1.7cm]{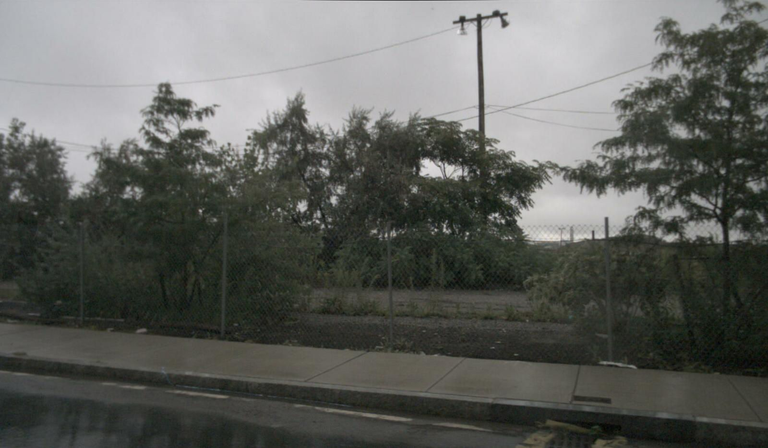}
\includegraphics[width=0.16\linewidth,height=1.7cm]{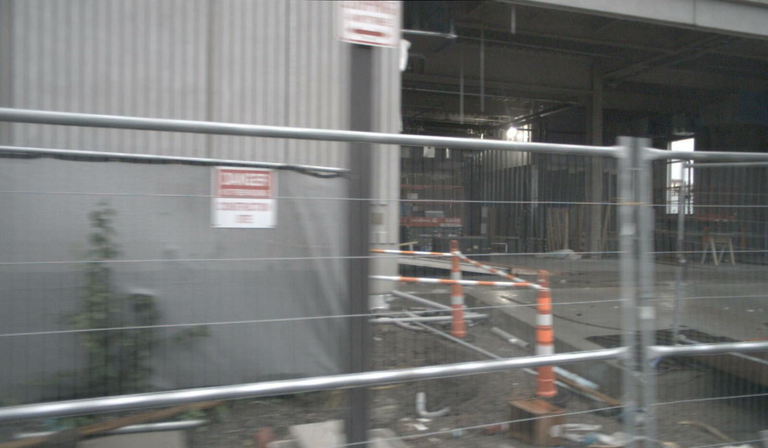}
\includegraphics[width=0.16\linewidth,height=1.7cm]{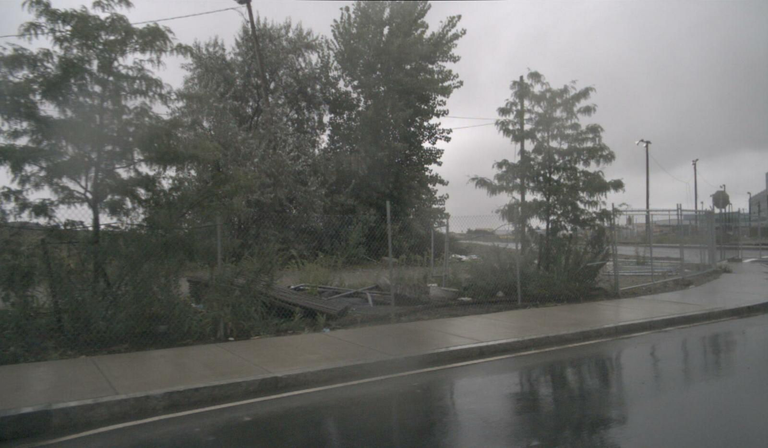}
\includegraphics[width=0.16\linewidth,height=1.7cm]{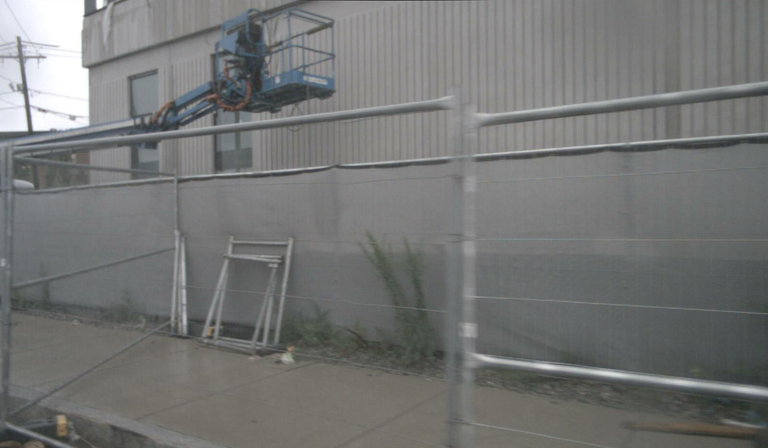}
}
\vspace{-4mm}
\\
\subfloat{
\includegraphics[width=0.16\linewidth,height=1.7cm]{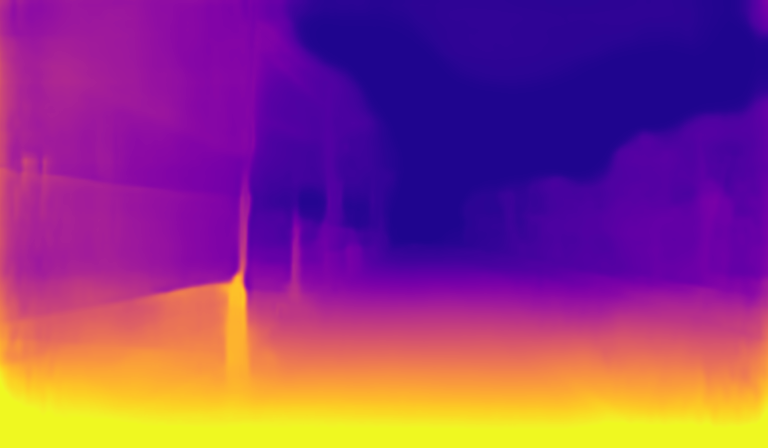}
\includegraphics[width=0.16\linewidth,height=1.7cm]{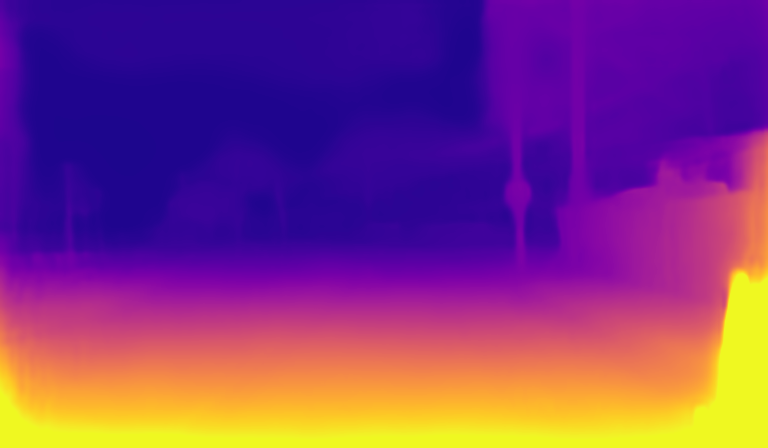}
\includegraphics[width=0.16\linewidth,height=1.7cm]{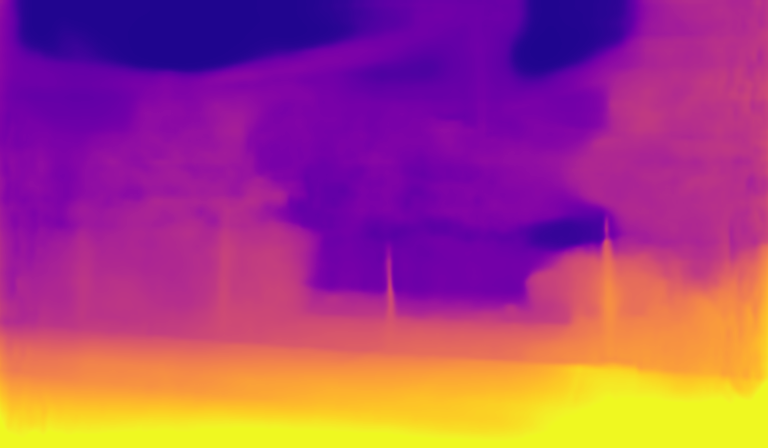}
\includegraphics[width=0.16\linewidth,height=1.7cm]{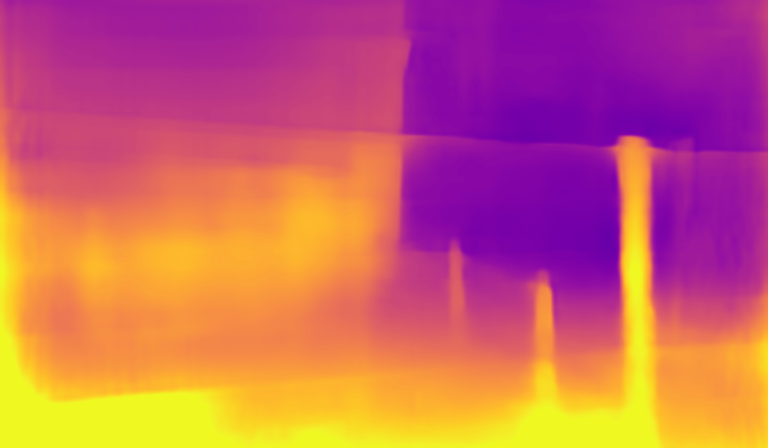}
\includegraphics[width=0.16\linewidth,height=1.7cm]{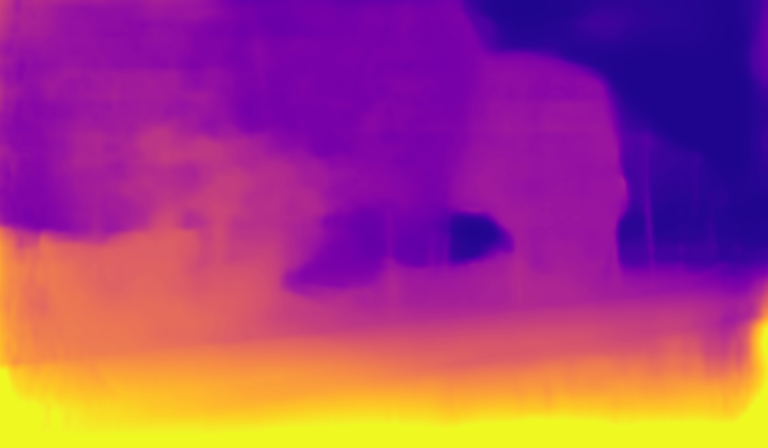}
\includegraphics[width=0.16\linewidth,height=1.7cm]{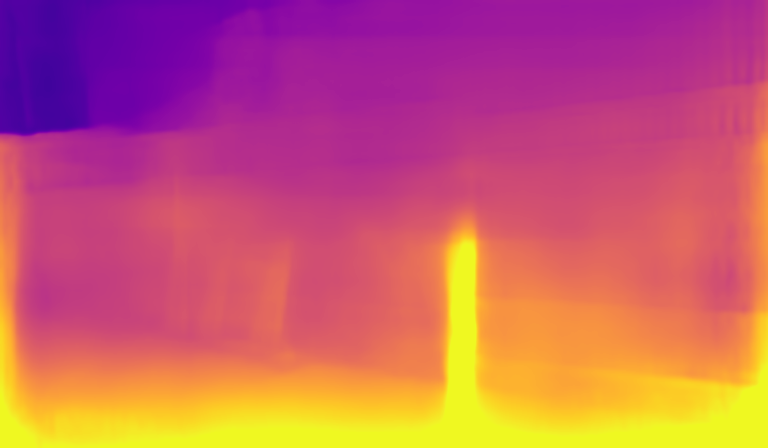}
}
\vspace{-2mm}
\\
\subfloat{
\includegraphics[width=0.16\linewidth,height=1.7cm]{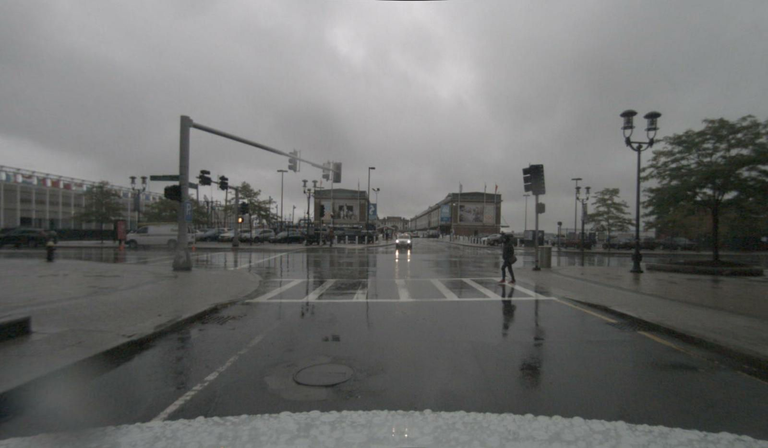}
\includegraphics[width=0.16\linewidth,height=1.7cm]{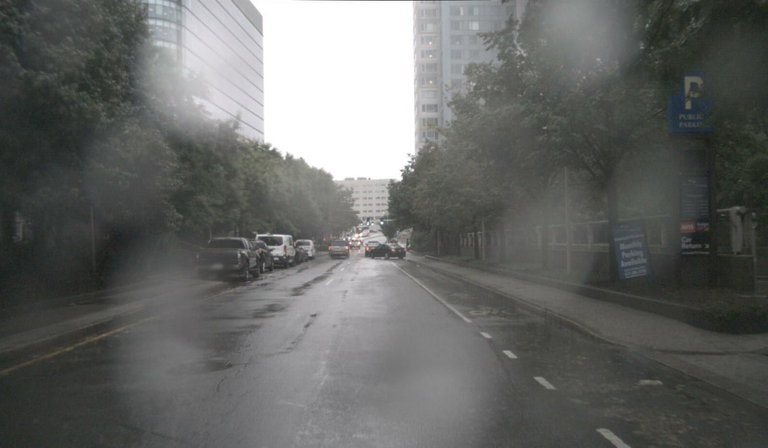}
\includegraphics[width=0.16\linewidth,height=1.7cm]{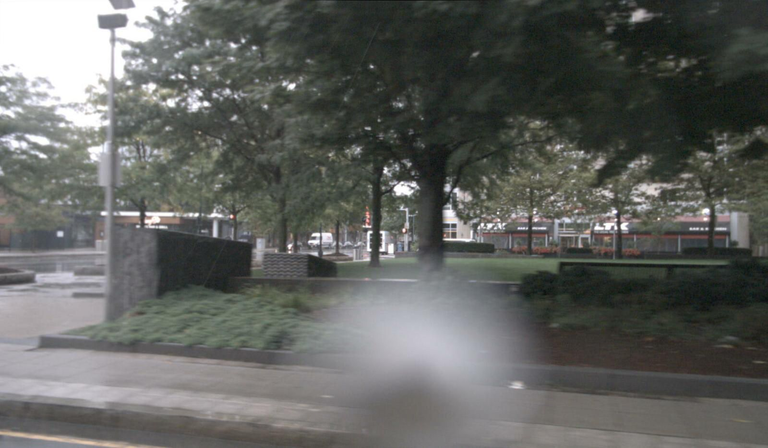}
\includegraphics[width=0.16\linewidth,height=1.7cm]{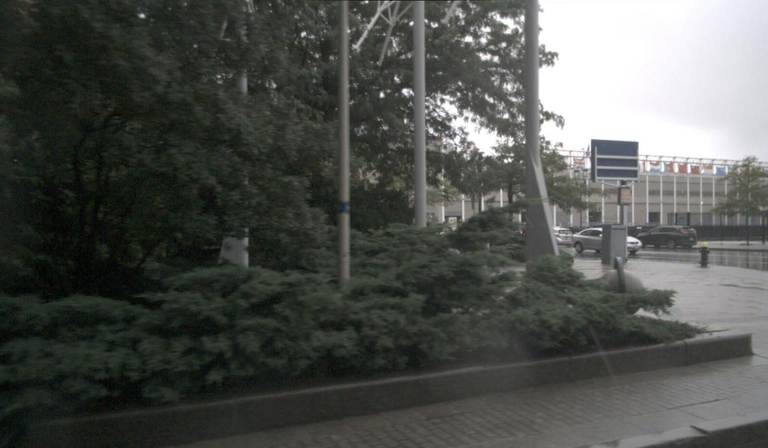}
\includegraphics[width=0.16\linewidth,height=1.7cm]{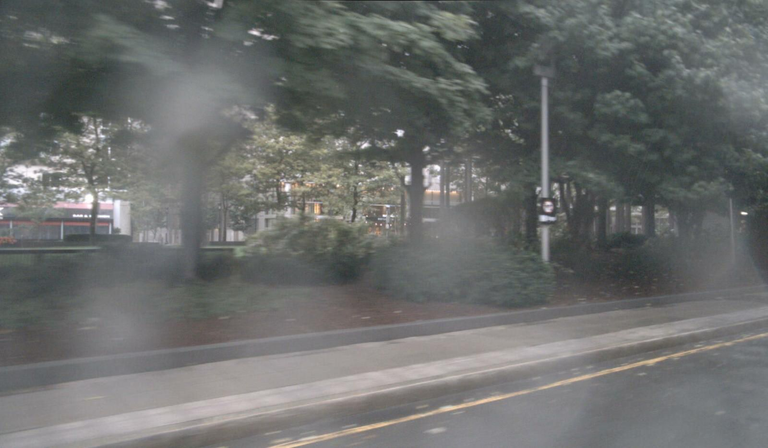}
\includegraphics[width=0.16\linewidth,height=1.7cm]{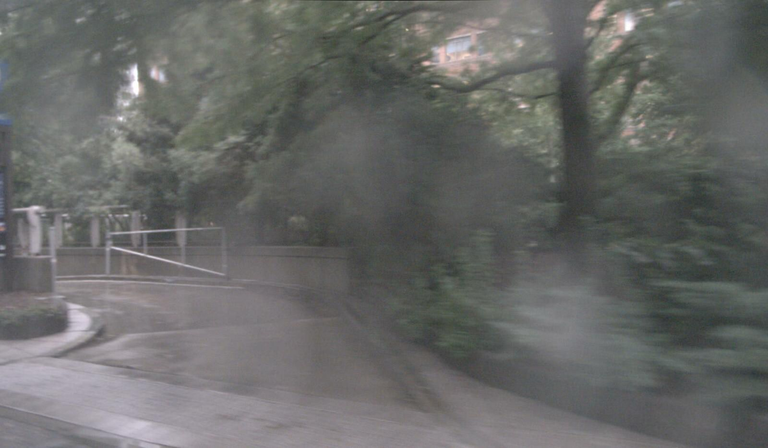}
}
\vspace{-4mm}
\\
\subfloat{
\includegraphics[width=0.16\linewidth,height=1.7cm]{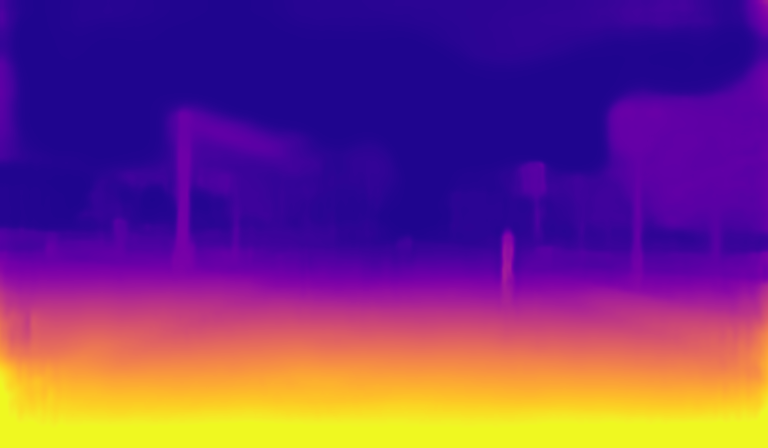}
\includegraphics[width=0.16\linewidth,height=1.7cm]{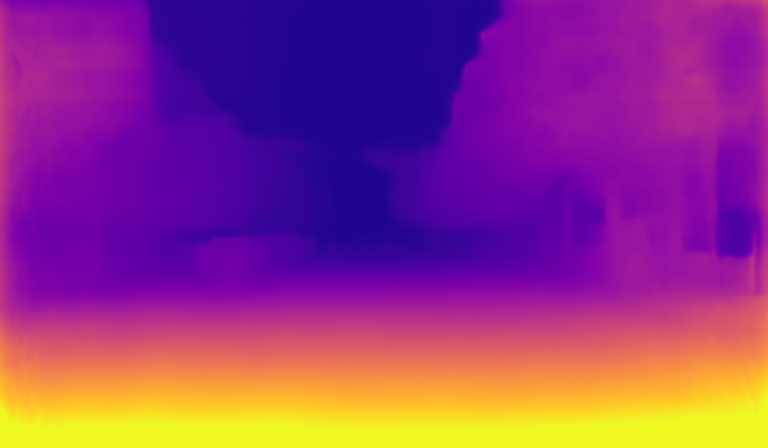}
\includegraphics[width=0.16\linewidth,height=1.7cm]{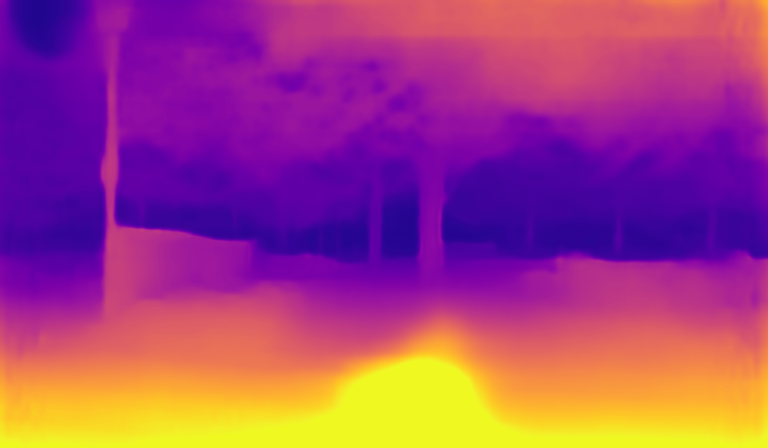}
\includegraphics[width=0.16\linewidth,height=1.7cm]{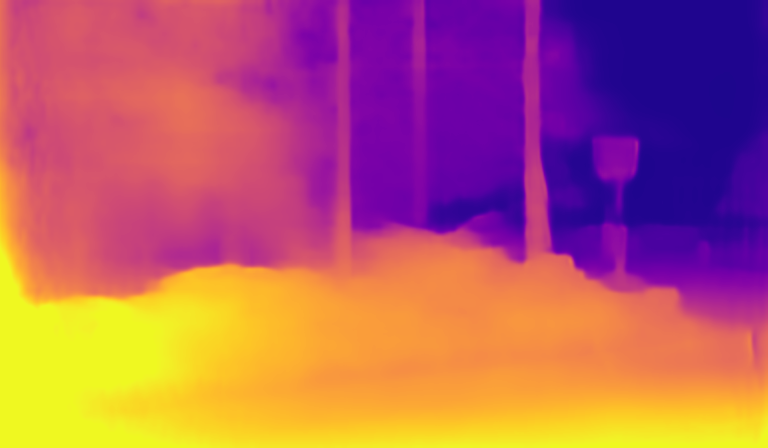}
\includegraphics[width=0.16\linewidth,height=1.7cm]{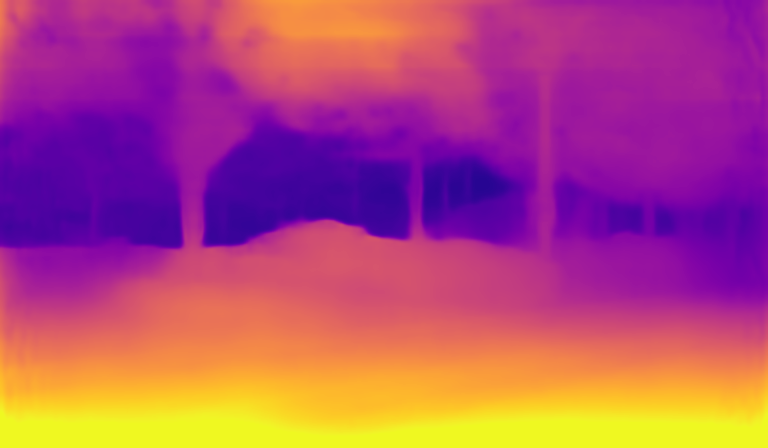}
\includegraphics[width=0.16\linewidth,height=1.7cm]{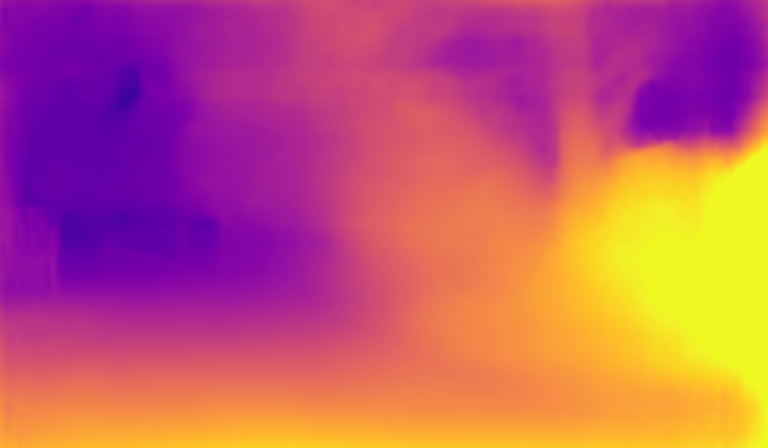}
}
\vspace{-2mm}
\\
\subfloat{
\includegraphics[width=0.16\linewidth,height=1.7cm]{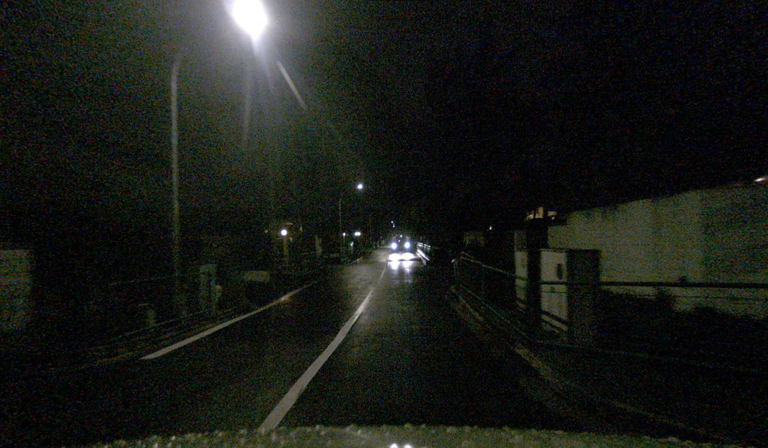}
\includegraphics[width=0.16\linewidth,height=1.7cm]{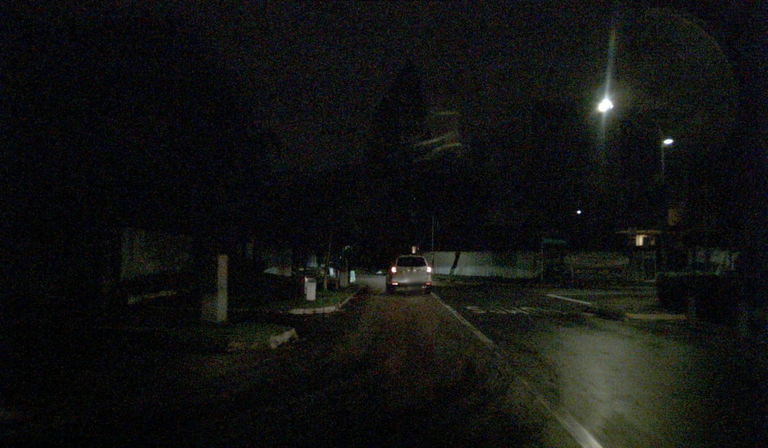}
\includegraphics[width=0.16\linewidth,height=1.7cm]{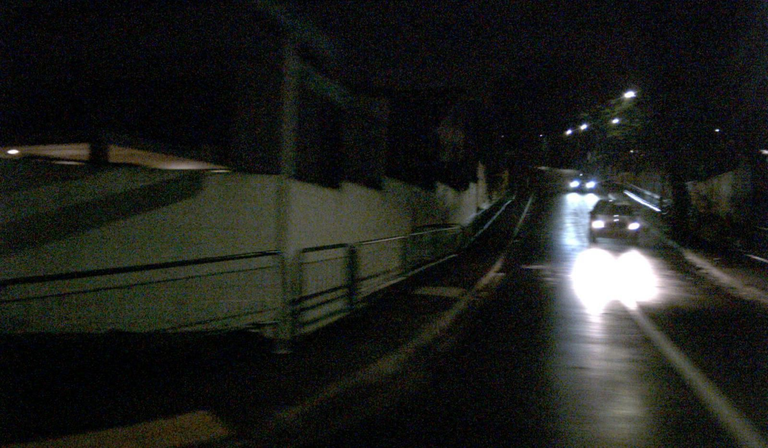}
\includegraphics[width=0.16\linewidth,height=1.7cm]{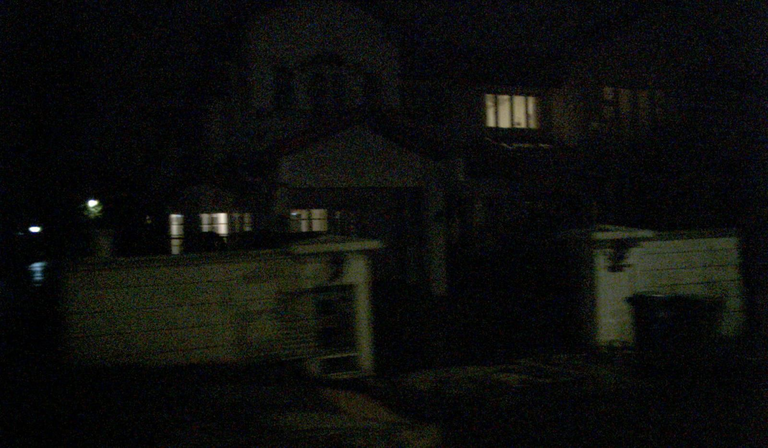}
\includegraphics[width=0.16\linewidth,height=1.7cm]{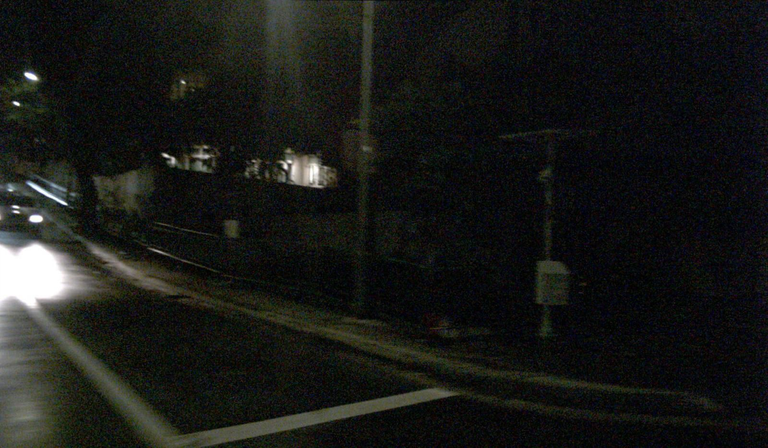}
\includegraphics[width=0.16\linewidth,height=1.7cm]{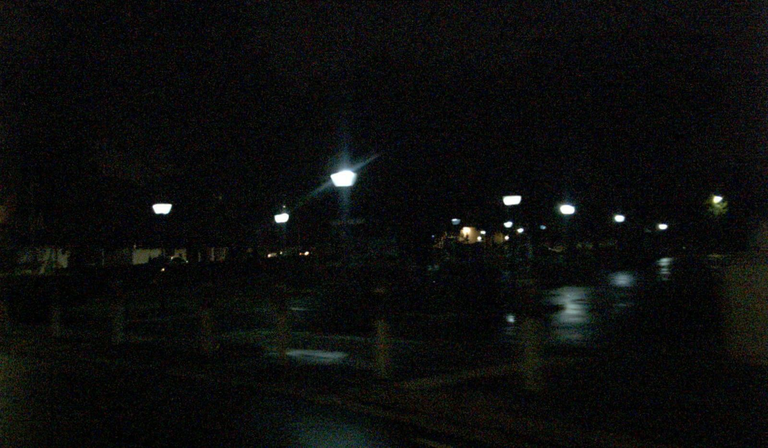}
}
\vspace{-4mm}
\\
\subfloat{
\includegraphics[width=0.16\linewidth,height=1.7cm]{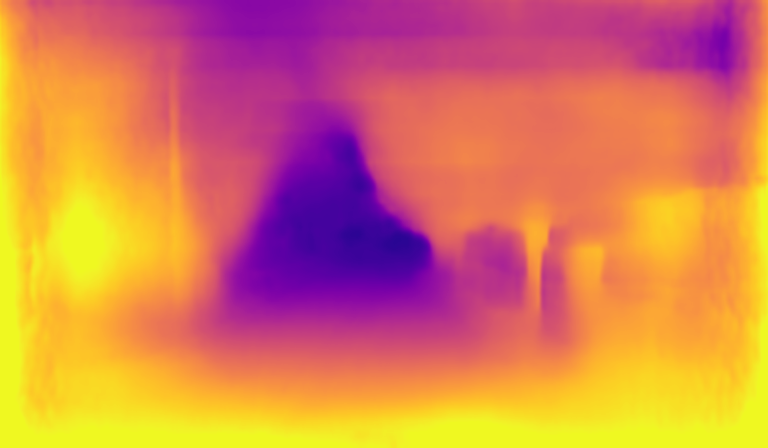}
\includegraphics[width=0.16\linewidth,height=1.7cm]{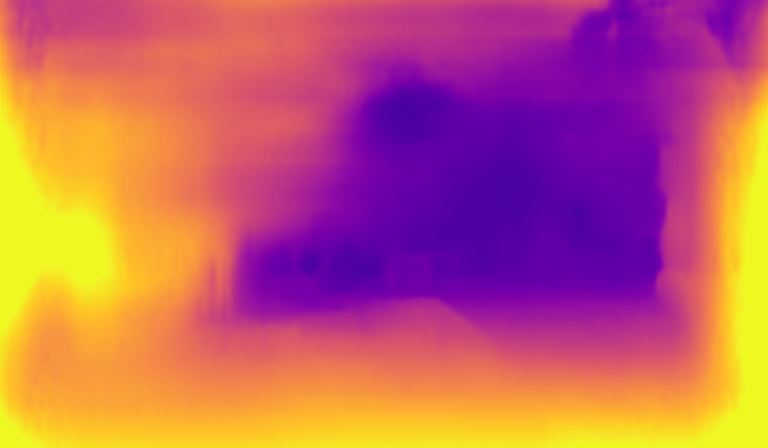}
\includegraphics[width=0.16\linewidth,height=1.7cm]{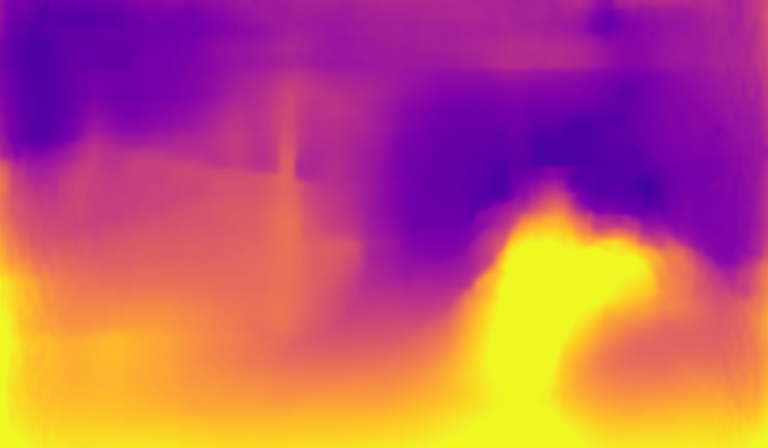}
\includegraphics[width=0.16\linewidth,height=1.7cm]{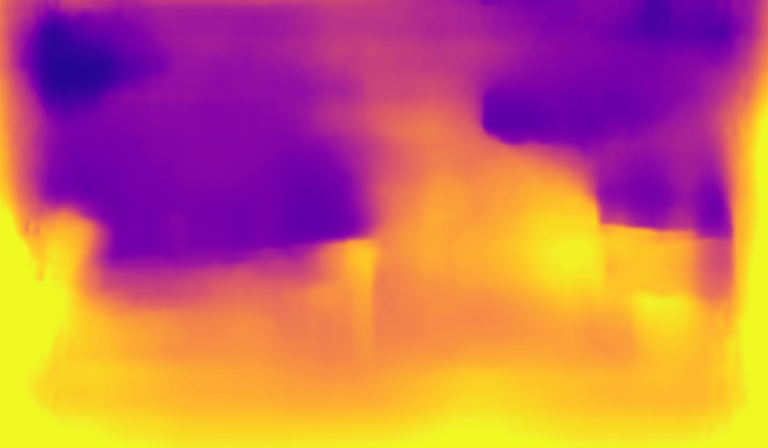}
\includegraphics[width=0.16\linewidth,height=1.7cm]{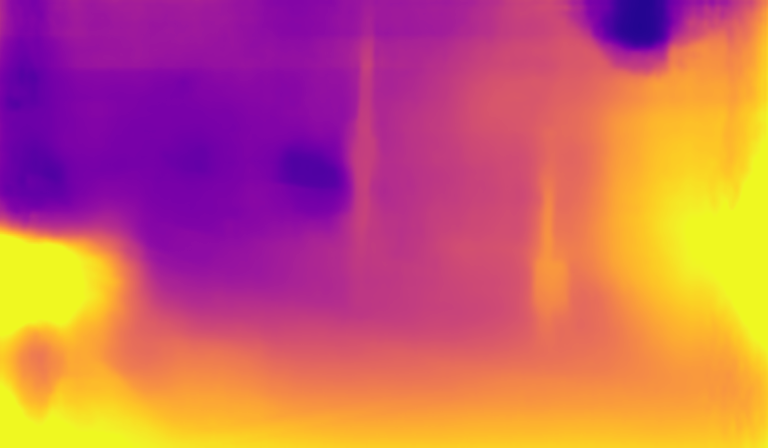}
\includegraphics[width=0.16\linewidth,height=1.7cm]{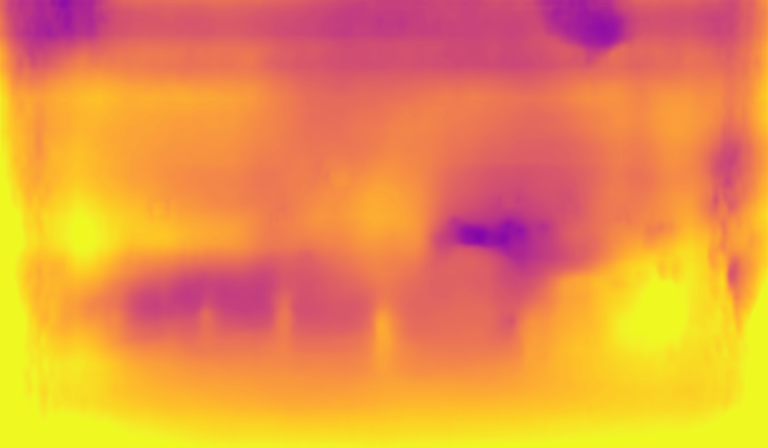}
}
\caption{\textbf{Self-Supervised depth estimation FSM results} on the \textit{nuScenes} dataset. The lower resolution and smaller overlap between cameras, combined with diverse weather conditions, time of day and observed structures (especially on side-facing cameras), make \textit{nuScenes} a particularly challenging dataset for self-supervised multi-camera depth estimation.}
\vspace{-2mm}
\label{fig:nuscenes_qualitative_supp}
\end{figure*}

%% file: supp_mat/overlapping.tex
\begin{figure*}[t!]
\centering
\subfloat[\textit{DDAD}]{
\includegraphics[width=0.92\linewidth]{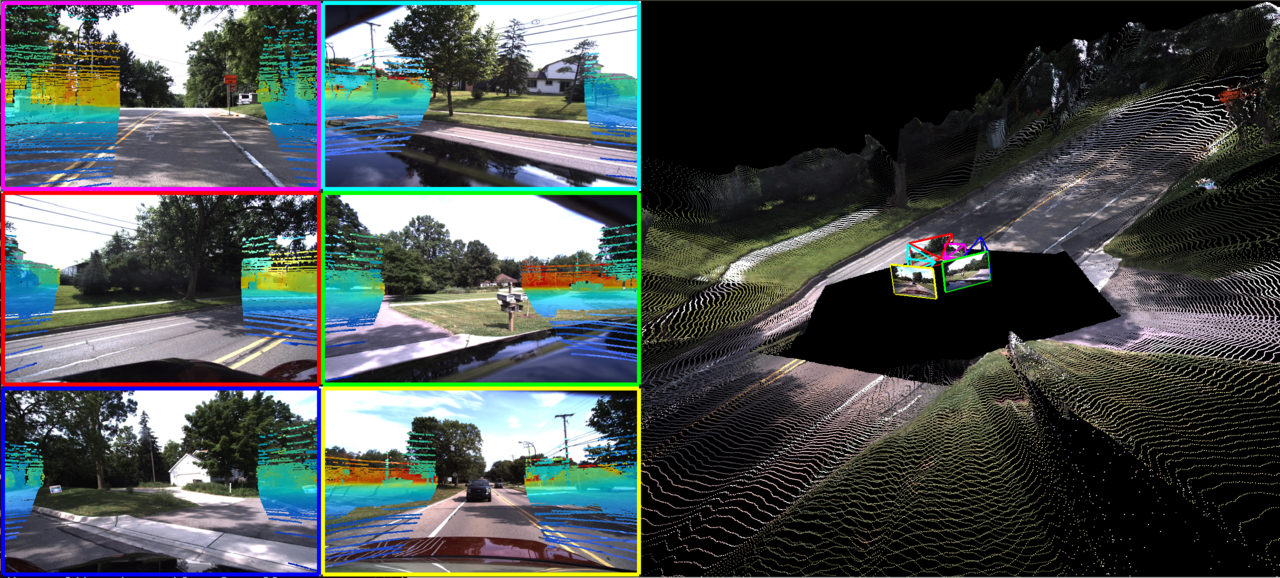}
}
\\
\vspace{-3mm}
\subfloat[\textit{nuScenes}]{
\includegraphics[width=0.92\linewidth]{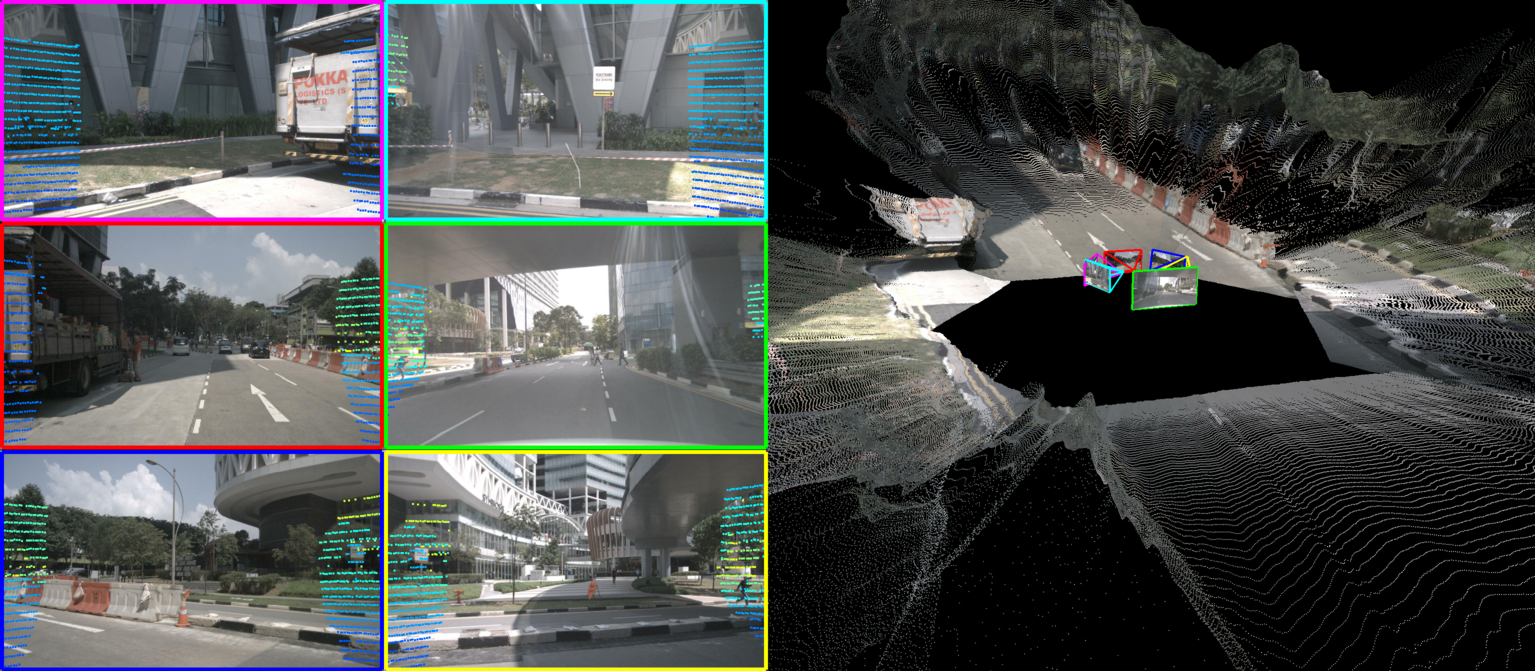}
}
\label{fig:pred_pcl}
\caption{
\textbf{Predicted 360$\degree$ scale-aware FSM pointclouds} on different multi-camera datasets. Each image is processed by the same network, and the predicted depth maps are lifted to 3D using camera intrinsics and extrinsics. On the left, LiDAR points from adjacent cameras are overlaid to show how much overlap there is between views.
}
\label{fig:pcl360_overlapping}
\vspace{-3mm}
\end{figure*}